\documentclass{article}
\usepackage{graphicx}
\usepackage{float}

\usepackage[main, final]{neurips_2026}
\usepackage{amsmath}
\usepackage{xcolor}

\usepackage{wrapfig}
\usepackage[utf8]{inputenc} 
\usepackage[T1]{fontenc}    
\usepackage{hyperref}       
\usepackage{url}            
\usepackage{booktabs}       
\usepackage{amsfonts}       
\usepackage{nicefrac}       
\usepackage{microtype}      
\usepackage{xcolor}         

\title{Sparse Prototype Code Underlies \\ Classification and Prediction Across Modalities}

\author{%
  Yehonatan Avidan\textsuperscript{1,2}, Daniel D. Lee\textsuperscript{3}, Haim Sompolinsky\textsuperscript{1,2,4}\\
  \textsuperscript{1}The Racah Institute of Physics, Hebrew University\\ 
  \textsuperscript{2}Edmond and Lily Safra Center for Brain Sciences, Hebrew University \\
  \textsuperscript{3}Department of Electrical and Computer Engineering, Cornell Tech, Cornell University \\ 
  \textsuperscript{4}Center for Brain Science, Harvard University \\
  \texttt{Yehonatan.Avidan@mail.huji.ac.il}
}
\begin{document}

\maketitle

\begin{abstract}
Neural representations have become a central tool for studying the internal mechanisms of modern AI models, yet their complex high-dimensional structure makes them difficult to interpret. We show that classification tasks give rise to a universal representational geometry, shared across state-of-the-art models in vision, audio, and language processing. The key structure is that within-class variability is not random in representation space. Instead, its classifier-relevant component has strong and structured correlations with the class’s own centroid and with the centroids of its competing classes. Building on this observation, we derive an analytical mean-field theory governed mainly by the variability along true-class and rival-class centroid coordinates, together with a global renormalization of the class radius that compensates for the non-Gaussian statistics of real representations. The theory accurately predicts classification accuracy across architectures and modalities. The relevant geometric quantities improve systematically with model scale, mirroring the observed gains in accuracy. A striking feature of the theory is its sparsity: accurate prediction requires only a small set of centroid coordinates associated with the true class and its strongest rivals - connecting our framework to sparse-feature extraction approaches such as sparse autoencoders. Together, these results provide a parsimonious predictive theory of neural representations and suggest that classification in deep networks is governed by a sparse, centroid-aligned structure embedded within the full high-dimensional representation space.
\end{abstract}
\vspace{-0.5cm}

\section{Introduction}\label{Sec:intro}
\vspace{-0.2cm}
One of the central challenges in modern Deep Neural Network (DNN)
research is to open the "black box" of
computation and to understand the internal mechanisms by which networks
transform their inputs into useful predictions. Such understanding
is essential for building safer, more reliable AI systems \citep{bommasani2021opportunities,hendrycks2021unsolved}. A primary tool toward this goal is the analysis of representations - the hidden-layer activations induced by specific inputs. These representations form structured point clouds whose geometry reflects both the data and the model, but extracting interpretable principles from this structure remains a significant challenge.

A growing body of work in mechanistic interpretability addresses this
challenge by extracting sparse, human-interpretable features from
network activations \citep{elhage2022toy,bricken2023monosemanticity,huben2024sparse,templeton2026scaling}. Sparse autoencoders, feature dictionaries, and circuit analyses have shown that semantically meaningful
concepts are often encoded along specific directions in activation
space and can be isolated within otherwise opaque representations. These approaches are powerful tools
for identifying the essential features of the network; they are typically not
aimed, however, at providing a quantitative theory linking the geometric
structure of a representation to how well it supports a downstream
task. Such a theory would help address fundamental questions, such
as why some networks generalize better than others.


Predictive theories of representational geometry ---manifold capacity \citep{chung2018classification,cohen2020separability,stephenson2021geometry,wakhloo2023linear} and neural collapse \citep{papyan2020prevalence,han2021neural, galanti2021role} --- have made important progress in connecting geometry to performance within their respective regimes. Capacity theory studies the linear separability of object manifolds under random binary dichotomies, relating global capacity to geometric quantities such as manifold radius, effective dimension, and inter-manifold correlations. Neural collapse describes the final phase of supervised training, in which within-class variability collapses and class means converge to a simplex equiangular tight frame. Modern pretrained foundation models, however, typically operate in a different regime: they are evaluated using multiclass accuracy and are often trained with early stopping or self-supervised objectives, leaving substantial within-class variability. Recent work connects pairwise class geometry to performance and generalization bounds in few-shot learning \citep{sorscher2022neural,luthra2026directional}. A predictive geometric theory of multiclass accuracy in the finite-radius regime that accounts for both competition among multiple classes and the rich within-class statistics of many examples is currently missing.

Our analytical lens is prototype-based classification. A long line
of work - from nearest-class-mean classifiers \citep{mensink2013distance} to prototypical networks
for few-shot learning \citep{snell2017prototypical}
- has established that classifying inputs by proximity to class centroids 
is a strong strategy whenever representations
are well-structured. 
Adopting the centroid-based classifier in this work has  a clear advantage:
the centroids are geometric objects, allowing classification
accuracy to be related directly to geometric measures of the representation
- while remaining relevant for realistic classification
settings. Thus, we leverage
this framework to close the theoretical gap, deriving and empirically
validating a quantitative geometric theory of multi-class classification
accuracy using prototype classification, and apply it across
architectures and modalities.

\begin{figure}[t]
\centering
\includegraphics[width=\linewidth]{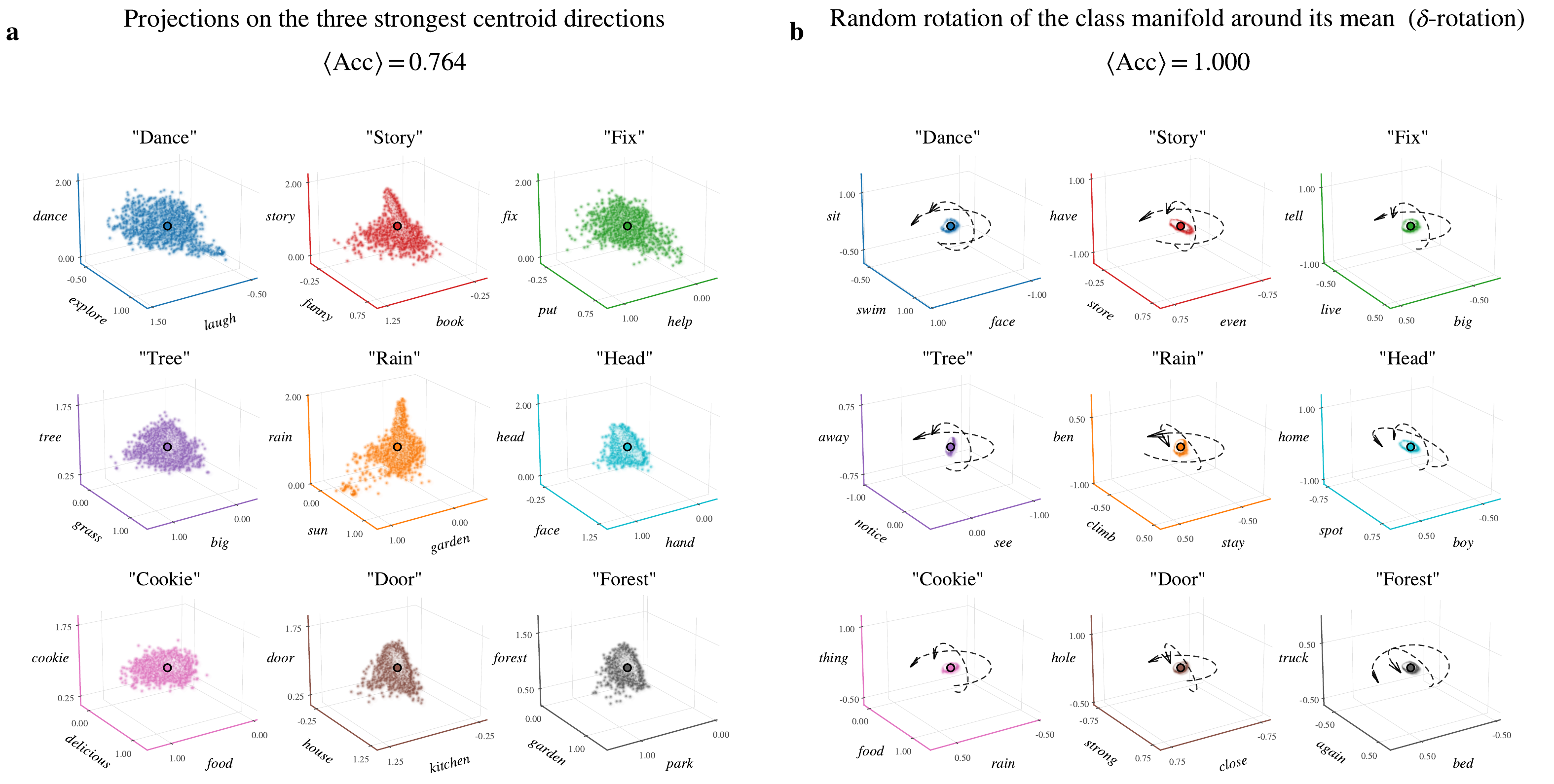}
\caption{\textbf{Centroid-aligned variability, Qwen3-32B, TinyStories
representations.} The variability of each class is highly correlated with other class centroids. \textbf{(a)} To visualize it, we use Qwen3-32B representations \citep{yang2025qwen3} on the dataset TinyStories \citep{eldan2023tinystories}. We project each example onto the three centroids with the largest correlations among the class variability. For 217/320 classes ($68\%$), the true class centroid is among these top 3 (the nine shown here are of this sort). The rivals carry semantic meaning (e.g., the rivals of "cookie" are "food" and "delicious"; the rivals of "head" are "face" and "hand"). The mean prototype classification accuracy of these classes (see Def.~2.1) is $76.4\%$. \textbf{(b)} To causally test the role of the variability-centroid correlations, we rotate the class manifold around its mean (denoted as $\delta$-rotation), breaking these correlations while leaving the class radius and the distance between class centroids unchanged. The manipulation removes all classification errors, and the projections onto the top-3 directions shrink
to $11.3\%$ of their original magnitude on average, and those directions no longer carry
semantic meaning.}
\label{fig:visualization}
\end{figure}

Our main contributions are:

\textbf{Universal centroid-aligned structure:} We identify a universal
representational geometry shared across vision, language, and audio backbones. Its defining property is that within-class
variability is strongly correlated with both the true-class centroid
and the centroids of competing classes (see Fig.~\ref{fig:visualization}a). We explicitly demonstrate
this by surgically removing only these centroid-aligned correlations
(using rotations, see Fig.~\ref{fig:visualization}b) - leaving centroid-centroid correlations, total
variance, and example-example structure intact - and find that doing
so drives per-class accuracy to essentially 100\%. The centroid-aligned
correlations are therefore precisely the geometric source of classification
errors.

\textbf{Sparsity of competition:} These correlations do not spread evenly across
the other classes; for each class, the geometry concentrates on a small set of active rivals, which is sufficient to recover the full geometric picture and reproduce the predicted per-class accuracy. These rivals also appear to carry semantically meaningful relations to the true class (see Fig.~\ref{fig:visualization}a).
Each example's classification is therefore faithfully captured by
a sparse, semantically meaningful centroid-aligned basis embedded in the full high-dimensional
representation - a grounded analog of the sparse-feature
picture pursued in mechanistic interpretability.

\textbf{Compact predictive mean-field theory:} To connect representational geometry
to classification, we adopt a prototype classifier - scoring examples
by dot product with unit-normalized class centroids \citep{snell2017prototypical} - which serves as an empirically reliable proxy for the trained linear
head (see Appendix \ref{app sec: prototype linear comparison}). Building on this, we derive an analytical mean-field theory in
the sparse centroid basis (Sec.~\ref{sec:theory}). Its inputs are the variability along the true- and rival-class centroid directions
and a single global radius renormalization that
absorbs the heavy-tailed statistics of real representations (see Fig.~\ref{fig:sparsity and heavy tail}c). Crucially, this factor
is tightly correlated with a direct measure of tail heaviness (Fig.~\ref{fig:theory}e). From these inputs alone, the theory predicts per-class
accuracy across classes, architectures, and modalities (see Fig.~\ref{fig:theory}(a-d), Appendix \ref{app:manymodels}).

\textbf{Geometry across model scales:} Tracking the theory's geometric ingredients across model families, we find that they systematically improve as the models scale up, mirroring the gains in accuracy (Table~\ref{tab:scale_depth}). Importantly, we show that larger models achieve their
accuracy advantage without having the smallest overall within-class variance. Our theory identifies the specific geometric measures that
drive this improvement - taking a step toward understanding the key
geometrical ingredients required for good generalization in real state-of-the-art
models. We hypothesize that these measures are practically actionable:
targeting them during training could improve generalization; matching
them between teacher and student could guide more geometry-aware distillation
\citep{hinton2015distilling} and could help identify the best-suited
pretrained backbone for transfer \citep{kornblith2018better}.

\section{Setup and Notation}\label{sec:definitions}

We study how the geometry of neural representations determines classification accuracy. For each input we consider a representation vector taken from some layer of a neural network, and the task is to assign it to one of $P$ classes.

\textbf{Prototype classification:} Rather than training a linear readout on top of the representation, we use a prototype classifier \citep{snell2017prototypical}, in which the class-specific readout vector is not a learned weight but the centroid - the empirical mean of the class in representation space (Def.~2.1), normalized to a unit vector. Centroids are estimated from a labeled set, and at inference, the learner classifies one example against all $P$ centroids (computed without it, "Leave one out" \citep{lachenbruch1968estimation}). This has a direct geometric definition while remaining relevant for realistic classification, can be applied at every layer of every model without additional fine-tuning, and empirically tracks trained linear readouts closely (see Appendix~\ref{app sec: prototype linear comparison}), making it both a flexible evaluator in its own right and a useful proxy for the standard linear head. 

\textbf{Experimental setup:} We evaluate frozen representations from pretrained backbones spanning vision (ImageNet-1K \citep{deng2009imagenet}), language (masked and next-token prediction on TinyStories \citep{eldan2023tinystories}), and audio (VGGSound \citep{chen2020vggsound}), across transformer and convolutional architectures, supervised, self-supervised, and contrastive objectives. Full lists of datasets, models, and extraction details are in Appendix~\ref{app:setup}. For language, the task is to predict a content word from its representation. In autoregressive models this is the embedding of the token preceding the word; in bidirectional models it is the embedding at the word's masked position. We group each word's inflections under a single lemma label (e.g. say/says/said, friend/friends), and keep only lemmas with at least 1000 examples, so their statistics are well estimated. Full technical details are in Appendix~\ref{app:setup}.



\textbf{Definition 2.1} (\emph{prototype classification}):
Let $\mathbf{x}\in\mathbb{R}^{N}$ be the representation of an input at a given layer of a neural network, and consider a $P$-class task with labels $y\in\{1,\dots,P\}$ and dataset $\mathcal{D}=\{(\mathbf{x}_{i},y_{i})\}_{i=1}^{n}$. For each class $\mu$, define the centroid $\mathbf{c}_{\mu}=\mathbb{E}[\mathbf{x}\mid y=\mu]$ and its unit direction $\hat{\mathbf{c}}_{\mu}=\mathbf{c}_{\mu}/\|\mathbf{c}_{\mu}\|$, which serves as the prototype class readout. Throughout the paper, we use the terms centroid and prototype interchangeably. The prototype logits and accuracy are defined as

\begin{equation} t_{\mu}(\mathbf{x})=\hat{\mathbf{c}}_{\mu}\cdot\mathbf{x}, \qquad \text{Acc}=\mathbb{E}_{(\mathbf{x},y)\in\mathcal{D}}[\mathbf{1}\{\arg\max_{\mu}t_{\mu}=y\}]. \label{eq: t and acc definition}
\end{equation}

\vspace{-0.1cm}
\textbf{Definition 2.2} (\emph{manifold geometry}):
The class manifold is defined as the point cloud of a particular class $k$: $\mathcal{M}_{k}=\{\mathbf{x}_{i}:y_{i}=k\}$. Each $\mathbf{x}\in\mathcal{M}_{k}$ decomposes as $\mathbf{x}=\mathbf{c}_{k}+\delta\mathbf{x}$ with $\mathbb{E}_{\mathbf{x}\in {\scriptstyle \mathcal{M}_k}}[\delta\mathbf{x}]=\mathbf{0}$. We name $\delta\mathbf{x}$ the \textbf{residual} vector. The class radius $R$ measures the overall within-class variance, and $\sigma_{\mu}^{2}\in[0,1]$ is the fraction of the residual variance along a particular centroid direction $\hat{\mathbf{c}}_{\mu}$:

\vspace{-3mm}

\begin{equation}
R^{2}\;=\;\frac{\mathbb{E}_{k}[\|\delta\mathbf{x}\|^{2}]}{\|\mathbf{c}_{k}\|^{2}},
\qquad
\sigma_{\mu}^{2}\;=\;\frac{\mathbb{E}_{k}[(\delta\mathbf{x}\cdot\hat{\mathbf{c}}_{\mu})^{2}]}{\mathbb{E}_{k}[\|\delta\mathbf{x}\|^{2}]}.\label{eq:R and sigma definition}
\end{equation}

\vspace{-3mm}

Throughout, geometric quantities depend on a fixed true class $k$. We treat $k$ as fixed and drop the subscript to lighten notation: $R \equiv R_k$, $\sigma_{k\mu} \equiv \sigma_{\mu}$, and $\sigma_k^2$ is the fraction of residual variance along the true centroid $\hat{\mathbf{c}}_k$. In addition, $\mathbb{E}_k$, denotes the average over all the points in the manifold $k$, $\mathbb{E}_k\equiv\mathbb{E}_{{\mathbf x}\in\mathcal{M}_k}$.

\vspace{-2mm}
\section{Isotropic Uncorrelated Model}
\label{sec:theory:isotropic}

We first examine a baseline of an uncorrelated model, in which the centroids
$\mathbf{c}_\mu$ and the residuals $\delta\mathbf{x}$ are drawn independently
from an isotropic Gaussian distribution. The model yields testable predictions
for how geometry shapes accuracy, which we evaluate in Sec.\ref{section:empirical} on real data: deviations
from these predictions quantify the role of correlations in the classification task. 

\textbf{Assumption 3.1} (\emph{uncorrelated statistics}).
For each class $\mu=1,\dots,P$, the centroid is drawn as
$\mathbf{c}_{\mu}\!\sim_{\mathrm{iid}}\!\mathcal{N}(\mathbf{0},I_{N})$.
Conditional on the centroids, examples of class $k$ are generated by sampling
$\mathbf{x}=\mathbf{c}_{k}+\delta\mathbf{x}$, with residuals drawn as
$\delta\mathbf{x}\!\sim_{\mathrm{iid}}\!\mathcal{N}(\mathbf{0},R^{2}I_{N})$.

\textbf{Assumption 3.2} (\emph{high-dimensional geometry}).
We work in the high-dimensional limit $N \to \infty$, with $P=N^{\mathcal{O}(1)}$
classes and a radius $R=R(N)$ that may scale with $N$.\\ We refer to
Assumptions 3.1-3.2 jointly as the isotropic model.

\textbf{Theorem 3.3} (\emph{isotropic model accuracy}).
Under the isotropic model,
\vspace{-2mm}
\begin{equation}
\mathrm{Acc}^{\mathrm{iso.}}(R,N,P)
=\mathbb{E}_{z\sim\mathcal{N}(0,1)}\!\left[
\Phi^{P-1}\!\left(\frac{Rz+\sqrt{N}}{\sqrt{R^{2}+1}}\right)\right],
\label{eq:isotropic-acc}
\end{equation}

where $\Phi$ is the standard Gaussian CDF.
\\ The supporting lemmas (centroid-overlap CLT, self-averaging of
$\mathrm{Acc}$, and the joint Gaussian distribution of the logits), together
with the proof of Theorem~3.3, are given in Appendix~\ref{app:isotropic-proofs}.

\textbf{Predictions.}
Several consequences follow directly from Theorem~3.3:
\\
(i)~For any $R=\mathcal O(1)$, the classes are well
separated, and classification is trivial: $\mathrm{Acc}^{\mathrm{iso.}}\to 1$.\\
(ii)~In the infinite radii limit, $R/\sqrt N\to\infty$, the isotropic residual
dominates the input, which thus retains no information about the true class:
all $P$ classes are equally likely to be predicted, and accuracy collapses
to chance, $\mathrm{Acc}^{\mathrm{iso.}}\to 1/P$.\\
(iii)~Non-perfect accuracy requires large radii, $\mathrm{Acc}^{\mathrm{iso.}}<1\Rightarrow R=\Omega(\sqrt {N/\ln P})$.\\
(iv)~In the high-dimensional regime, the accuracy is self-averaging: its variance across
centroid realizations is $\mathcal O((PR^2)^{-1})$. We note that according to (iii), $R$ also scales with $N$ for nontrivial accuracy.\\
These predictions are rigorously derived in Appendix~\ref{app:isotropic-proofs}
(Corollaries~C.1.6-C.1.9), and verified numerically with synthetic data in Appendix Fig.\ref{fig:v19-synth-isotropic}.

\begin{figure}[t]
\centering
\includegraphics[width=0.8\linewidth]{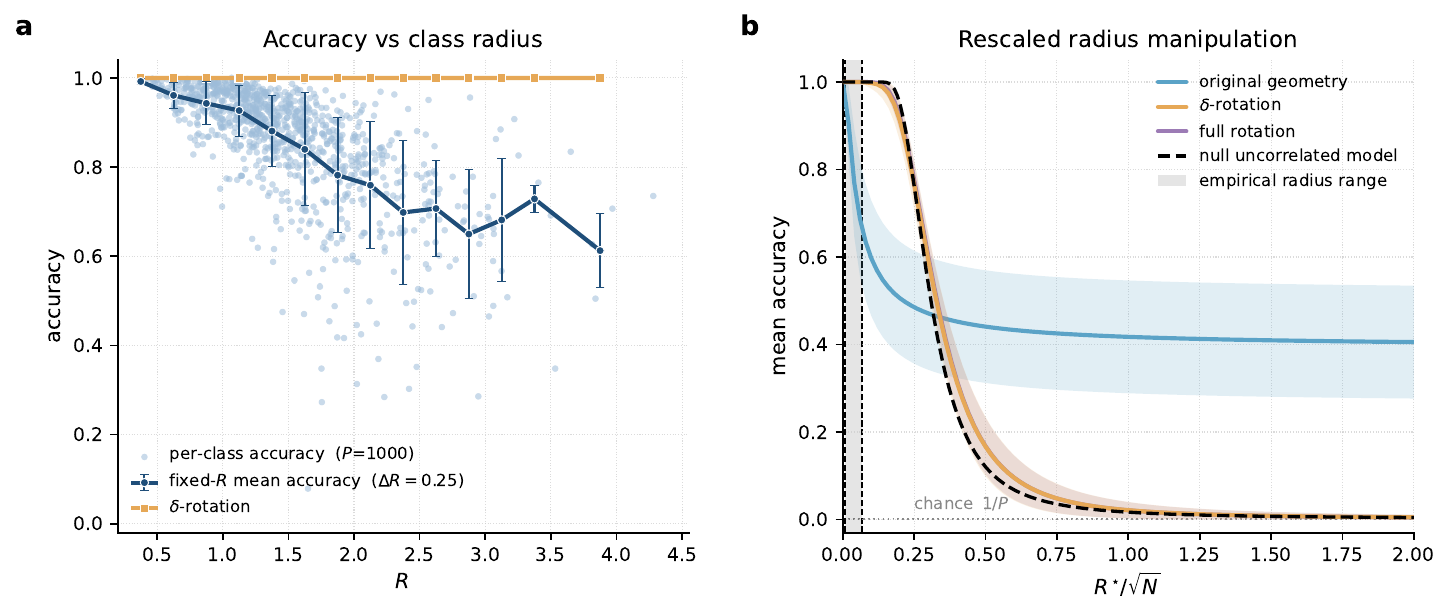}
\caption{\label{fig:empirical}\textbf{Geometric signatures of multiclass classification}
\textbf{(a)} Per-class accuracy vs.\ empirical radius $R$: sharp decrease across the narrow empirical $R$ range and a wide spread at fixed $R$. After random rotation of within-class variability $\delta \textbf{x}$ ($\delta$-rotation), the accuracy is perfect.
\textbf{(b)} We artificially scale the variability  $\delta\mathbf{x}\!\to\!(R^\star/R)\delta\mathbf{x}$. The accuracy decreases fast but stabilizes at $~39\%$ accuracy for $R^\star\rightarrow \infty$  due to strong true centroid correlations (blue line). After random rotation of the variability $\delta$-rotation, the accuracy is well described by the isotropic model ($\mathrm{MSE}=1.26\times10^{-3}$, orange line). Additional breaking of centroid-centroid correlations (full rotation) has a very small effect; lines overlap. 
} 

\end{figure}

\vspace{-2mm}
\section{Empirical Results}\label{section:empirical}

In this section we characterize the geometric structure of trained representations that governs prototype classification. We begin from the predictions of the isotropic model (Sec.\ref{sec:theory:isotropic}) and show that they are not satisfied by real representations. This pattern of failure is informative: it identifies which correlations dominate the geometry responsible for accuracy, and isolates the components needed to build the analytical framework of Sec.\ref{sec:theory}. Results are presented for DINOv3 7B \citep{simeoni2025dinov3} representations on ImageNet-1K \citep{deng2009imagenet}, and later replicated in many models and modalities (Sec.~\ref{sec:universality}).

\textbf{Accuracy changes on an $\mathcal{O}(1)$ radius scale:} The empirical class radii lie in a narrow $R\in[0.4,4.3]$ range.
 In Sec.\ref{sec:theory:isotropic}, we saw that an uncorrelated model predicts
that accuracy should remain unchanged throughout $\mathcal{O}(1)$ radii, and decline only in $ R=\Omega(\sqrt {N/\ln P})\sim20$. However, we observe a dramatic change in accuracy, ranging from $100\%$ to $61\%$ averaged over a small $R$ window (see Fig.\ref{fig:empirical}a). The sharp accuracy decline therefore
indicates structure that is not described by the isotropic model.

\textbf{Large class-to-class variability at fixed radius:} The isotropic model predicts that class-to-class fluctuations are of order $1/R\sqrt{P}$ ($\sim1\%-3\%$).
Empirically, however, the spread of accuracies at fixed $R$ is
large. For example, the 89 classes in the range $R\in\left[1.2,1.3\right]$ have
a mean accuracy of $90.4\%$, with a minimum accuracy of $61.8\%$, and a maximum of $99.6\%$ (see Fig.\ref{fig:empirical}a). The radius alone therefore cannot explain classification
performance: additional class-dependent geometric structure must determine
which classes are easy to classify and which are difficult.

\vspace{-5mm}
\subsection{Geometric manipulations}
To causally test which components cause the deviation from the isotropic model, we use
geometric manipulations to probe specific structures of the representations. We apply two geometric manipulations: \\
(i) \textbf{Variability inflation}, where we set each class radius to a chosen $R^\star$ artificially by scaling its deviations,
$\mathbf{x}(R^\star)=\mathbf{c}_{k}+(R^\star/R)\delta\mathbf{x}$, where $R$ is the empirical radius as defined in Sec.\ref{sec:definitions}. \\
(ii) \textbf{Geometric rotations}, 
where we surgically destroy specific correlations in the data while preserving all other aspects of the representation
geometry.

\textbf{Strong residual–true-centroid correlations:} We first examine the representations under variability inflation $\mathbf{x}(R^\star)=\mathbf{c}_{k}+(R^\star/R)\delta\mathbf{x}$.
Naively, as $R^\star\to\infty$ one expects chance accuracy, as we saw in the isotropic model. Empirically, the large-radius accuracy
instead stays far above chance -  accuracy approaches
$39.2\%$, over two orders of magnitude above $1/P\!=\!0.1\%$ (Fig.\ref{fig:empirical}b).  The within-class variability is thus far more correlated with
the true centroid than with the rival centroids. To quantify this directly, we measure the projection variance along
each centroid direction $\sigma_{\!\mu}$ (see Sec.~\ref{sec:definitions}).
The true-class $\sigma_{k}$ exceeds all rivals in $93\%$ of classes.
Moreover, this dominance
is itself predictive of a well-structured class geometry: in the $930$ classes
where $\sigma_{k}$ exceeds all rivals, the mean accuracy is $87.4\%$,
whereas in the remaining $70$ classes it drops to $60.1\%$. 

\textbf{Residual-centroid correlations are the main drivers of errors:}
We manipulate the representations by rotating the residuals $\delta\mathbf{x}$
around their own class centroid ($\delta$-rotations): for each class $k$ we replace $\mathbf{x}=\mathbf{c}_{k}+\delta\mathbf{x}$
by $\mathbf{x}^{\mathrm{\delta rot}}=\mathbf{c}_{k}+O_{k}\delta\mathbf{x}$,
with $O_{k}$ a class-specific random orthogonal matrix (see Fig.~\ref{fig:visualization}b for illustration). This
preserves the centroid-centroid correlations, the class radii, and
the example-example relations, while breaking the correlations between
$\delta\mathbf{x}$ and the centroid directions (residual-centroid correlations).
At the empirical radius, the class-averaged
accuracy rises from the baseline of $85.5\%$ to $100\%$, 
completely eliminating classification errors (see Fig.\ref{fig:empirical}a, orange line).
This is achieved \emph{without}
breaking centroid-centroid correlations. 

\textbf{Breaking residual–centroid correlations recovers the isotropic model:}
To look for any further role of residual–centroid correlations, we go
outside the empirical radius range, where $\delta$-rotation has already
saturated the accuracy. We set each class radius to a chosen $R^\star$ by variability inflation. 
Strikingly, after removing only the residual-centroid correlations, the data is already well described by the isotropic model accuracy (Eq.\ref{eq:isotropic-acc}), with $\mathrm{MSE}=1.26\times10^{-3}$. The accuracy declines only at $R=\Omega(\sqrt{N/\ln {P}})$ 
, far beyond the empirical $R$ range  (see Fig.\ref{fig:empirical}b). The accuracy at $R^\star\rightarrow\infty$ is chance ($0.1\%$), and the per-class standard deviation around the mean accuracy shrinks by an order of magnitude to $2.3\times10^{-3}$ compared to $1.7\times10^{-2}$ in the unmanipulated representations.

\textbf{Centroid-centroid correlations play a minor role:} To test the role of centroid-centroid correlations, we apply a stronger manipulation: rotation of the whole class manifold around the
origin by another random orthogonal matrix after $\delta$-rotations,
$\mathbf{x}^{\mathrm{full\text{-}rot}}=U_{k}\bigl(\mathbf{c}_{k}+O_{k}\delta\mathbf{x}\bigr)$,
which breaks centroid–centroid correlations on top of the $\delta$-rotation.
Even at these larger radii the extra effect is small: the full-rotation
curve differs from the $\delta$-rotation curve by only
$\mathrm{MSE}=2.0\!\times\!10^{-5}$ (see Fig.~\ref{fig:empirical}b). We infer that centroid–centroid correlations
play a minor role in accuracy. This echoes the picture from neural
collapse~\citep{papyan2020prevalence}, where the centroids are as spread as possible.

Together, these empirical results identify the key geometric structure of trained representations: the class-relevant geometry is governed by strong correlations between within-class variability and a sparse set of centroid directions—dominated by the true-class centroid.

\vspace{-3mm}

\section{Centroid-Aligned Variability Model}
\label{sec:theory:caligned}
\label{sec:theory}

The empirical results of Sec.~\ref{section:empirical} identify the geometry of prototype classification in pretrained
representations: Within-class variability $\delta\mathbf{x}$ is
\emph{centroid-aligned}. Surgically removing only the $\delta\mathbf{x}$-centroid
correlations drives accuracy to $\sim\!100\%$, while breaking
centroid-centroid correlations has only a secondary effect. 



We introduce the centroid-aligned model that incorporates this geometry. The resulting theory takes as input geometric quantities measurable from the representation, together with a single global renormalization of the class radius (Sec.~\ref{sec:theory:lambda}), and predicts per-class accuracy across architectures and modalities.
All proofs, lemmas, and
intermediate computations are deferred to Appendix~\ref{app:caligned-proofs};
the main text retains only the core assumptions and central
results. We stress that the following assumptions are introduced to define an analytically solvable model and are not intended as exact claims about the statistics of the data. Nevertheless, the model is motivated by empirical observations (Sec.~\ref{section:empirical}) and, as we show below, yields accurate predictions across models spanning different architectures and modalities. We therefore believe that it captures important aspects of real model representations.


\textbf{Definition 5.1} (\emph{rival support}). Fix a class $k$ and an integer $K$. Define its rival support $\mathcal{R}_K$ as the set of indices $\mu\neq k$ corresponding to the $K$ largest values of $\sigma_\mu$ (Def.~2.2), and set $\mathcal{R}_K^{+}\equiv\{k\}\cup\mathcal{R}_K$. For each $\mu\in\mathcal{R}_K$, define the true-rival centroid correlation by $g_\mu=\hat{\mathbf{c}}_k\cdot\hat{\mathbf{c}}_\mu$. 

\textbf{Definition 5.2} (\emph{standardized centroid projections}).
Conditioned on class $k$, each example is represented by the following projection coefficients:
\begin{equation}\label{eq:s}
s_\mu=\frac{\delta\mathbf{x}\cdot\hat{\mathbf{c}}_{\mu}}{\mathrm{std}_{k}\left[\delta\mathbf{x}\cdot\hat{\mathbf{c}}_{\mu}\right]}, \quad \mu\in\mathcal{R}_K^+
\end{equation}
\noindent
\textbf{Assumption 5.3} (\emph{centroid-aligned variability model}). Let $G_{\mu\nu}=\hat{\mathbf{c}}_\mu\cdot\hat{\mathbf{c}}_\nu$ be the centroids Gram matrix. 
Assume that the restricted centroid Gram matrix of the rival support $(G_{\mu\nu})_{\mu,\nu\in\mathcal{R}_K^{+}}$ is invertible.
We assume the following generative model of examples of class $k$
\begin{equation}
\mathbf{x}(\mathbf{s})=\left\Vert \mathbf{c}_{k}\right\Vert \left[\hat{\mathbf{c}}_{k}+R\left(\sum_{\mu,\nu\in\mathcal{R}^{+}_{K}}\!\!\!\sigma_{\mu}s_{\mu}\,G^{-1}_{\mu\nu}\hat{\mathbf{c}}_{\nu}+\sigma_{_\perp}s_{_{\perp}}\hat{\mathbf{e}}_{_{\perp}}\right)\right]
\label{eq:caligned}
\end{equation}
where $~\hat{\mathbf{e}}_{_\perp}$ is perpendicular to the $~\mathrm{span}(\hat{\mathbf{c}}_{\mu\in\mathcal{R}_k^+})$,
$\sigma_{\mu}^{2}$ is the variance fraction along the centroid $\mu$ direction, and $R$ is the empirical class
radius (see Def.~2.2). Eq.~\ref{eq:caligned} encodes the geometry identified in Sec.~\ref{section:empirical} by placing the residuals
$\delta\mathbf{x}$ along rival centroid directions in the rival support.  

\textbf{Assumption 5.4} (\emph{Gaussian statistics}).
We assume the projection coefficients $s_\mu$ are jointly
Gaussian with zero mean and unit variance, and that their covariance has the following form
\begin{equation}
\mathrm{Cov}\left(s_{\mu},s_{\nu}\right)=\delta_{\mu\nu}\left(1-\rho^{2}_{\mu}\right)+\rho_{\mu}\rho_{\nu},
\end{equation}

\vspace{-2mm}
where $\rho_\mu=\mathbb{E}_{k}\left[s_{k}s_{\mu}\right]$, and by definition $\rho_k=1$. These are the Pearson correlations between the residual projections $\delta \mathbf{x}\cdot\hat{{\mathbf c}}_k$ and $\delta \mathbf{x}\cdot\hat{{\mathbf c}}_\mu$. Equivalently, all dependence among the rival projections is captured by $s_k$: conditioned on $s_k$, they are mutually independent.

\textbf{Theorem 5.5} (\emph{centroid-aligned
model accuracy}). Under assumptions 5.3-5.4
\begin{equation}
\mathrm{Acc}^{\mathrm{th}}_k(R)=\mathbb{E}_{z\sim\mathcal{N}(0,1)}\!\left[\prod_{\mu\in\mathcal{R}_{K}}\Phi\left(\frac{\sigma_{k}-\rho_{\mu}\sigma_{\mu}}{\sigma_{\mu}\sqrt{1-\rho^{2}_{\mu}}}z+\frac{1-g_{\mu}}{R\sigma_{\mu}\sqrt{1-\rho^{2}_{\mu}}}\right)\right].
\label{eq:caligned-acc}
\end{equation}

The inputs to Eq.~\ref{eq:caligned-acc} are entirely geometric and class-specific: the
empirical radius $R$, the rival projected variances
$\{\sigma_{\mu}\}$, the centroid-centroid overlaps $\{g_{\mu}\}$, and the correlation terms $\rho_\mu$ are all measured directly from the representations. For convenience, we summarize our notations and how we measure them in Table.~\ref{tab:defs}.

\begin{figure}[hbt!]
\centering
\includegraphics[width=\linewidth]{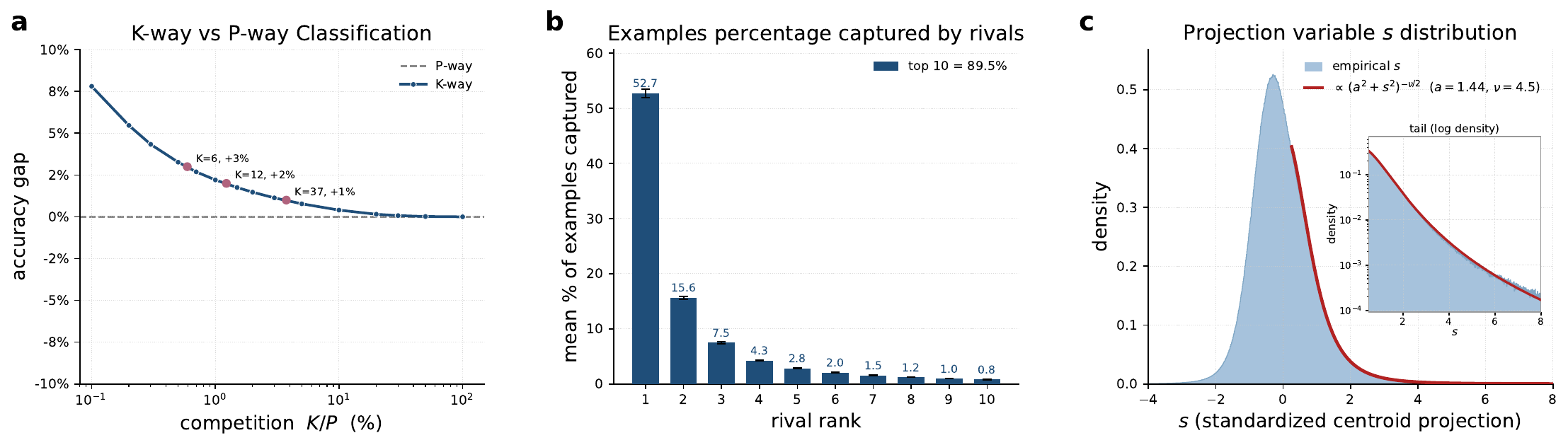}
\caption{\label{fig:sparsity and heavy tail} \textbf{Sparsity and heavy tail statistics:} \textbf{(a)} K-way classification of rivals chosen by centroid projection variance $\sigma_\mu$ converges quickly to the full $P$-way classification. \textbf{(b)} We define the strongest rival of each example $\mu^{*}(\mathbf{x})=\arg\max_{\mu\neq k} t_{\mu}(\mathbf{x})$, and count the number of examples captured by each rival. The strongest rival of each class captures more than half of the examples on average, with $90\%$ of the examples captured by only 14.4 rivals. \textbf{(c)} The standardized centroid projections $s_\mu$ (see Eq.\ref{eq:s}) have heavy tail power law statistics, which inflate the extreme values
that control classification accuracy. The magnified panel shows the log density of the far tail.}
\end{figure}

\vspace{-3mm}

\subsection{Sparsity}\label{sec:sparsity}

The rival count $K$ is a hyperparameter of the theoretical 
model 
controlling how many rival directions enter the prediction in 
Eq.~\ref{eq:caligned-acc}. Because $\Phi$ approaches one quickly as its argument grows, small projections $\sigma_\mu$ have very little effect on the overall accuracy. 
Thus, when ranking rivals by their projection 
magnitudes $\sigma_{\mu}$, predictive power plateaus rapidly once $K$ exceeds 
the head of the spectrum. 

We assess the effective number of rivals both theoretically and empirically. We first replace the full P-way task with a K-way task restricted to the top rivals ranked by $\sigma_\mu$. We find that only 6 rivals are needed to come within $3\%$ of the full P-way classification accuracy, 12 within $2\%$, and 37 within $1\%$ (see Fig.\ref{fig:sparsity and heavy tail}a). The competition is thus not about the full set of $P$ (1000) classes, but rather a sparse set of strongly correlated rivals. In Appendix \ref{app sec: K}, we compare several rival-selection methods and show that ranking rivals by $\sigma_\mu$ yields results comparable to more complex selection rules, while remaining simple, interpretable, and easy to compute. In particular, it outperforms selection based on centroid proximity, providing additional evidence that centroid correlations play only a secondary role in the task.

We further examine whether the correlations are homogeneously distributed across the rivals or dominated by a few. For each example $\mathbf{x}$ of class $k$, we record its strongest rival, $\mu^{*}(\mathbf{x})=\arg\max_{\mu\neq k} t_{\mu}(\mathbf{x})$, and count the number of distinct rivals that appear across the $\sim1300$ examples of that class. Competition is highly concentrated: on average only $67$ distinct rivals appear, $14.4$ of them cover $90\%$ of the examples, and one rival alone accounts for more than half (Fig.~\ref{fig:sparsity and heavy tail}b).

The theory's predictions also saturate quickly as the number of rivals $K$ increases. We find that $K\in[10,20]$ is already enough in most models to achieve accurate predictions, with very small gain from increasing $K$ further (see Appendix \ref{app sec: K}). The results in the main text are shown for $K=20$ for all models. 

We also note that in many cases, the rivals chosen by $\sigma_\mu$ carry semantic meaning.
In the Qwen3-32B \citep{yang2025qwen3} model on TinyStories \citep{eldan2023tinystories},
many rival words are similar in both grammatical role ("smile"$\mapsto$"reply", "nod",
"say") and semantic meaning ("forest"$\mapsto$"park", "garden", "grass"). In vision
(DINOv3 ViT-7B), we observe either similar objects ("leopard"$\mapsto$"jaguar", "cheetah")
or frequently co-occurring objects ("king crab"$\mapsto$"plate", "flute"$\mapsto$"stage").
For more examples, see Table~\ref{tab:rival_semantics} and Fig.~\ref{fig:visualization}a.

\begin{table}[t]
\centering
\small
\begin{tabular}{@{}lll@{}}
\toprule
model & class & top-5 rivals (by $\sigma_\mu$) \\
\midrule
{Qwen3-32B}
 & \textit{happy}  & excite, proud, glad, surprised, curious \\
 & \textit{bad}    & sad, sorry, scare, angry, tired \\
 & \textit{voice}  & loud, noise, sound, big, funny \\
 & \textit{smile}  & reply, nod, say, think, tell \\
 & \textit{cake}   & delicious, cookie, food, ice, big \\
 & \textit{forest} & park, garden, grass, world, sky \\
 & \textit{ground} & floor, grass, swing, slide, garden \\
\midrule
{DINOv3 ViT-7B}
 & \textit{leopard}   & snow leopard, jaguar, cheetah, impala, tusker \\
 & \textit{king crab} & Dungeness crab, rock crab, hermit crab, plate, American lobster \\
 & \textit{flute}     & panpipe, oboe, ocarina, bassoon, stage \\
 & \textit{orange}    & lemon, banana, grocery store, Granny Smith, strawberry \\
\bottomrule
\end{tabular}
\caption{Rival classes selected by the projection's variance $\sigma_\mu$ are semantically related words (Qwen3-32B, TinyStories) or visually similar and co-occurring objects
(DINOv3 ViT-7B, ImageNet). For more examples see also Fig.~\ref{fig:visualization}a.}
\label{tab:rival_semantics}
\end{table}


\subsection{Non-Gaussian tails and a global radius renormalization}
\label{sec:theory:lambda}
\setlength{\intextsep}{0pt}
\begin{wraptable}{r}{0.56\textwidth}
\small
\renewcommand{\arraystretch}{1.3}
\centering
\begin{tabular}{c | l | l}
\hline
\textbf{Notation} & \textbf{Name} & \textbf{Definition} \\ \hline
$\mathbf{c}_{k}$ & Centroid & $\mathbb{E}_{k}[\mathbf{x}]$ \\ \hline
$\hat{\mathbf c}_{\mu}$ & Unit centroid & $\mathbf{c}_{\mu}/\lVert\mathbf{c}_{\mu}\rVert$ \\ \hline
$\delta\mathbf{x}$ & Residual & $\mathbf{x}-\mathbf{c}_{k}$ \\ \hline
$R^{2}$ & Radius$^2$ & $\mathbb{E}_{k}\!\big[\lVert\delta\mathbf{x}\rVert^{2}\big]\big/\lVert\mathbf{c}_{k}\rVert^{2}$ \\ \hline
$\sigma_{\mu}^{2}$ & Variance fraction & $\mathbb{E}_{k}\!\big[(\delta\mathbf{x}\cdot\hat{\mathbf c}_{\mu})^{2}\big]\big/\mathbb{E}_{k}\!\big[\lVert\delta\mathbf{x}\rVert^{2}\big]$ \\ \hline
$s_{\mu}$ & Std.\ projection & $\left(\delta\mathbf{x}\cdot\hat{\mathbf{c}}_{\mu}\right)/\mathrm{std}_{k}\left[\delta\mathbf{x}\cdot\hat{\mathbf{c}}_{\mu}\right]$ \\ \hline
$\rho_{\mu}$ & Proj.\ correlation & $\mathrm{Corr}_{k}\left(\delta\mathbf{x}\cdot\hat{\mathbf{c}}_{k},\delta\mathbf{x}\cdot\hat{\mathbf{c}}_{\mu}\right)$ \\ \hline
$g_{\mu}$ & Centroid overlap & $\hat{\mathbf c}_{k}\cdot\hat{\mathbf c}_{\mu}$ \\ \hline
$\mathcal{R}_{K}$ & Rival set & $\{\,\mu\neq k:\ K\ \text{largest}\ \sigma_{\mu}\,\}$ \\ \hline
\end{tabular}
\caption{Definitions. $\mathbb{E}_k$ is defined as average over all samples $\mathbf{x}$ belong to the manifold $k$}
\label{tab:defs}
\end{wraptable}
The theoretical accuracy predicted by Eq.\ref{eq:caligned-acc} alone consistently overestimates the accuracy on all the models we checked (see Appendix \ref{app sec: lambda}). Our main simplifying assumption (Assumption~5.4) is that the statistics of the standardized centroid projections $s_{\mu}$ are Gaussian.
Empirically, the projections have markedly heavy-tailed statistics (see Fig.\ref{fig:sparsity and heavy tail}c). Although the
variance of $s_\mu$ is normalized, classification depends on extreme statistics
($\max_{\mu}t_{\mu}$), and heavier tails inflate these
extremes relative to a Gaussian of equal variance. We find that the right tail of the variables $s_\mu$ that controls the max projection on a certain centroid is well fitted by a Student's-t power-law distribution $p(s)\propto (a^2+s^2{})^{-\nu/2}$ (see Fig.\ref{fig:sparsity and heavy tail}c), where the exponent $\nu$ shapes the decay of the tail.


We adopt the simplest solution: a uniform global variance inflation of all manifolds $R\mapsto\lambda R$ in Theorem~5.5, which adjusts the scale of the Gaussian theory such that its extreme matches the empirical heavy-tailed distribution. 

The global parameter $\lambda$ is fitted once per model
by
\vspace{-1mm}
\begin{equation}
\lambda^{\star}=\arg\min_{\lambda}\left(\sum_{k=1}^{P}\!\left[\mathrm{Acc}_{k}^{\mathrm{emp}}(R)-\mathrm{Acc}_{k}^{\mathrm{th}}(\lambda R)\right]^{2}\right),
\label{eq:lambda-fit}
\end{equation}

\vspace{-1mm}

\begin{figure}[hbt!]
\centering
\includegraphics[width=\linewidth]{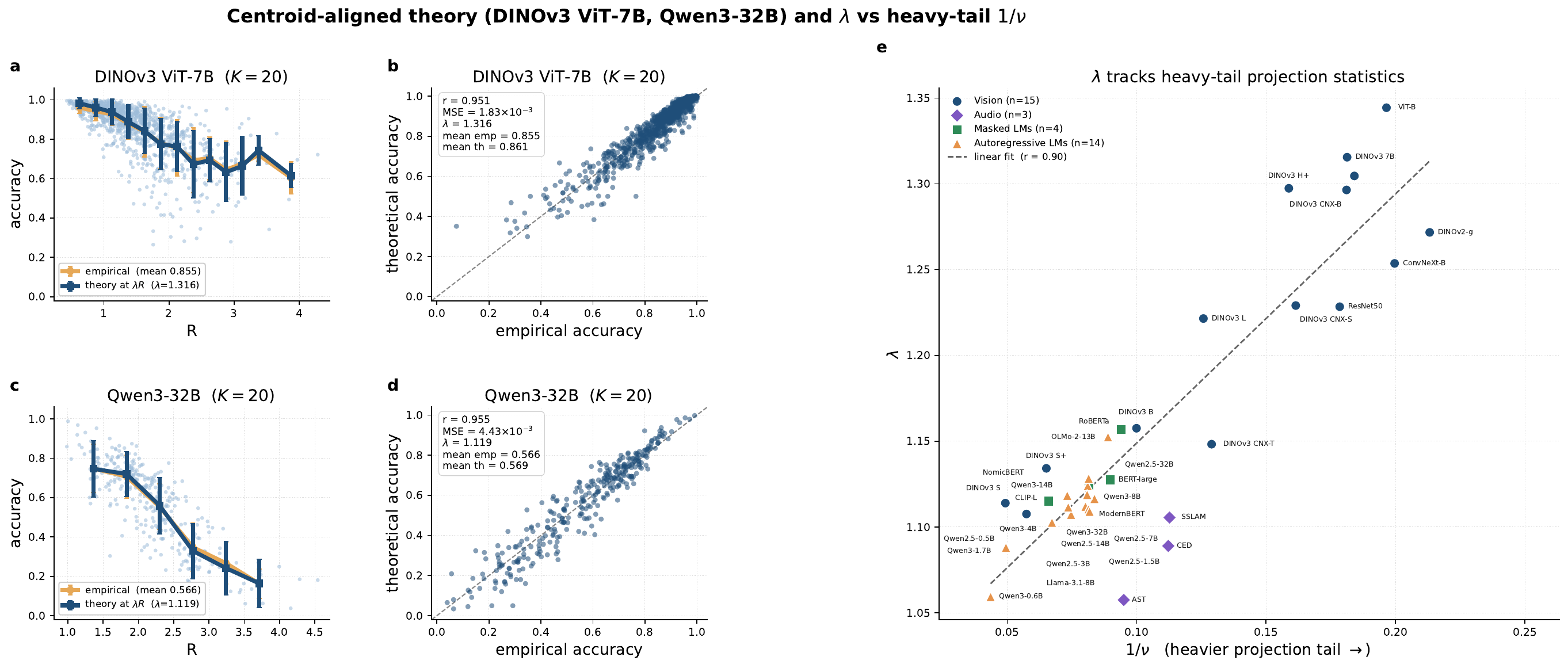}
\caption{\textbf{\label{fig:theory} Centroid-aligned theory matches
per-class accuracy across modalities.}
\textbf{(a, c)} Per-class accuracy vs.\ empirical radius $R$ for
DINOv3~ViT-7B on ImageNet ($P{=}1000$, $K{=}20$) and Qwen3-32B
on TinyStories next-token prediction ($P{=}320$, $K{=}20$). Per-class
scatter is presented with empirical binned mean. The $\lambda$-rescaled theory
($\lambda^{\star}{=}1.316$, $1.119$ resp.) accounts for the empirical accuracy accurately.
\textbf{(b, d)} Accurate per-class prediction of the accuracy. MSE${=}1.83\times10^{-3}$ (vision) and $4.43\times10^{-3}$ (language),
Pearson $r{=}0.951$ and $0.955$. 
\textbf{(e)} Optimal $\lambda^{\star}$ vs.\ heavy tail measurement (1/$\nu$)
on representations for 15 vision backbones, 14 autoregressive language models, 4 masked language models, and 3 audio models. Overall Pearson correlations between $\lambda$ and $1/\nu$ are $r{=}0.90$, confirming that $\lambda$ is a byproduct of the heavy-tail shape of the projection variables.}
\end{figure}

Thus $\lambda$ is a single global calibration of
non-Gaussian extreme-value statistics of a specific model, not a per-class fitting
parameter. Empirically $\lambda^{\star}$
is well correlated with the inverse tail exponent $1/\nu$ of the model, $r=0.90$ (see Fig.~\ref{fig:theory}e). This relation is global across 36 models spanning vision, audio, and language,
supporting our interpretation of $\lambda$ as a compensation for non-Gaussian statistics.

\subsection{Shared geometry across architectures and modalities}
\label{sec:universality}

The empirical structure of Sections \ref{section:empirical}-\ref{sec:theory} was characterized on
DINOv3 ViT-7B. Table~\ref{tab:universality} reports the same measurements across
a diverse set of vision, audio, and language backbones: Every observation persists: $R$ is in an $\mathcal{O}(1)$ range for any
$N$, with large accuracy fluctuations for fixed $R$ (independent of $N$). $\delta$-rotation drives the class-averaged accuracy to near-perfect
on all backbones; the saturation accuracy ($R^\star\rightarrow\infty$) is consistently orders of magnitude
above chance, indicating true-class dominance among the residual correlations. $K_{90\%}$ is the number of rivals that are the strongest competitor for 90\% of a class's examples. This rival set is sparse,
$K_{90\%}\!\ll\!P$ in all cases. The projection $s_\mu$ right tail is heavy on every
backbone. The geometric characterizations required by the theory of
Section~\ref{sec:theory} are therefore present and consistent across different architectures, 
modalities, and training procedures. Appendix~\ref{app:manymodels} and Figs.~\ref{app:grid-vision}-\ref{app:grid-audio} report the theory-empirical comparison for all $36$ backbones.

\begin{table}
\caption{Universality of the centroid-aligned geometry. For each backbone, we report: the number of
classes $P$; feature dimension $N$; per-class radius range; per-class mean accuracy
$\langle\mathrm{Acc}\rangle$ and its fluctuation around the mean accuracy at a fixed $R$;
mean accuracy after $\delta$-rotation; saturation accuracy at $R^{\star}\rightarrow\infty$ (controlled only by the residual $\delta{\bf x}$); $K_{90\%}$ (mean rivals covering $90\%$
of examples); right-tail Student-$t$ tail index $\nu$ ($K{=}20$).}

\label{tab:universality}\centering\small\setlength{\tabcolsep}{4pt}
\begin{tabular}{lccccccccc}
\toprule
Backbone & $P$ & $N$ & $R$ range & $\langle\mathrm{Acc}\rangle$ & \shortstack{Fixed $R$\\ fluctuations} &
$\delta$-rot & $R^{\star}\rightarrow\infty$ & $\langle K_{90\%}\rangle$ & $\nu$ \\
\midrule
\multicolumn{10}{l}{\textit{Vision -- ImageNet-1K }} \\
DINOv3 ViT-7B/16     & 1000 & 4096 & [0.4, 4.3] & 0.855 & 0.099 & 1.000 & 0.392 & 14.4 & 4.5 \\
DINOv3 ConvNeXt-L    & 1000 & 1536 & [0.2, 3.1] & 0.814 & 0.100 & 1.000 & 0.312 & 14.2 & 4.4 \\
CLIP ViT-L/14        & 1000 &  768 & [0.6, 2.7] & 0.792 & 0.124 & 0.999 & 0.077 & 13.9 & 16.4 \\
ViT-B/16 (IN21K)     & 1000 &  768 & [0.5, 2.7] & 0.853 & 0.088 & 1.000 & 0.265 & 14.1 & 4.1 \\[2pt]
\multicolumn{10}{l}{\textit{Audio -- VGGSound}} \\
SSLAM                &  310 &  768 & [0.1, 3.7] & 0.628 & 0.155 & 0.999 & 0.067 & 11.0 & 7.9 \\
CED             &  310 &  768 & [0.1, 4.0] & 0.628 & 0.160 & 1.000 & 0.062 & 11.3 & 7.9 \\[2pt]
\multicolumn{10}{l}{\textit{Language - TinyStories masked-token }} \\
RoBERTa-large        &  320 & 1024 & [0.6, 2.6] & 0.767 & 0.108 & 1.000 & 0.120 & 13.3 & 9.6 \\
ModernBERT-large     &  320 & 1024 & [0.6, 2.7] & 0.734 & 0.112 & 1.000 & 0.118 & 12.4 & 11.3 \\
BERT-large           &  320 & 1024 & [0.6, 2.7] & 0.710 & 0.112 & 1.000 & 0.120 & 15.8 & 10.1 \\[2pt]
\multicolumn{10}{l}{\textit{Language - TinyStories next-token }} \\
Qwen3-8B (L34, $-2$) &  320 & 4096 & [0.9, 4.2] & 0.540 & 0.118 & 1.000 & 0.107 & 33.7 & 11.3 \\
OLMo2-13B (L36, $-4$) &  320 & 5120 & [1.0, 4.8] & 0.563 & 0.116 & 1.000 & 0.112 & 35.8 & 10.2 \\
Llama-3.1-8B (L29, $-3$) &  320 & 4096 & [0.8, 4.1] & 0.550 & 0.119 & 1.000 & 0.101 & 31.7 & 12.4 \\
\bottomrule
\end{tabular}
\end{table}

\vspace{-2mm}

\section{Geometry and Accuracy Across Model Scales}
\label{sec:scaling}

The theory of Sec.~\ref{sec:theory} reduces accuracy to a handful of ingredients of the centroid-aligned geometry. We use these quantities as a diagnostic lens across the DINOv3 and  Qwen3 model families at different scales ($S(21M)/S_+(29M)/B(86M)/L(300M)/H_+(840M)/7B$, $0.6B/1.7B/4B/8B/14B/32B$ resp.). The theory reveals which geometric reorganizations drive the improvements in classification performance of larger models.

\label{sec:scaling:scale}

A natural concern is that the accuracy gains of larger models are a trivial variability-reduction effect — that bigger models simply produce tighter manifolds. The Qwen3 family rules this out directly: the mean radius actually grows with scale ($\langle R\rangle = 1.98 \to 2.20$ from 0.6B to 32B), yet accuracy rises monotonically from $41\%$ to $57\%$, the opposite of what a naive approach predicts. In the DINOv3 family, the radius does shrink with scale, but the trend breaks at the largest model: ViT-7B carries a mean radius $0.21$ larger than ViT-H$^+$ yet matches its accuracy. Reduced variability, therefore, cannot be the mechanism behind improved generalization. Our other geometric measures are independent of $R$, so they isolate genuine reorganizations of the geometry across scale, and correctly identify the components that make the larger models' representations better.

We specifically look at the worst rival $\mu^*$, measured by the maximum $\sigma_\mu$ (most correlated centroid among the variability) for each class, and average its geometric statistics over all classes. The theory exposes a coherent set of reorganizations, shared among the two families. 
Within-class variability becomes more aligned with the true centroid and less aligned with the rival (the ratio $\sigma_k/\sigma_\mu^*$ increases, $\sigma_\mu^\ast$ decreases in both families). The centroids decorrelate - the overlap between the true-centroid and its hardest rival $g^{*}_{\mu}$ drops - and the projections of the residual on the centroids also become less correlated ($\rho^{*}$ decreases). In language, we see similar trends to vision but with less favorable values, which explains the overall lower accuracy. Finally, representations become more heavy-tailed with scale ($\nu$ decreases, $19.3\mapsto4.5$ in vision and $21.9\mapsto11.4$ in language) . While we cannot directly link this trend to improved accuracy, it is consistent.

A large body of literature \citep{peters2018dissecting, liu2019linguistic, tenney2019bert, skean2025layer} reports that in many cases, the best representations for tasks are found in the middle hidden layers, and not in the final layer. We add our geometric angle to it in finding that in autoregressive models, the best prototype classification is consistently 1-3 layers before the final layer. We report here the geometry for the best layer found for prototype classification, and analyze how the geometry changes across depth in Appendix \ref{app sec: qwen}.

\begin{table}
\centering
\small
\setlength{\tabcolsep}{4pt}
\caption{\textbf{Geometric measures across model scales:} DINOv3 ViT family on ImageNet-1K (top block) and the
Qwen3 language-model family on TinyStories (bottom block).
The geometric quantities are defined in Sec.~\ref{sec:theory} and Table~\ref{tab:defs}; the
superscript $^{*}$ flags the hardest competitor of a class, identified by $\sigma_{\mu}$.
Best measure within each family marked in bold.}

\label{tab:scale_depth}
\begin{tabular}{l|c|cccc|c|c}
\hline
Model & $\langle R\rangle$
      & $\sigma_k/\sigma^{*}_{\mu}$ & $\sigma^{*}_{\mu}$ & $g^{*}_{\mu}$ & $\rho_\mu^{*}$
      & $\nu$
      & Acc \\
\hline
DINOv3 ViT-S           & 1.68 & 1.06 & 0.172 & 0.574 & 0.558 & 19.3 & 0.706 \\
DINOv3 ViT-S$+$        & 1.60 & 1.09 & 0.172 & 0.555 & 0.535 & 14.4 & 0.738 \\
DINOv3 ViT-B           & 1.52 & 1.19 & 0.163 & 0.542 & 0.529 & 9.0 & 0.794 \\
DINOv3 ViT-L           & 1.39 & 1.36 & 0.147 & 0.447 & 0.322 & 7.0 & 0.845 \\
DINOv3 ViT-H$+$        & \textbf{1.29} & 1.59 & 0.164 & \textbf{0.236} & \textbf{-0.011} & 5.3 & \textbf{0.855} \\
DINOv3 ViT-7B          & 1.50 & \textbf{2.01} & \textbf{0.120} & 0.289 & 0.038 & \textbf{4.5} & \textbf{0.855} \\
\hline
Qwen3-0.6B (L27, $-1$) & \textbf{1.98} & 0.85 & 0.241 & 0.576 & 0.515 & 21.9 & 0.406 \\
Qwen3-1.7B (L26, $-2$) & 2.09 & 0.95 & 0.212 & 0.542 & 0.494 & 13.9 & 0.474 \\
Qwen3-4B (L34, $-2$)   & 2.05 & 0.98 & 0.202 & 0.526 & 0.463 & 12.7 & 0.519 \\
Qwen3-8B (L34, $-2$)   & 2.12 & 0.99 & 0.188 & 0.514 & 0.453 & \textbf{11.3} & 0.540 \\
Qwen3-14B (L37, $-3$)  & 2.10 & 1.00 & 0.186 & 0.504 & \textbf{0.447} & \textbf{11.3} & 0.556 \\
Qwen3-32B (L61, $-3$)  & 2.20 & \textbf{1.01} & \textbf{0.175} & \textbf{0.503} & 0.451 & 11.4 & \textbf{0.566} \\
\hline
\end{tabular}
\end{table}

\vspace{-5mm}
\section{Discussion}
\label{sec:discussion}

\vspace{-3mm}

This work identifies geometric organization principles for multiclass representations: prototype accuracy is controlled by a sparse, centroid-aligned code embedded in the full activation space. Across vision, audio, and language models, and across architectures and training objectives, the same geometry recurs. Within-class variability is not isotropic around each class center; its classifier-relevant components are arranged along the true centroid and a small set of rival centroids. The evidence is causal as well as descriptive: breaking this alignment, while preserving the remaining parts of the geometry, nearly eliminates prototype errors. Thus the full $N$-dimensional representation can be replaced, for the purpose of predicting class accuracy, by a much smaller set of task-aligned coordinates.

The theory makes this reduction quantitative. Per-class accuracy is predicted from the geometric measures of centroid-projection statistics, and a single global scale $\lambda$ that captures the effect of heavy tails in the marginal projection distributions. Radius alone is insufficient to explain the results. We show that as models become larger, the improved accuracy is not due to reduced class variance
but because of reorganization of the geometry: variability shifts toward the true-class coordinate, correlations with the strongest rivals decrease, and centroid directions become more decorrelated.

Importantly, the theory tracks not only prototype classification but also standard linear classification (Appendix \ref{app sec: prototype linear comparison}). For linear readouts, a train–test split is essential, since a readout can fit arbitrary labels whenever the data remain separable. Prototype classification, by contrast, achieves high accuracy only for genuinely well-organized geometry — shuffled labels or random representations perform near chance — so the train–test split (calculating the centroids on different data than the one classified) contributes only finite-sample effects, negligible for large sample sizes when train and test are drawn from the same distribution.

This perspective complements existing theories of representational geometry. Manifold-capacity theory~\citep{chung2018classification, cohen2020separability}
models class manifolds as ellipsoids and characterizes their linear
separability under random binary dichotomies. In that setting, the primary component of the geometry is the random structure of the whole $P$ manifolds, and correlations enter as
corrections on top of the random theory \citep{wakhloo2023linear}. We showed that in our task the centroid-aligned component of
within-class variability is the dominant part, while the
random fluctuations of inter-centroid overlaps have little to no effect. The two frameworks are complementary
regimes of the same underlying geometry, and it would be interesting to further investigate where the two descriptions meet.

The sparse code uncovered here also suggests links to interpretability and neuroscience. Sparse autoencoders and dictionary-learning methods seek low-dimensional, semantically meaningful directions in activations~\citep{makhzani2013k,huben2024sparse,templeton2026scaling,gao2025scaling}; our results identify a complementary sparse class-supervised set of directions whose functional role is directly tied to classification. In parallel, deep networks have been shown to predict responses in primate ventral-stream areas, including IT cortex~\citep{yamins2014performance,schrimpf2020integrative}, and object-manifold theory has been applied to biological populations~\citep{froudarakis2020object,sorscher2022neural, kuoch2024probing}. It would be interesting to examine whether the universal geometry we identify across many artificial models also applies to representations in the brain.

Finally, the theory points to practical applications. Because its ingredients are compact and inexpensive to estimate, they can serve as diagnostics for representation quality, to select which pretrained model is most suitable for fine-tuning on a new task \citep{kornblith2018better}. They could also suggest geometry-aware objectives: training could explicitly encourage strong own-centroid alignment, weak rival alignment, and sparse competition, while distillation could match the teacher and student geometry rather than the entire $N$-dimensional representation or the logit distribution~\citep{hinton2015distilling}. Following these directions would turn the centroid-aligned description from a predictive theory of existing representations into a constructive principle for designing better ones.

\textbf{Acknowledgments:} We thank David G. Clark, Yoni Ankri, and Nadav Lederman for fruitful discussions. This research is supported by the Gatsby Charitable Foundation, the Kempner Institute for the Study of Natural and Artificial Intelligence
at Harvard University, the Office of
Naval Research grant No. N0014-23-1-2051, and the Institute of Information and Communications Technology Planning and Evaluation (IITP), grant funded by the Korean government (MSIT) (No. RS-2024-00457882, National AI Research Lab Project).

\bibliographystyle{plainnat}
\bibliography{bib}

@article{honnibal2020spacy,
author = {Honnibal, Matthew and Montani, Ines and Van Landeghem, Sofie and Boyd, Adriane},
doi = {10.5281/zenodo.1212303},
title = {{spaCy: Industrial-strength Natural Language Processing in Python}},
year = {2020}
}

@article{galanti2021role,
  title={On the role of neural collapse in transfer learning},
  author={Galanti, Tomer and Gy{\"o}rgy, Andr{\'a}s and Hutter, Marcus},
  journal={arXiv preprint arXiv:2112.15121},
  year={2021}
}

@article{luthra2026directional,
  title={Directional Neural Collapse Explains Few-Shot Transfer in Self-Supervised Learning},
  author={Luthra, Achleshwar and Salunkhe, Yash and Galanti, Tomer},
  journal={arXiv preprint arXiv:2603.03530},
  year={2026}
}

@article{bricken2023monosemanticity,
       title={Towards Monosemanticity: Decomposing Language Models With Dictionary Learning},
       author={Bricken, Trenton and Templeton, Adly and Batson, Joshua and Chen, Brian and Jermyn, Adam and Conerly, Tom and Turner, Nick and Anil, Cem and Denison, Carson and Askell, Amanda and Lasenby, Robert and Wu, Yifan and Kravec, Shauna and Schiefer, Nicholas and Maxwell, Tim and Joseph, Nicholas and Hatfield-Dodds, Zac and Tamkin, Alex and Nguyen, Karina and McLean, Brayden and Burke, Josiah E and Hume, Tristan and Carter, Shan and Henighan, Tom and Olah, Christopher},
       year={2023},
       journal={Transformer Circuits Thread},
       note={https://transformer-circuits.pub/2023/monosemantic-features/index.html}
    }

@article{bommasani2021opportunities,
  title={On the opportunities and risks of foundation models},
  author={Bommasani, Rishi and Hudson, Drew A and Adeli, Ehsan and Altman, Russ and Arora, Simran and von Arx, Sydney and Bernstein, Michael S and Bohg, Jeannette and Bosselut, Antoine and Brunskill, Emma and others},
  journal={arXiv preprint arXiv:2108.07258},
  year={2021}
}

@article{eldan2023tinystories,
  title={Tinystories: How small can language models be and still speak coherent english?},
  author={Eldan, Ronen and Li, Yuanzhi},
  journal={arXiv preprint arXiv:2305.07759},
  year={2023}
}

@inproceedings{arps2024multilingual,
  title={Multilingual nonce dependency treebanks: Understanding how language models represent and process syntactic structure},
  author={Arps, David and Kallmeyer, Laura and Samih, Younes and Sajjad, Hassan},
  booktitle={Proceedings of the 2024 Conference of the North American Chapter of the Association for Computational Linguistics: Human Language Technologies (Volume 1: Long Papers)},
  pages={7822--7844},
  year={2024}
}

@article{de2021universal,
  title={Universal dependencies},
  author={De Marneffe, Marie-Catherine and Manning, Christopher D and Nivre, Joakim and Zeman, Daniel},
  journal={Computational linguistics},
  volume={47},
  number={2},
  pages={255--308},
  year={2021},
  publisher={MIT Press One Rogers Street, Cambridge, MA 02142-1209, USA journals-info~…}
}

@inproceedings{peters2018dissecting,
  title={Dissecting contextual word embeddings: Architecture and representation},
  author={Peters, Matthew E and Neumann, Mark and Zettlemoyer, Luke and Yih, Wen-tau},
  booktitle={Proceedings of the 2018 conference on empirical methods in natural language processing},
  pages={1499--1509},
  year={2018}
}

@inproceedings{liu2019linguistic,
  title={Linguistic knowledge and transferability of contextual representations},
  author={Liu, Nelson F and Gardner, Matt and Belinkov, Yonatan and Peters, Matthew E and Smith, Noah A},
  booktitle={Proceedings of the 2019 conference of the North American chapter of the association for computational linguistics: human language technologies, volume 1 (long and short papers)},
  pages={1073--1094},
  year={2019}
}

@inproceedings{tenney2019bert,
  title={BERT rediscovers the classical NLP pipeline},
  author={Tenney, Ian and Das, Dipanjan and Pavlick, Ellie},
  booktitle={Proceedings of the 57th annual meeting of the association for computational linguistics},
  pages={4593--4601},
  year={2019}
}

@article{skean2025layer,
  title={Layer by layer: Uncovering hidden representations in language models},
  author={Skean, Oscar and Arefin, Md Rifat and Zhao, Dan and Patel, Niket and Naghiyev, Jalal and LeCun, Yann and Shwartz-Ziv, Ravid},
  journal={arXiv preprint arXiv:2502.02013},
  year={2025}
}

@inproceedings{kuoch2024probing,
  title={Probing biological and artificial neural networks with task-dependent neural manifolds},
  author={Kuoch, Michael and Chou, Chi-Ning and Parthasarathy, Nikhil and Dapello, Joel and DiCarlo, James J and Sompolinsky, Haim and Chung, SueYeon},
  booktitle={Conference on Parsimony and Learning},
  pages={395--418},
  year={2024},
  organization={PMLR}
}

@article{froudarakis2020object,
  title={Object manifold geometry across the mouse cortical visual hierarchy},
  author={Froudarakis, Emmanouil and Cohen, Uri and Diamantaki, Maria and Patel, Saumil and Tan, Zheng and Muhammad, Taliah and Walker, Edgar Y and Reimer, Jacob and Berens, Philipp and Sompolinsky, Haim and others},
  journal={BioRxiv},
  pages={2020--08},
  year={2020},
  publisher={Cold Spring Harbor Laboratory}
}

@article{hendrycks2021unsolved,
  title={Unsolved problems in ml safety},
  author={Hendrycks, Dan and Carlini, Nicholas and Schulman, John and Steinhardt, Jacob},
  journal={arXiv preprint arXiv:2109.13916},
  year={2021}
}

@article{elhage2022toy,
  title={Toy models of superposition},
  author={Elhage, Nelson and Hume, Tristan and Olsson, Catherine and Schiefer, Nicholas and Henighan, Tom and Kravec, Shauna and Hatfield-Dodds, Zac and Lasenby, Robert and Drain, Dawn and Chen, Carol and others},
  journal={arXiv preprint arXiv:2209.10652},
  year={2022}
}

@inproceedings{radford2021learning,
  title={Learning transferable visual models from natural language supervision},
  author={Radford, Alec and Kim, Jong Wook and Hallacy, Chris and Ramesh, Aditya and Goh, Gabriel and Agarwal, Sandhini and Sastry, Girish and Askell, Amanda and Mishkin, Pamela and Clark, Jack and others},
  booktitle={International conference on machine learning},
  pages={8748--8763},
  year={2021},
  organization={PmLR}
}

@article{dosovitskiy2020image,
  title={An image is worth 16x16 words: Transformers for image recognition at scale},
  author={Dosovitskiy, Alexey and Beyer, Lucas and Kolesnikov, Alexander and Weissenborn, Dirk and Zhai, Xiaohua and Unterthiner, Thomas and Dehghani, Mostafa and Minderer, Matthias and Heigold, Georg and Gelly, Sylvain and others},
  journal={arXiv preprint arXiv:2010.11929},
  year={2020}
}

@inproceedings{liu2022convnet,
  title={A convnet for the 2020s},
  author={Liu, Zhuang and Mao, Hanzi and Wu, Chao-Yuan and Feichtenhofer, Christoph and Darrell, Trevor and Xie, Saining},
  booktitle={2022 IEEE/CVF conference on computer vision and pattern recognition (CVPR)},
  pages={11966--11976},
  year={2022},
  organization={IEEE}
}

@inproceedings{he2016deep,
  title={Deep residual learning for image recognition},
  author={He, Kaiming and Zhang, Xiangyu and Ren, Shaoqing and Sun, Jian},
  booktitle={Proceedings of the IEEE conference on computer vision and pattern recognition},
  pages={770--778},
  year={2016}
}

@inproceedings{huben2024sparse,
  title={Sparse autoencoders find highly interpretable features in language models},
  author={Huben, Robert and Cunningham, Hoagy and Smith, Logan and Ewart, Aidan and Sharkey, Lee},
  booktitle={International Conference on Learning Representations},
  volume={2024},
  pages={7827--7845},
  year={2024}
}

@article{templeton2026scaling,
  title={Scaling monosemanticity: Extracting interpretable features from claude 3 sonnet},
  author={Templeton, Adly and Conerly, Tom and Marcus, Jonathan and Lindsey, Jack and Bricken, Trenton and Chen, Brian and Pearce, Adam and Citro, Craig and Ameisen, Emmanuel and Jones, Andy and others},
  journal={arXiv preprint arXiv:2605.29358},
  year={2026}
}

@inproceedings{gao2025scaling,
  title={Scaling and evaluating sparse autoencoders},
  author={Gao, Leo and Dupre la Tour, Tom and Tillman, Henk and Goh, Gabriel and Troll, Rajan and Radford, Alec and Sutskever, Ilya and Leike, Jan and Wu, Jeffrey},
  booktitle={International Conference on Learning Representations},
  volume={2025},
  pages={26721--26754},
  year={2025}
}

@article{makhzani2013k,
  title={K-sparse autoencoders},
  author={Makhzani, Alireza and Frey, Brendan},
  journal={arXiv preprint arXiv:1312.5663},
  year={2013}
}

@article{lachenbruch1968estimation,
  title={Estimation of error rates in discriminant analysis},
  author={Lachenbruch, Peter A and Mickey, M Ray},
  journal={Technometrics},
  volume={10},
  number={1},
  pages={1--11},
  year={1968},
  publisher={Taylor \& Francis}
}

@article{chung2018classification,
  title={Classification and geometry of general perceptual manifolds},
  author={Chung, SueYeon and Lee, Daniel D and Sompolinsky, Haim},
  journal={Physical Review X},
  volume={8},
  number={3},
  pages={031003},
  year={2018},
  publisher={APS}
}

@article{cohen2020separability,
  title={Separability and geometry of object manifolds in deep neural networks},
  author={Cohen, Uri and Chung, SueYeon and Lee, Daniel D and Sompolinsky, Haim},
  journal={Nature communications},
  volume={11},
  number={1},
  pages={746},
  year={2020},
  publisher={Nature Publishing Group UK London}
}

@article{wakhloo2023linear,
  title={Linear classification of neural manifolds with correlated variability},
  author={Wakhloo, Albert J and Sussman, Tamara J and Chung, SueYeon},
  journal={Physical Review Letters},
  volume={131},
  number={2},
  pages={027301},
  year={2023},
  publisher={APS}
}

@article{papyan2020prevalence,
  title={Prevalence of neural collapse during the terminal phase of deep learning training},
  author={Papyan, Vardan and Han, Xiao Y and Donoho, David L},
  journal={Proceedings of the National Academy of Sciences},
  volume={117},
  number={40},
  pages={24652--24663},
  year={2020},
  publisher={National Academy of Sciences}
}

@article{han2021neural,
  title={Neural collapse under mse loss: Proximity to and dynamics on the central path},
  author={Han, XY and Papyan, Vardan and Donoho, David L},
  journal={arXiv preprint arXiv:2106.02073},
  year={2021}
}

@article{oquab2023dinov2,
  title={Dinov2: Learning robust visual features without supervision},
  author={Oquab, Maxime and Darcet, Timoth{\'e}e and Moutakanni, Th{\'e}o and Vo, Huy and Szafraniec, Marc and Khalidov, Vasil and Fernandez, Pierre and Haziza, Daniel and Massa, Francisco and El-Nouby, Alaaeldin and others},
  journal={arXiv preprint arXiv:2304.07193},
  year={2023}
}

@article{simeoni2025dinov3,
  title={Dinov3},
  author={Sim{\'e}oni, Oriane and Vo, Huy V and Seitzer, Maximilian and Baldassarre, Federico and Oquab, Maxime and Jose, Cijo and Khalidov, Vasil and Szafraniec, Marc and Yi, Seungeun and Ramamonjisoa, Micha{\"e}l and others},
  journal={arXiv preprint arXiv:2508.10104},
  year={2025}
}

@inproceedings{devlin2019bert,
  title={Bert: Pre-training of deep bidirectional transformers for language understanding},
  author={Devlin, Jacob and Chang, Ming-Wei and Lee, Kenton and Toutanova, Kristina},
  booktitle={Proceedings of the 2019 conference of the North American chapter of the association for computational linguistics: human language technologies, volume 1 (long and short papers)},
  pages={4171--4186},
  year={2019}
}

@article{mensink2013distance,
  title={Distance-based image classification: Generalizing to new classes at near-zero cost},
  author={Mensink, Thomas and Verbeek, Jakob and Perronnin, Florent and Csurka, Gabriela},
  journal={IEEE transactions on pattern analysis and machine intelligence},
  volume={35},
  number={11},
  pages={2624--2637},
  year={2013},
  publisher={IEEE}
}

@article{snell2017prototypical,
  title={Prototypical networks for few-shot learning},
  author={Snell, Jake and Swersky, Kevin and Zemel, Richard},
  journal={Advances in neural information processing systems},
  volume={30},
  year={2017}
}

@inproceedings{deng2009imagenet,
  title={Imagenet: A large-scale hierarchical image database},
  author={Deng, Jia and Dong, Wei and Socher, Richard and Li, Li-Jia and Li, Kai and Fei-Fei, Li},
  booktitle={2009 IEEE conference on computer vision and pattern recognition},
  pages={248--255},
  year={2009},
  organization={Ieee}
}

@article{hinton2015distilling,
  title={Distilling the knowledge in a neural network},
  author={Hinton, Geoffrey and Vinyals, Oriol and Dean, Jeff},
  journal={arXiv preprint arXiv:1503.02531},
  year={2015}
}

@article{kornblith2018better,
  title={Do better imagenet models transfer better?},
  author={Kornblith, Simon and Shlens, Jonathon and Le, Quoc V},
  journal={arXiv preprint arXiv:1805.08974},
  year={2018}
}

@article{yamins2014performance,
  title={Performance-optimized hierarchical models predict neural responses in higher visual cortex},
  author={Yamins, Daniel LK and Hong, Ha and Cadieu, Charles F and Solomon, Ethan A and Seibert, Darren and DiCarlo, James J},
  journal={Proceedings of the national academy of sciences},
  volume={111},
  number={23},
  pages={8619--8624},
  year={2014},
  publisher={National Academy of Sciences}
}

@article{schrimpf2020integrative,
  title={Integrative benchmarking to advance neurally mechanistic models of human intelligence},
  author={Schrimpf, Martin and Kubilius, Jonas and Lee, Michael J and Murty, N Apurva Ratan and Ajemian, Robert and DiCarlo, James J},
  journal={Neuron},
  volume={108},
  number={3},
  pages={413--423},
  year={2020},
  publisher={Elsevier}
}

@article{sorscher2022neural,
  title={Neural representational geometry underlies few-shot concept learning},
  author={Sorscher, Ben and Ganguli, Surya and Sompolinsky, Haim},
  journal={Proceedings of the National Academy of Sciences},
  volume={119},
  number={43},
  pages={e2200800119},
  year={2022},
  publisher={National Academy of Sciences}
}

@article{liu2019roberta,
  title={Roberta: A robustly optimized bert pretraining approach},
  author={Liu, Yinhan and Ott, Myle and Goyal, Naman and Du, Jingfei and Joshi, Mandar and Chen, Danqi and Levy, Omer and Lewis, Mike and Zettlemoyer, Luke and Stoyanov, Veselin},
  journal={arXiv preprint arXiv:1907.11692},
  year={2019}
}

@inproceedings{chen2020vggsound,
  title={Vggsound: A large-scale audio-visual dataset},
  author={Chen, Honglie and Xie, Weidi and Vedaldi, Andrea and Zisserman, Andrew},
  booktitle={ICASSP 2020-2020 IEEE International Conference on Acoustics, Speech and Signal Processing (ICASSP)},
  pages={721--725},
  year={2020},
  organization={IEEE}
}

@article{gong2021ast,
  title={Ast: Audio spectrogram transformer},
  author={Gong, Yuan and Chung, Yu-An and Glass, James},
  journal={arXiv preprint arXiv:2104.01778},
  year={2021}
}

@article{stephenson2021geometry,
  title={On the geometry of generalization and memorization in deep neural networks},
  author={Stephenson, Cory and Padhy, Suchismita and Ganesh, Abhinav and Hui, Yue and Tang, Hanlin and Chung, SueYeon},
  journal={arXiv preprint arXiv:2105.14602},
  year={2021}
}

@inproceedings{warner2025smarter,
  title={Smarter, better, faster, longer: A modern bidirectional encoder for fast, memory efficient, and long context finetuning and inference},
  author={Warner, Benjamin and Chaffin, Antoine and Clavi{\'e}, Benjamin and Weller, Orion and Hallstr{\"o}m, Oskar and Taghadouini, Said and Gallagher, Alexis and Biswas, Raja and Ladhak, Faisal and Aarsen, Tom and others},
  booktitle={Proceedings of the 63rd annual meeting of the association for computational linguistics (volume 1: Long papers)},
  pages={2526--2547},
  year={2025}
}

@article{nussbaum2024nomic,
  title={Nomic embed: Training a reproducible long context text embedder},
  author={Nussbaum, Zach and Morris, John X and Duderstadt, Brandon and Mulyar, Andriy},
  journal={arXiv preprint arXiv:2402.01613},
  year={2024}
}

@misc{qwen2025qwen25technicalreport,
      title={Qwen2.5 Technical Report}, 
      author={Qwen and : and An Yang and Baosong Yang and Beichen Zhang and Binyuan Hui and Bo Zheng and Bowen Yu and Chengyuan Li and Dayiheng Liu and Fei Huang and Haoran Wei and Huan Lin and Jian Yang and Jianhong Tu and Jianwei Zhang and Jianxin Yang and Jiaxi Yang and Jingren Zhou and Junyang Lin and Kai Dang and Keming Lu and Keqin Bao and Kexin Yang and Le Yu and Mei Li and Mingfeng Xue and Pei Zhang and Qin Zhu and Rui Men and Runji Lin and Tianhao Li and Tianyi Tang and Tingyu Xia and Xingzhang Ren and Xuancheng Ren and Yang Fan and Yang Su and Yichang Zhang and Yu Wan and Yuqiong Liu and Zeyu Cui and Zhenru Zhang and Zihan Qiu},
      year={2025},
      eprint={2412.15115},
      archivePrefix={arXiv},
      primaryClass={cs.CL},
      url={https://arxiv.org/abs/2412.15115}, 
}

@article{yang2025qwen3,
  title={Qwen3 technical report},
  author={Yang, An and Li, Anfeng and Yang, Baosong and Zhang, Beichen and Hui, Binyuan and Zheng, Bo and Yu, Bowen and Gao, Chang and Huang, Chengen and Lv, Chenxu and others},
  journal={arXiv preprint arXiv:2505.09388},
  year={2025}
}

@article{grattafiori2024llama,
  title={The llama 3 herd of models},
  author={Grattafiori, Aaron and Dubey, Abhimanyu and Jauhri, Abhinav and Pandey, Abhinav and Kadian, Abhishek and Al-Dahle, Ahmad and Letman, Aiesha and Mathur, Akhil and Schelten, Alan and Vaughan, Alex and others},
  journal={arXiv preprint arXiv:2407.21783},
  year={2024}
}

@article{crammer2001algorithmic,
  title={On the algorithmic implementation of multiclass kernel-based vector machines},
  author={Crammer, Koby and Singer, Yoram},
  journal={Journal of machine learning research},
  volume={2},
  number={Dec},
  pages={265--292},
  year={2001}
}

@article{olmo20242,
  title={2 OLMo 2 Furious},
  author={OLMo, Team and Walsh, Pete and Soldaini, Luca and Groeneveld, Dirk and Lo, Kyle and Arora, Shane and Bhagia, Akshita and Gu, Yuling and Huang, Shengyi and Jordan, Matt and others},
  journal={arXiv preprint arXiv:2501.00656},
  year={2024}
}

@inproceedings{dinkel2024ced,
  title={CED: Consistent ensemble distillation for audio tagging},
  author={Dinkel, Heinrich and Wang, Yongqing and Yan, Zhiyong and Zhang, Junbo and Wang, Yujun},
  booktitle={ICASSP 2024-2024 IEEE International Conference on Acoustics, Speech and Signal Processing (ICASSP)},
  pages={291--295},
  year={2024},
  organization={IEEE}
}

@inproceedings{alex2025sslam,
  title={Sslam: Enhancing self-supervised models with audio mixtures for polyphonic soundscapes},
  author={Alex, Tony and Atito, Sara and Mustafa, Armin and Awais, Muhammad and Jackson, Philip},
  booktitle={International Conference on Learning Representations},
  volume={2025},
  pages={22608--22626},
  year={2025}
}
\newpage
\appendix

\section{Prototype Classification}

\subsection{Prototype classification and linear readout comparison}\label{app sec: prototype linear comparison}

\begin{figure}[H]
\centering
\includegraphics[width=0.85\linewidth]{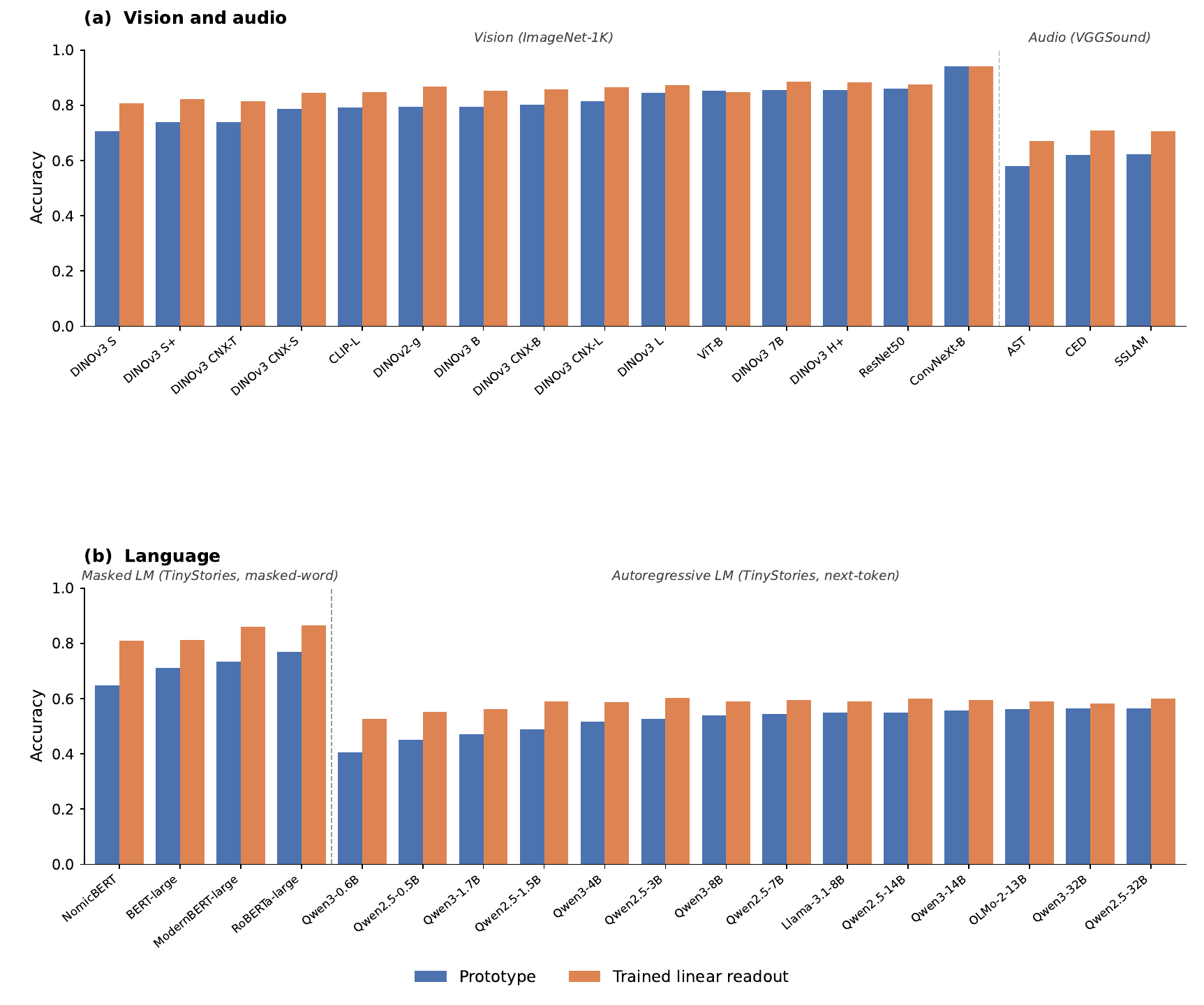}
\caption{Centroid vs trained linear-readout accuracy, per
backbone. The gap between methods is small (see also Table \ref{tab:proto-vs-readout-pearson}).}
\label{fig:proto-vs-linear-models}
\end{figure}

One natural concern regarding our work is that it is applied only to prototype classification, with no relation to the more common linear readout. We test this by splitting the data into $90\%$ train and $10\%$ validation, training a max-margin SVM (using the Crammer–Singer method \citep{crammer2001algorithmic}) on the train data, and evaluating on the validation set. We compare it with computing the centroids over the $90\%$ train representations, and computing the prototype scores on the $10\%$. The gap between the two methods is small (see Fig.\ref{fig:proto-vs-linear-models} and Table \ref{tab:proto-vs-readout-pearson}), especially in vision (0.047) and autoregressive models (0.063). We note that in the bidirectional language models, the gap is substantially larger than in the other modalities (0.121), which is also reflected in a lower per-class correlation between the two classifiers ($r=0.58$, against $0.87$ in vision and $0.76$ in the autoregressive models). The additional structure that allows the linear readout to separate the data better than the centroids in these models remains an open question.

\begin{table}[t]
\centering
\small
\setlength{\tabcolsep}{5pt}
\renewcommand{\arraystretch}{1.05}
\begin{tabular}[t]{lcc}
\toprule
\textbf{Backbone} & \textbf{$r$} & $\Delta$Acc \\
\midrule
\multicolumn{3}{l}{\emph{Vision (ImageNet-1K)}} \\
\quad ViT-B/16 (IN21K)             & $0.937$ & $-0.005$ \\
\quad ConvNeXt-B (IN1K) & $0.893$ & $-0.001$ \\
\quad DINOv3 ViT-L/16              & $0.885$ & $+0.028$ \\
\quad DINOv3 ViT-H$^{+}$/16        & $0.873$ & $+0.028$ \\
\quad DINOv3 ConvNeXt-S            & $0.872$ & $+0.058$ \\
\quad CLIP ViT-L/14                & $0.869$ & $+0.057$ \\
\quad DINOv3 ConvNeXt-T            & $0.865$ & $+0.076$ \\
\quad DINOv3 ViT-S$^{+}$/16        & $0.864$ & $+0.084$ \\
\quad ResNet-50                    & $0.861$ & $+0.016$ \\
\quad DINOv3 ViT-S/16              & $0.859$ & $+0.100$ \\
\quad DINOv3 ViT-B/16              & $0.857$ & $+0.058$ \\
\quad DINOv3 ConvNeXt-B            & $0.856$ & $+0.054$ \\
\quad DINOv3 ConvNeXt-L            & $0.854$ & $+0.052$ \\
\quad DINOv3 ViT-7B/16             & $0.839$ & $+0.030$ \\
\quad DINOv2 ViT-g/14              & $0.791$ & $+0.074$ \\
\cmidrule(lr){1-3}
\quad \textit{mean}                & $\mathbf{0.865}$ & $\mathbf{+0.047}$ \\
\midrule
\multicolumn{3}{l}{\emph{Audio (VGGSound)}} \\
\quad SSLAM                        & $0.821$ & $+0.084$ \\
\quad AST                          & $0.815$ & $+0.091$ \\
\quad CED                         & $0.808$ & $+0.088$ \\
\cmidrule(lr){1-3}
\quad \textit{mean}                & $\mathbf{0.815}$ & $\mathbf{+0.088}$ \\
\bottomrule
\end{tabular}
\hfill
\begin{tabular}[t]{lcc}
\toprule
\textbf{Backbone} & \textbf{$r$} & $\Delta$Acc \\
\midrule
\multicolumn{3}{l}{\emph{Masked LM (TinyStories, masked-word)}} \\
\quad ModernBERT-large             & $0.584$ & $+0.126$ \\
\quad NomicBERT                    & $0.581$ & $+0.159$ \\
\quad BERT-large                   & $0.568$ & $+0.102$ \\
\quad RoBERTa-large                & $0.565$ & $+0.096$ \\
\cmidrule(lr){1-3}
\quad \textit{mean}                & $\mathbf{0.575}$ & $\mathbf{+0.121}$ \\
\midrule
\multicolumn{3}{l}{\emph{Autoregressive LM (TinyStories, next-token)}} \\
\quad Qwen3-4B                     & $0.785$ & $+0.070$ \\
\quad Qwen3-1.7B                   & $0.777$ & $+0.089$ \\
\quad Qwen3-8B                     & $0.772$ & $+0.050$ \\
\quad Qwen2.5-0.5B                 & $0.769$ & $+0.102$ \\
\quad Qwen3-0.6B                   & $0.765$ & $+0.122$ \\
\quad Llama-3.1-8B                 & $0.764$ & $+0.041$ \\
\quad Qwen2.5-3B                   & $0.762$ & $+0.076$ \\
\quad Qwen3-32B                    & $0.761$ & $+0.019$ \\
\quad Qwen3-14B                    & $0.754$ & $+0.038$ \\
\quad OLMo-2-13B                   & $0.750$ & $+0.029$ \\
\quad Qwen2.5-1.5B                 & $0.749$ & $+0.102$ \\
\quad Qwen2.5-32B                  & $0.747$ & $+0.036$ \\
\quad Qwen2.5-7B                   & $0.740$ & $+0.052$ \\
\quad Qwen2.5-14B                  & $0.736$ & $+0.050$ \\
\cmidrule(lr){1-3}
\quad \textit{mean}                & $\mathbf{0.759}$ & $\mathbf{+0.063}$ \\
\midrule
\textbf{Overall}                   & $\mathbf{0.787}$ & $\mathbf{+0.065}$ \\
\bottomrule
\end{tabular}
\caption{Per-class Pearson correlation $r$ between trained linear-readout and prototype-classifier accuracy, together with the mean accuracy gap $\Delta\mathrm{Acc}$ (readout minus prototype) on the held-out split. The two classifiers align closely, with the exception of masked language models (see Sec.~\ref{app sec: prototype linear comparison}). }
\label{tab:proto-vs-readout-pearson}
\end{table}

\section{Experimental Setup}
\label{app:setup}

The main text evaluates frozen representations from pretrained backbones spanning
vision, language, and audio, across transformer and convolutional architectures,
supervised, self-supervised, and contrastive objectives.
This appendix gives the full list of datasets, models, and extraction details. 
No backbone is fine-tuned or changed for our analysis.

\subsection{Datasets}
\label{app:B-data}

\paragraph{Vision-ImageNet-1K:}
We use ImageNet-1K (ILSVRC~2012) \citep{deng2009imagenet}, the standard $1000$-class
image benchmark with $\approx\!1.28$M images.

\paragraph{Language-TinyStories:}
For language, the task is token prediction on the validation dataset of TinyStories \citep{eldan2023tinystories}, a
corpus of 20K short synthetic children's stories. We treat token prediction as a multi-class classification problem.

\emph{Content words and lemmatizing}: In order to get semantically meaningful
representations, we keep only data points of \emph{content words}. Following
\citep{arps2024multilingual}, content words are those whose part-of-speech is an
open lexical class of the Universal Dependencies scheme~\citep{de2021universal}: \textsc{noun}, \textsc{propn}, \textsc{verb}, \textsc{adj}, or \textsc{adv}. We
assign part-of-speech tags and lemmas with the spaCy pipeline~\citep{honnibal2020spacy}, and unify all tokens sharing a lemma into a
single class (for example, "say, said, says" or "friend, friends"). We keep
only classes with at least $1000$ occurrences, so that the manifold geometry is
well estimated. Finally, to obtain an identical class set across models with
different tokenizers, we discard the $11$ lemmas that are not realized
consistently in every tokenizer - fragmented or unified into/from sub-words differently in at least one model ("benny, bunny, fire, frog,
grandma, grow, mia, mommy, sweet, window, yummy", for example one tokenizer treats "fireman"$\,\mapsto\,$"fire+man"). This leaves $320$ classes
shared by all backbones.

\emph{Masked-word task (bidirectional encoders).} We mask a single content-word position
at a time, run a full
bidirectional forward pass, and read out the encoder's representation at the masked
position. The representation is the final hidden state feeding the model's masked-LM head. The
class label is the \emph{lemma} of the masked word, restricted to the same open-class content-word construction as above.

\emph{Next-token task (autoregressive models):} From a standard causal forward pass, the
representation is the hidden state (at a specified layer) at the last token \emph{preceding} the
predicted word, and the class label is the lemma of the actual next word, restricted to the same open-class content-word construction as above. 

\paragraph{Audio - VGGSound.}
For audio we use VGGSound \citep{chen2020vggsound}, a large-scale dataset of $\sim\!10$-second
clips sourced from web video. We evaluate on the train split: $310$ single-label classes and
$183,630$ clips ($\sim592$ for each class), decoded to $16$\,kHz mono.

\subsection{Models}
\label{app:B-models}
The canonical roster comprises $36$ pretrained backbones across the four modality groups
(Table~\ref{tab:roster}). Vision contains the self-supervised DINOv3 family
\citep{simeoni2025dinov3}: DINOv3 ViT-S/16 through ViT-7B/16 and DINOv3 ConvNeXt-T through
ConvNeXt-L, together with the self-supervised DINOv2 ViT-g/14 \citep{oquab2023dinov2},
contrastive CLIP ViT-L/14 \citep{radford2021learning}, the supervised ImageNet-21K ViT-B/16
\citep{dosovitskiy2020image}, the supervised ImageNet-1K ConvNeXt-B \citep{liu2022convnet},
and the supervised ResNet-50 \citep{he2016deep}. 

The language backbones are evaluated on the TinyStories dataset \citep{eldan2023tinystories}. The masked-word encoders are RoBERTa-large
\citep{liu2019roberta}, BERT-large \citep{devlin2019bert}, ModernBERT-large
\citep{warner2025smarter}, and NomicBERT \citep{nussbaum2024nomic}. The next-token
autoregressive models are the Qwen2.5-Instruct family ($0.5$B--$32$B)
\citep{qwen2025qwen25technicalreport}, the Qwen3 family ($0.6$B--$32$B)
\citep{yang2025qwen3}, Llama-3.1-8B \citep{grattafiori2024llama}, and
OLMo-2-13B \citep{olmo20242}.

The audio backbones are SSLAM \citep{alex2025sslam}, an EAT-style self-supervised model
fine-tuned on AudioSet-2M; CED-base \citep{dinkel2024ced}, an AudioSet-supervised distilled
transformer; and AST \citep{gong2021ast}, an AudioSet-supervised spectrogram transformer.

\begin{table}[t]
\centering
\footnotesize
\setlength{\tabcolsep}{4pt}          
\caption{Model roster ($36$ backbones). $N$ is the feature dimension of the
extracted representation; $P$ is the number of classes in the task.}
\label{tab:roster}
\begin{tabular}{llrlrl}
\toprule
Family & Model & $N$ & Data / task & $P$ & Objective \\
\midrule
\multicolumn{6}{l}{\emph{Vision --- ImageNet-1K}} \\
Vision & DINOv3 ViT-S/16      & $384$  & ImageNet-1K & $1000$ & Self-supervised \\
Vision & DINOv3 ViT-S+/16     & $384$  & ImageNet-1K & $1000$ & Self-supervised \\
Vision & DINOv3 ViT-B/16      & $768$  & ImageNet-1K & $1000$ & Self-supervised \\
Vision & DINOv3 ViT-L/16      & $1024$ & ImageNet-1K & $1000$ & Self-supervised \\
Vision & DINOv3 ViT-H+/16     & $1280$ & ImageNet-1K & $1000$ & Self-supervised \\
Vision & DINOv3 ViT-7B/16     & $4096$ & ImageNet-1K & $1000$ & Self-supervised \\
Vision & DINOv3 ConvNeXt-T    & $768$  & ImageNet-1K & $1000$ & Self-supervised \\
Vision & DINOv3 ConvNeXt-S    & $768$  & ImageNet-1K & $1000$ & Self-supervised \\
Vision & DINOv3 ConvNeXt-B    & $1024$ & ImageNet-1K & $1000$ & Self-supervised \\
Vision & DINOv3 ConvNeXt-L    & $1536$ & ImageNet-1K & $1000$ & Self-supervised \\
Vision & DINOv2 ViT-g/14      & $1536$ & ImageNet-1K & $1000$ & Self-supervised \\
Vision & CLIP ViT-L/14        & $768$  & ImageNet-1K & $1000$ & Contrastive (image-text) \\
Vision & ViT-B/16 (IN21K)     & $768$  & ImageNet-1K & $1000$ & Supervised (IN21K) \\
Vision & ConvNeXt-B (IN1K)    & $1024$ & ImageNet-1K & $1000$ & Supervised (IN1K) \\
Vision & ResNet-50            & $2048$ & ImageNet-1K & $1000$ & Supervised (IN1K) \\
\midrule
\multicolumn{6}{l}{\emph{Audio --- VGGSound}} \\
Audio & SSLAM     & $768$ & VGGSound & $310$ & Self-supervised + AudioSet-2M FT \\
Audio & CED-base  & $768$ & VGGSound & $310$ & Supervised distillation (AudioSet) \\
Audio & AST       & $768$ & VGGSound & $310$ & Supervised (AudioSet) \\
\midrule
\multicolumn{6}{l}{\emph{Masked LM --- TinyStories masked-word}} \\
Masked LM & RoBERTa-large    & $1024$ & TinyStories (masked) & $320$ & Masked language modeling \\
Masked LM & BERT-large       & $1024$ & TinyStories (masked) & $320$ & Masked language modeling \\
Masked LM & ModernBERT-large & $1024$ & TinyStories (masked) & $320$ & Masked language modeling \\
Masked LM & NomicBERT        & $768$  & TinyStories (masked) & $320$ & Masked language modeling \\
\midrule
\multicolumn{6}{l}{\emph{Autoregressive LM --- TinyStories next-token}} \\
Autoregressive LM & Qwen2.5-0.5B    & $896$  & TinyStories (next-token) & $320$ & Causal language modeling \\
Autoregressive LM & Qwen2.5-1.5B    & $1536$ & TinyStories (next-token) & $320$ & Causal language modeling \\
Autoregressive LM & Qwen2.5-3B      & $2048$ & TinyStories (next-token) & $320$ & Causal language modeling \\
Autoregressive LM & Qwen2.5-7B      & $3584$ & TinyStories (next-token) & $320$ & Causal language modeling \\
Autoregressive LM & Qwen2.5-14B     & $5120$ & TinyStories (next-token) & $320$ & Causal language modeling \\
Autoregressive LM & Qwen2.5-32B     & $5120$ & TinyStories (next-token) & $320$ & Causal language modeling \\
Autoregressive LM & Qwen3-0.6B      & $1024$ & TinyStories (next-token) & $320$ & Causal language modeling \\
Autoregressive LM & Qwen3-1.7B      & $2048$ & TinyStories (next-token) & $320$ & Causal language modeling \\
Autoregressive LM & Qwen3-4B        & $2560$ & TinyStories (next-token) & $320$ & Causal language modeling \\
Autoregressive LM & Qwen3-8B        & $4096$ & TinyStories (next-token) & $320$ & Causal language modeling \\
Autoregressive LM & Qwen3-14B       & $5120$ & TinyStories (next-token) & $320$ & Causal language modeling \\
Autoregressive LM & Qwen3-32B       & $5120$ & TinyStories (next-token) & $320$ & Causal language modeling \\
Autoregressive LM & Llama-3.1-8B    & $4096$ & TinyStories (next-token) & $320$ & Causal language modeling \\
Autoregressive LM & OLMo-2-13B      & $5120$ & TinyStories (next-token) & $320$ & Causal language modeling \\
\bottomrule
\end{tabular}
\end{table}

\subsection{Representation extraction}
\label{app:B-extract}

For vision transformers, we take the class token of the final layer as the
representation. For convolutional architectures (ConvNeXt, ResNet), we take the
global-average-pooled feature vector of the final stage. For audio backbones, we take the
model's pre-classifier pooled embedding: SSLAM its fc-norm of the
class token, CED-base the
mean over its final-layer hidden states, AST its standard
pre-classifier pooled outputs. For masked language models the representation is the final hidden state at the masked position — the input to the masked-LM head. For autoregressive language models the
representation is the hidden state (in a specific layer, see Sec.~\ref{app sec: qwen}) of the last token preceding the predicted
word. In every case, the feature dimensions $N$ reported in Table~\ref{tab:roster} are those
of the stored arrays.

Before any analysis, we z-score each representation
dimension-wise, using the empirical mean and standard deviation of the data.

\subsection{Resources}
\label{app:B-resources}

Feature extraction ran on GPU (A100 and H100); the numerical integration of the theory and
the fitting of the rescaling factor $\lambda$ ran on CPU. The largest extractions were Qwen-14B/32B and OLMo on language, and DINOv3 ViT-7B on vision.

\section{Theoretical Proofs}\label{app sec:theory}

This appendix contains all material deferred from Sec.~\ref{sec:theory:isotropic} and
Sec.~\ref{sec:theory}. Appendix~\ref{app:isotropic-proofs} works out the
isotropic uncorrelated model and derives the corollaries that show how the model fails qualitatively on real data. Appendix~\ref{app:caligned-proofs}
gives the full proofs of the centroid-aligned theory used in the main
text.

\subsection{Isotropic model}
\label{app:isotropic-proofs}

We analyze the maximally unstructured baseline, in which class centroids and
within-class fluctuations are both isotropic Gaussian. It retains only the
ambient dimension $N$, the number of classes $P$, and the noise-to-signal
scale $R$, discarding every correlation between centroids and fluctuations.
We derive its exact large-$N$ accuracy, show that accuracy is self-averaging,
and extract corollaries delineating its regimes. In Section~\ref{section:empirical} these predictions are shown to match
real representations only after the residual-centroid correlations are removed.

\medskip\noindent\textbf{Assumption C.1.1} \emph{(Isotropic uncorrelated model)}.
For each class $\mu=1,\dots,P$ the centroid is drawn
$\mathbf c_\mu\sim_{\mathrm{iid}}\mathcal N(\mathbf 0,I_N)$ and held fixed
(quenched). Conditioned on $\mathbf c_k$, an example of class $k$ is
$\mathbf x=\mathbf c_k+\delta\mathbf x$ with
$\delta\mathbf x\sim\mathcal N(\mathbf 0,R^2 I_N)$, sampled independently across examples.

\medskip\noindent\textbf{Assumption C.1.2} \emph{(High-dimensional regime).}
$N\to\infty$, the number of classes grows at most polynomially,
$P=N^{\mathcal O(1)}$, and the radius $R=R(N)$ may scale with $N$.

\medskip\noindent
We refer to Assumptions~C.1.1--C.1.2 jointly as the \emph{isotropic model}. The
classifier is the unit-centroid (prototype) rule
$\hat y(\mathbf x)=\arg\max_{\mu}\,\hat{\mathbf c}_\mu\!\cdot\!\mathbf x$, with
$\hat{\mathbf c}_\mu:=\mathbf c_\mu/\lVert\mathbf c_\mu\rVert$, and its logits
are $t_\mu:=\hat{\mathbf c}_\mu\!\cdot\!\mathbf x$.

\medskip\noindent\textbf{Lemma C.1.3 }\emph{(Overlaps).}
Let $\mathbf u,\mathbf v\sim_{\mathrm{iid}}\mathcal N(\mathbf 0,I_N)$.
Then, as $N\to\infty$, $\lVert\mathbf u\rVert/\sqrt N\xrightarrow{p}1$ and
$\sqrt N\,(\hat{\mathbf u}\!\cdot\!\hat{\mathbf v})\xrightarrow{d}\mathcal N(0,1)$.
Consequently, the dot product of the target centroid with
the true and rival directions obeys
\begin{equation}
\hat{\mathbf c}_\mu\!\cdot\!\mathbf c_k
=\varepsilon_\mu+\delta_{\mu k}\big(\sqrt N-\varepsilon_\mu\big),
\qquad \varepsilon_\mu\sim_{\mathrm{iid}}\mathcal N(0,1).
\end{equation}

\smallskip\noindent\emph{Proof.}
$\mathbf u\!\cdot\!\mathbf v=\sum_{i} u_i v_i$ is a sum of $N$ i.i.d.\ centred
unit-variance variables, so $\mathbf u\!\cdot\!\mathbf v/\sqrt N\xrightarrow{d}
\mathcal N(0,1)$ by the CLT, while $\lVert\mathbf u\rVert/\sqrt N\xrightarrow{p}1$
by the law of large numbers. 
\hfill$\square$

\medskip\noindent\textbf{Theorem C.1.4 (Mean accuracy).}
\textit{Under the isotropic model,}
\begin{equation}\label{eq:isotropic-app}
\overline{\mathbb{E}_{\delta\mathbf{x}}[\operatorname{Acc}]}
=\mathbb E_{z\sim\mathcal N(0,1)}
\!\left[\Phi^{P-1}\!\left(\frac{\sqrt N+Rz}{\sqrt{1+R^2}}\right)\right].
\end{equation}

\smallskip\noindent
where $\overline{\mathbb{E}_{\mathbf{\delta x}}[\operatorname{Acc}]}$ denotes the average over the quenched disorder $\{\mathbf{c}_\mu\}$ after averaging over the thermal $\delta \mathbf{x}$.\\
\emph{Proof.}
Conditional on the example and centroids, the accuracy is the probability that
the correct logit is the largest,
\begin{equation}
\operatorname{Acc}\,\big|\,\delta\mathbf x,\{\mathbf c_\nu\}
=\int_{-\infty}^{\infty}\! dt_k\,P(t_k)\prod_{\mu\neq k}
\int_{-\infty}^{t_k}\! dt_\mu\,P(t_\mu),
\qquad P(t_\mu)=\delta\!\big(t_\mu-\hat{\mathbf c}_\mu\!\cdot\!\mathbf x\big).
\end{equation}
Writing each $\delta$ through its Fourier representation introduces conjugate
variables $\hat t_\mu$:
\begin{equation}
\operatorname{Acc}\,\big|\,\delta\mathbf x,\{\mathbf c_\nu\}
=\int_{-\infty}^{\infty}\!\frac{dt_k\,d\hat t_k}{2\pi}
\prod_{\mu\neq k}\int_{-\infty}^{t_k}\!\frac{dt_\mu\,d\hat t_\mu}{2\pi}
\exp\!\Big(i\sum_{\mu=1}^{P}\hat t_\mu\big(t_\mu-\hat{\mathbf c}_\mu\!\cdot\!
(\mathbf c_k+\delta\mathbf x)\big)\Big).
\end{equation}

Carrying out the Gaussian
$\delta\mathbf{x}$ and $\hat t_\mu$ integrals,
\begin{equation}\label{eq:intermediate avg}
\mathbb E_{\delta x}\!\big[\operatorname{Acc}\,\big|\,\{\mathbf c_\mu\}\big]
=\int_{-\infty}^{\infty}\!\frac{dt_k}{\sqrt{2\pi R^2}}
\prod_{\mu\neq k}\int_{-\infty}^{t_k}\!\frac{dt_\mu}{\sqrt{2\pi R^2}}
\exp\!\Big(-\frac{1}{2R^2}\sum_{\mu=1}^{P}
\big(t_\mu-\hat{\mathbf c}_\mu\!\cdot\!\mathbf c_k\big)^2\Big).
\end{equation}
Lemma~C.1.3 sets
$\hat{\mathbf c}_\mu\!\cdot\!\mathbf c_k=\varepsilon_\mu+\delta_{\mu k}
(\sqrt N-\varepsilon_\mu)$. The target mean concentrates at $\sqrt N$, while
integrating each rival overlap $\varepsilon_\mu\sim\mathcal N(0,1)$
inflates the rival variance $R^2\to1+R^2$:
\begin{equation}
\overline{\operatorname{Acc}}
=\int_{-\infty}^{\infty}\!\frac{dt_k}{\sqrt{2\pi R^2}}
\prod_{\mu\neq k}\int_{-\infty}^{t_k}\!\frac{dt_\mu}{\sqrt{2\pi(1+R^2)}}
\exp\!\Big(-\frac{(t_k-\sqrt N)^2}{2R^2}
-\frac12\sum_{\mu\neq k}\frac{t_\mu^2}{1+R^2}\Big).
\end{equation}
The rival integrals factorize into $\Phi\!\big(t_k/\sqrt{1+R^2}\big)$;
substituting $t_k=\sqrt N+Rz$ yields Eq.~\ref{eq:isotropic-app}.
\hfill$\square$

\medskip\noindent
The average Eq.~\ref{eq:isotropic-app} is taken over both variability $\delta{\bf x}$ and centroids; the
next result shows that the accuracy is self-averaging, such that taking the annealed average over the centroids disorder is valid.

\medskip\noindent\textbf{Proposition C.1.5} \emph{(Self-averaging).}
Under the isotropic model, the accuracy is self-averaging.
$\operatorname{Var}(\operatorname{Acc})=\mathcal{O}\!\big(1/(PR^2)\big)\to0$.

\smallskip\noindent\emph{Proof.}
We bound the variance of the disorder-averaged accuracy. Introduce two
independent replicas $\mathbf x,\mathbf x'$ of class $k$ sharing the
centroids. As in Eq.\ref{eq:intermediate avg}, we get,
\begin{align}
\mathbb E_{\delta x}\!\big[\operatorname{Acc}^2\,\big|\,\{\mathbf c_\mu\}\big]
&=\int_{-\infty}^{\infty}\!\frac{dt_k\,dt_k'}{2\pi R^2}
\prod_{\mu\neq k}\int_{-\infty}^{t_k}\!\frac{dt_\mu}{\sqrt{2\pi R^2}}
\int_{-\infty}^{t_k'}\!\frac{dt_\mu'}{\sqrt{2\pi R^2}} \nonumber\\
&\hphantom{{}={}}\exp\!\Big(-\frac{1}{2R^2}\sum_{\mu}\!\big[(t_\mu-\hat{\mathbf c}_\mu\cdot\mathbf c_k)^2
+(t_\mu'-\hat{\mathbf c}_\mu\cdot\mathbf c_k)^2\big]\Big).
\end{align}
Averaging over the centroids
couples the replicas through the shared mean. The integral becomes bivariate normal with variance $1+R^2$ and
covariance $1$,
\begin{equation}
\overline{\operatorname{Acc}^2}
=\int\!\frac{dt_k\,dt_k'}{2\pi R^2}
\prod_{\mu\neq k}\frac{1}{\mathcal{Z}}\int_{-\infty}^{t_k+\sqrt N}dt_\mu
\int_{-\infty}^{t_k'+\sqrt N}dt_\mu'
\exp\!\Big(-\frac{t_k^2+t_k'^2}{2R^2}
-\frac{(1{+}R^2)(t_\mu^2+t_\mu'^2)-2t_\mu t_\mu'}{2R^2(R^2+2)}\Big),
\end{equation}
with the normalization $\mathcal Z=2\pi R\sqrt{R^2+2}$.

The regime $R=\mathcal{O}(1)$: The target shift $\sqrt N\to\infty$ pushes the rival
limits to $+\infty$, the rival integrals yield $1$ up to an exponentially small correction whenever $\ln{P}=o(N)$, and
$\overline{\operatorname{Acc}^2}=\overline{\operatorname{Acc}}^2=1$, so
$\operatorname{Var}\to0$.

The regime $R\sim\sqrt N$: Set $r=R/\sqrt N$ and rescale. The inter-replica
coupling is weak, of order $1/(Nr^2)$, and we can expand in the coupling. Solving the uncoupled Gaussian integrals yields:
\begin{equation}
\begin{aligned}
\overline{\operatorname{Acc}^2}
&=\mathbb E_{z,z'}\!\bigg[\Big(\Phi(z{+}\tfrac1r)\Phi(z'{+}\tfrac1r)
+\tfrac{1}{Nr^2}\phi(z{+}\tfrac1r)\phi(z'{+}\tfrac1r)+\mathcal{O}(N^{-2})\Big)^{P-1}\bigg]\\
&=\overline{\operatorname{Acc}}^2
+\frac{P-1}{Nr^2}\Big(\mathbb E_z\big[\Phi^{P-2}(z{+}\tfrac1r)
\phi(z{+}\tfrac1r)\big]\Big)^2+\mathcal{O}\!\Big(\tfrac{\ln P}{N^2P^2}\Big).
\end{aligned}
\end{equation}
where $\phi(x)$ is a standard normal PDF, and $\Phi(x)$ its CDF. The expectation can be performed using the Laplace method, yielding the classic extreme-value extremum  $z^\ast+1/r=\sqrt{2\ln P}$, $\phi(z^\ast{+}1/r)=\sqrt{2\ln P}/P$, and the width of the Laplace region $\Delta z=1/\left(2\sqrt{\ln {P}}\right)$. We note that 
the higher binomial terms are suppressed not by the $1/N$ prefactors but by the
Gaussian PDF densities $\phi$, each one of the order $\sqrt{2\ln P}/P$ , which cancel the binomial growth. 
\begin{equation}
\mathrm{Var(Acc)}=\frac{P-1}{Nr^2}\Big(\mathbb E_z\big[\Phi^{P-2}(z{+}\tfrac1r)
\phi(z{+}\tfrac1r)\big]\Big)^2=\mathcal{O}(\frac{1}{PNr^2})=\mathcal{O}(\frac{1}{PR^2})\rightarrow 0
\end{equation}
In both regimes $\operatorname{Var}(\operatorname{Acc})\to0$.
\hfill$\square$

\medskip\noindent
The mean accuracy Eq.\ref{eq:isotropic-app} has four immediate consequences,
delineating the regimes of the isotropic model.

\medskip\noindent\textbf{Corollary C.1.6} \emph{(Separable regime).}
For $R=\mathcal O(1)$ the classes are well separated and
classification is trivial: $\mathrm{Acc}^{\mathrm{}}\to1$.

\smallskip\noindent\emph{Proof.}
The scale of the maximum of $P$ i.i.d.
Gaussians with variance $\sigma^2=1+R^2$ converges to $\sigma\sqrt{2\ln P}$ for large $P$. The signal is $\mathbb{E}[t_k]=\sqrt{N}$. From Assumption C.1.2, $\ln P=\mathcal O(\ln N)$ thus for any $R=\mathcal{O}(1)$, $\sigma\sqrt{2\ln P}\ll\sqrt{N}$ which yields $\mathrm{Acc}\to1$.
\hfill$\square$

\medskip\noindent\textbf{Corollary C.1.7} \emph{(Saturation regime).}
In the infinite-radius limit $R/\sqrt N\to\infty$ the isotropic residual
dominates the input, which retains no information about the true class: all
$P$ classes are equally likely and accuracy collapses to chance,
$\mathrm{Acc}^{\mathrm{}}\to1/P$.

\smallskip\noindent\emph{Proof.}
Dividing through by $R$, the argument of $\Phi$ in Eq.\ref{eq:isotropic-app} tends to
$z$ pointwise, so bounded convergence gives
$\mathrm{Acc}^{\mathrm{}}\to\mathbb E_z[\Phi^{P-1}(z)]
=\mathbb P(\xi_k>\max_{\mu\neq k}\xi_\mu)=1/P$ for
$\xi_1,\dots,\xi_P\sim_{\mathrm{iid}}\mathcal N(0,1)$, by permutation symmetry.
\hfill$\square$

\medskip\noindent\textbf{Corollary C.1.8} \emph{(Critical radius).}
Non-perfect accuracy requires large radii:\\
$\mathrm{Acc}^{\mathrm{}}<1\Rightarrow R=\Omega(\sqrt{N/\ln P})$.

\smallskip\noindent\emph{Proof.}
Contrapositive. If $R=o(\sqrt{N/\ln P})$ then $1+R^2=o(N/\ln P)$, so the
argument of $\Phi$ in Eq.\ref{eq:isotropic-app} is
$(\sqrt N+Rz)/\sqrt{1+R^2}=\omega(\sqrt{\ln P})\gg\sqrt{2\ln P}$, and Corollary~C.1.6 forces $\mathrm{Acc}^{\mathrm{}}\to1$.
\hfill$\square$

\medskip\noindent\textbf{Corollary C.1.9} \emph{(Per-class fluctuations).}
Per-class accuracy fluctuations are of order $1/R\sqrt{P}$.

\smallskip\noindent\emph{Proof.}
Calculation of the variance in Proposition C.1.5 yields $\mathrm{Var(Acc)}=\mathcal{O}(1/{PR^2})$. Different classes are equivalent to a resampling of the centroids, so per-class fluctuations are of the same order. 
\hfill$\square$
\subsection{Centroid-aligned  model}
\label{app:caligned-proofs}

\medskip\noindent

\medskip\noindent\textbf{Definition C.2.1} \emph{(Centroid geometry).}
Fix the true class $k$. For $\mu,\nu\in\{1,\dots,P\}$ let
$G_{\mu\nu}:=\hat{\mathbf c}_\mu\!\cdot\!\hat{\mathbf c}_\nu$ be the Gram matrix of the unit
centroids, and let $g_\mu:=\hat{\mathbf c}_k\!\cdot\!\hat{\mathbf c}_\mu$ be the centroid overlap
of class $\mu$ with the true class, so that $g_k=1$. The rival support is $\mathcal R_K$ and
$\mathcal R^{+}_K:=\{k\}\cup\mathcal R_K$, as defined in Sec.~\ref{sec:theory}.

\medskip\noindent\textbf{Assumption C.2.2} \emph{(Deterministic centroids and Gaussian projections).}
We assume that the centroids $\{\mathbf c_\mu\}_{\mu=1}^{P}$ are deterministic and held fixed, and that the restricted Gram matrix of $\mathcal R^{+}_K$ is invertible. The source of stochasticity in the system is the sampling of Gaussian projection
coefficients $\{s_\mu\}_{\mu=1}^P$ and $s_{_\perp}$.

\medskip\noindent\textbf{Definition C.2.3} \emph{(Centroid-aligned generative model).}
An example of class $k$ is generated by
\begin{equation}\label{eq:cal-model}
\mathbf{x}(\mathbf{s})=\left\Vert \mathbf{c}_{k}\right\Vert \left[\hat{\mathbf{c}}_{k}
+R\left(\sum_{\mu,\nu\in\mathcal{R}^{+}_{K}}\!\!\sigma_{\mu}s_{\mu}\,G^{-1}_{\mu\nu}\hat{\mathbf{c}}_{\nu}
+\sigma_{_{\perp}}s_{_{\perp}}\hat{\mathbf{e}}_{_{\perp}}\right)\right],
\end{equation}
where $\hat{\mathbf e}_{_\perp}$ is orthogonal to
$\operatorname{span}(\hat{\mathbf c}_\mu)_{\mu\in\mathcal R^{+}_K}$, $R$ is the empirical class
radius, and $\sigma_\mu$ is the projected standard deviation along $\hat{\mathbf c}_\mu$.

\medskip\noindent\textbf{Definition C.2.4} \emph{(Normalized logits).}
The prototype logits of (Def~2.1) are defined as 
$t_\mu=\hat{\mathbf c}_\mu\!\cdot\!\mathbf x$. Throughout this section it is convenient to divide all of
them by the common factor $\lVert\mathbf c_k\rVert$, and we set
\begin{equation}\label{eq:cal-logit-def}
t_\mu:=\frac{\mathbf x\!\cdot\!\hat{\mathbf c}_\mu}{\lVert\mathbf c_k\rVert},
\qquad \mu\in\mathcal R^{+}_K .
\end{equation}
Since $\lVert\mathbf c_k\rVert$ is a common factor independent of $\mu$, the $\mathrm{argmax}$ properties of the logits do not change. The classifier and its
accuracy are therefore those of Definition~2.1, and no generality is lost. We restrict the logits only to the sparse set of rivals $\mathcal{R}_K$, and neglect weak competitors. 

\medskip\noindent\textbf{Lemma C.2.5} \emph{(Logit representation).}
Under Assumption~C.2.2 and Definitions~C.2.3--C.2.4, $t_\mu=g_\mu+R\,\sigma_\mu s_\mu$ for every
$\mu\in\mathcal R^{+}_K$.

\smallskip\noindent\emph{Proof.}
Taking the inner product of Eq.~\ref{eq:cal-model} with $\hat{\mathbf c}_\mu$ and dividing by
$\lVert\mathbf c_k\rVert$ gives
$t_\mu=g_\mu+R\big(\sum_{\alpha,\nu\in\mathcal R^{+}_K}\sigma_\alpha s_\alpha G^{-1}_{\alpha\nu}G_{\nu\mu}
+\sigma_{_\perp} s_{_\perp}\,\hat{\mathbf e}_{_\perp}\!\cdot\!\hat{\mathbf c}_\mu\big)$.
The last term vanishes since $\hat{\mathbf e}_{_\perp}\perp\hat{\mathbf c}_\mu$ for
$\mu\in\mathcal R^{+}_K$, and $\sum_\nu G^{-1}_{\alpha\nu}G_{\nu\mu}=\delta_{\alpha\mu}$ by
Assumption~C.2.2.
\hfill$\square$

\medskip\noindent\textbf{Definition C.2.6} \emph{(Projection covariance).}
The coefficients are centered and standardized by definition, $\mathbb E_k[s_\mu]=0$ and
$\mathbb E_k[s^{2}_\mu]=1$. We write
\begin{equation}\label{eq:cal-Sigma}
\mathbb E_k[s_\mu s_\nu]=\frac{\Sigma_{\mu\nu}}{R^{2}\sigma_\mu\sigma_\nu},
\qquad \mu,\nu\in\mathcal R^{+}_K .
\end{equation}
By Lemma~C.2.5 this makes $\Sigma_{\mu\nu}=\mathrm{Cov}_k[t_\mu,t_\nu]$ the covariance matrix of the
logit vector itself. No structure is imposed on $\Sigma$ beyond
that of a well-conditioned covariance matrix.


\medskip\noindent\textbf{Proposition C.2.7} \emph{(Accuracy as an orthant probability).}
Under Lemma~C.2.5, Definition~C.2.6 and Assumption~C.2.2,
\begin{equation}\label{eq:cal-orthant}
\mathrm{Acc}^{\mathrm{th}}(R)=\frac{1}{\sqrt{\det\Sigma}}
\int_{-\infty}^{\infty}\!\frac{dt_{k}}{\sqrt{2\pi}}
\prod_{\mu\in\mathcal{R}_{K}}\int_{-\infty}^{t_{k}}\!\frac{dt_{\mu}}{\sqrt{2\pi}}\,
\exp\!\left(-\frac{1}{2}\sum_{\mu,\nu\in\mathcal{R}^{+}_{K}}
\left(t_{\mu}-g_{\mu}\right)\Sigma^{-1}_{\mu\nu}\left(t_{\nu}-g_{\nu}\right)\right).
\end{equation}

\smallskip\noindent\emph{Proof.}
By Lemma~C.2.5, Definition~C.2.6 and Assumption~C.2.2 the vector
$(t_\mu)_{\mu\in\mathcal R^{+}_K}$ is Gaussian with mean $(g_\mu)$ and covariance $\Sigma$. An
example is classified correctly iff $t_\mu<t_k$ for all $\mu\in\mathcal R_K$.
\hfill$\square$

\medskip\noindent
Eq.~\ref{eq:cal-orthant} is a multivariate orthant probability, which in general can be
evaluated only numerically (for example by Monte-Carlo simulation). A specific structure admits an analytical solution.

\medskip\noindent\textbf{Lemma C.2.8} \emph{(Gaussian conditioning).}
Under Assumption~C.2.2, set $\rho_\mu:=\mathbb E_k[s_\mu s_k]$ for $\mu\in\mathcal R_K$. Then for
all $\mu,\nu\in\mathcal R_K$,
\begin{equation}\label{eq:cal-gausscond}
\mathbb{E}_{k}\!\left[t_{\mu}\mid t_{k}\right]=g_{\mu}+\rho_{\mu}\sigma_{\mu}\frac{t_{k}-1}{\sigma_{k}},
\qquad
\mathrm{Cov}\!\left[t_{\mu},t_{\nu}\mid t_{k}\right]
=R^{2}\sigma_{\mu}\sigma_{\nu}\left(\mathbb E_k[s_\mu s_\nu]-\rho_{\mu}\rho_{\nu}\right),
\end{equation}
and in particular the conditional covariance does not depend on $t_{k}$.

\smallskip\noindent\emph{Proof.}
By Definition~C.2.6, $\mathrm{Cov}(t_\mu,t_\nu)=\Sigma_{\mu\nu}$,
$\Sigma_{\mu k}=R^{2}\sigma_\mu\sigma_k\rho_\mu$ and $\Sigma_{kk}=R^{2}\sigma^{2}_k$, while
$\mathbb E_k[t_\mu]=g_\mu$ and $\mathbb E_k[t_k]=1$. For a jointly Gaussian vector,
$\mathbb E[t_\mu\mid t_k]=\mathbb E[t_\mu]+\Sigma_{\mu k}\Sigma_{kk}^{-1}(t_k-\mathbb E[t_k])$ and
$\mathrm{Cov}[t_\mu,t_\nu\mid t_k]=\Sigma_{\mu\nu}-\Sigma_{\mu k}\Sigma_{\nu k}\Sigma_{kk}^{-1}$.
Substituting gives Eq.~\ref{eq:cal-gausscond}.
\hfill$\square$

\medskip\noindent\textbf{Assumption C.2.9} \emph{(Conditional independence).}
Conditionally on $t_k$ the rival logits are uncorrelated: $\mathrm{Cov}[t_\mu,t_\nu\mid t_k]=0$ for
all $\mu\neq\nu$ in $\mathcal R_K$. Equivalently, by Lemma~C.2.8,
\begin{equation}\label{eq:cal-condind}
\mathbb E_k[s_\mu s_\nu]=\delta_{\mu\nu}\left(1-\rho^{2}_{\mu}\right)+\rho_{\mu}\rho_{\nu},
\qquad\text{so that}\qquad
\mathrm{Cov}\!\left[t_{\mu},t_{\nu}\mid t_{k}\right]
=\delta_{\mu\nu}R^{2}\sigma^{2}_{\mu}\left(1-\rho^{2}_{\mu}\right).
\end{equation}

\medskip\noindent\textbf{Theorem C.2.10} \emph{(Conditional factorization).}
Under Assumption~C.2.9,
\begin{equation}\label{eq:cal-factored}
\mathrm{Acc}^{\mathrm{th}}(R)=\int_{-\infty}^{\infty}\frac{dt_{k}}{R\sigma_{k}\sqrt{2\pi}}
\exp\left({-\frac{(t_k-1)^2}{2R^2\sigma_k^2}}\right)\prod_{\mu\in\mathcal{R}_{K}}\Phi\!\left(
\frac{t_{k}\left(1-\rho_{\mu}\sigma_{\mu}/\sigma_{k}\right)+\rho_{\mu}\sigma_{\mu}/\sigma_{k}-g_{\mu}}
{R\sigma_{\mu}\sqrt{1-\rho^{2}_{\mu}}}\right).
\end{equation}

\smallskip\noindent\emph{Proof.}
Condition on $t_k$. By Lemma~C.2.8 and Assumption~C.2.9 the variables
$(t_\mu)_{\mu\in\mathcal R_K}$ are then independent Gaussians with mean
$g_\mu+\rho_\mu\sigma_\mu(t_k-1)/\sigma_k$ and variance $R^{2}\sigma^{2}_\mu(1-\rho^{2}_\mu)$, so
the orthant probability factorizes into
$\prod_\mu\Pr[t_\mu<t_k\mid t_k]$, which yields Eq.~\ref{eq:cal-factored}.
\hfill$\square$

\medskip\noindent\textbf{Corollary C.2.11} \emph{(Standardized form).}
Substituting $t_k=R\sigma_k z+1$ in Eq.~\ref{eq:cal-factored} yields
\begin{equation}
\mathrm{Acc}^{\mathrm{th}}_k(R)=\mathbb{E}_{z\sim\mathcal N(0,1)}
\left[\prod_{\mu\in\mathcal{R}_{K}}\Phi\!\left(
\frac{\sigma_{k}-\rho_{\mu}\sigma_{\mu}}{\sigma_{\mu}\sqrt{1-\rho^{2}_{\mu}}}\,z
+\frac{1-g_{\mu}}{R\sigma_{\mu}\sqrt{1-\rho^{2}_{\mu}}}\right)\right],
\end{equation}
which is Theorem~5.5 of the main text.


\medskip\noindent\textbf{Proposition C.2.12} \emph{(Unconditional moments).}
Under Assumption~C.2.9, for $\mu,\nu\in\mathcal R_K$,
\begin{equation}
\mathbb{E}_{k}\!\left[t_{\mu}\right]=g_{\mu},
\qquad
\Sigma_{\mu\nu}=\mathrm{Cov}_{k}\!\left[t_{\mu},t_{\nu}\right]
=R^{2}\left(\delta_{\mu\nu}\sigma^{2}_{\mu}\left(1-\rho^{2}_{\mu}\right)
+\sigma_{\mu}\sigma_{\nu}\rho_{\mu}\rho_{\nu}\right),
\end{equation}
which is a rank-one correction to a diagonal covariance. This is the covariance structure of
$\mathbf s$ stated in Assumption~5.4 of the main text.

\smallskip\noindent\emph{Proof.}
From Lemma~C.2.5, $\mathbb E_k[t_\mu]=g_\mu$ and
$\Sigma_{\mu\nu}=R^{2}\sigma_\mu\sigma_\nu\mathbb E_k[s_\mu s_\nu]$. Inserting the covariance structure from
Eq.~\ref{eq:cal-condind} yields the result.
\hfill$\square$

\medskip\noindent\textbf{Remark C.2.13} \emph{(Quenched versus deterministic centroids).}
Assumption~C.2.2 treats the centroids as deterministic. Drawing them instead from a Gaussian
ensemble whose second moments reproduce the overlap structure $G$ does not change the result for
large $N$: the induced fluctuations of the coefficients $g_\mu$, $\sigma_\mu$ and $\rho_\mu$
entering Eq.~\ref{eq:cal-factored} are subleading corrections to their $\mathcal O(1)$ values, and
the accuracy can be shown to be self-averaging by the same argument as in Proposition~C.1.5.

\section{Synthetic Data}
\label{app sec: synthetic}

\begin{figure}
  \centering
  \includegraphics[width=\linewidth]{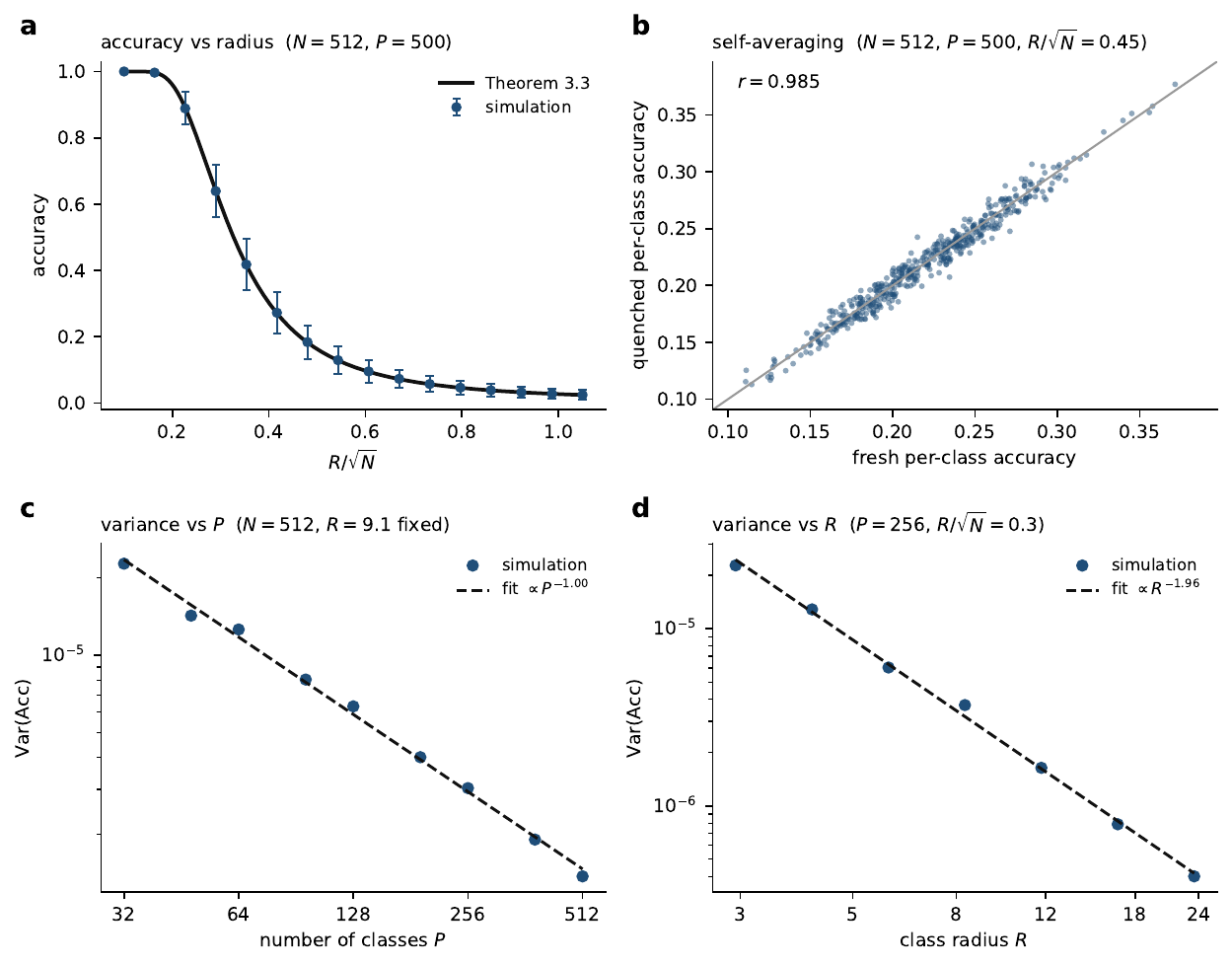}
  \caption{\textbf{Validation of the isotropic model's theory on synthetic data.}
  \textbf{(a)}~Class-averaged accuracy versus $R/\sqrt{N}$: simulation (markers,
  mean $\pm$ between-class std) against Theorem~3.3 (line), $N=512$, $P=500$.
  \textbf{(b)}~Self-averaging: quenched per-class accuracy (one shared centroid
  bank) versus the fresh-rival accuracy (rivals redrawn per example), the line is $y=x$. $r=0.985$, confirming self-averaging - the thermal average is equivalent to the quenched average. \textbf{(c)}~Across-realization variance of the accuracy versus $P$ at
  fixed radius $R=9.05$, fitted power law
  $\propto P^{-1.00}$, predicted by Proposition C.1.5. \textbf{(d)}~Variance versus radius $R$ (swept via the
  dimension at fixed $R/\sqrt{N}$), fitted power law $\propto R^{-1.96}$, predicted by Proposition C.1.5. }
  \label{fig:v19-synth-isotropic}
\end{figure}

\begin{figure}
  \centering
  \includegraphics[width=\linewidth]{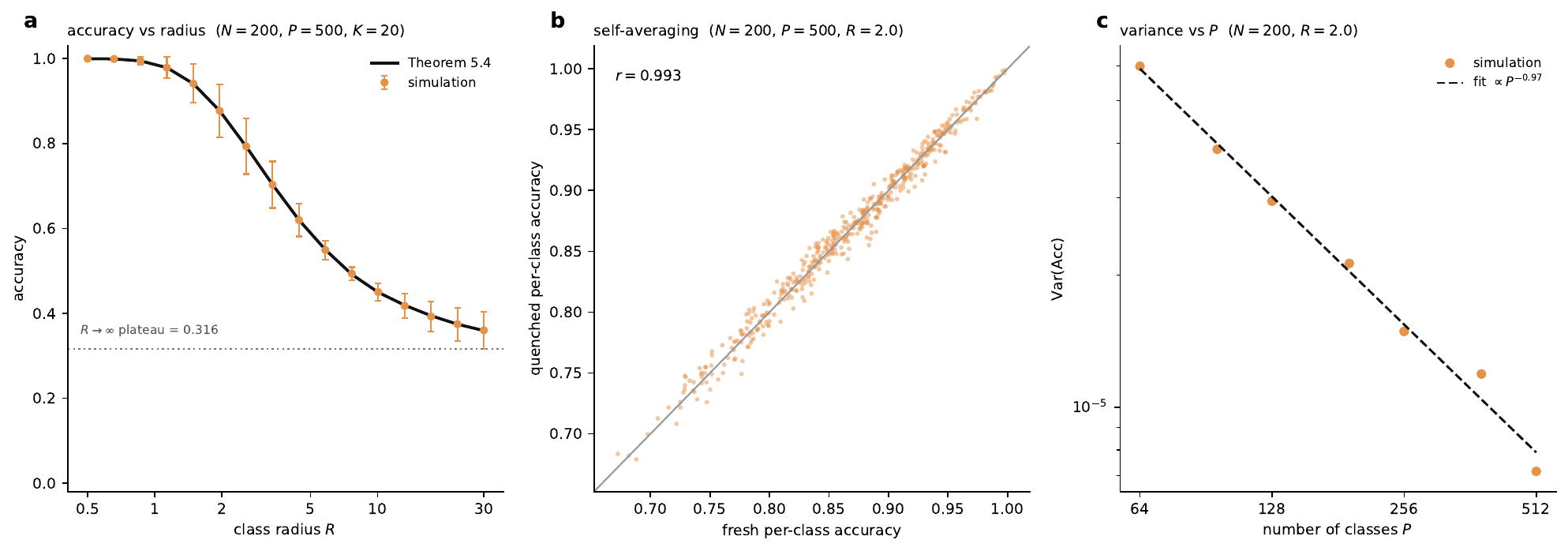}
  \caption{\textbf{Synthetic validation of the centroid-aligned model.} Examples are
  generated according to the centroid-aligned variability model (Sec.~\ref{sec:theory} and Sec.~\ref{app:caligned-proofs}) with $K=20$ rivals, $N=200$, $P=500$. The geometry is calibrated to DINOv3 ViT-7B: own-class projected std $\sigma_k=0.24$,
  rival stds $\sigma_\mu=0.12\,\mu^{-0.4}$.
  \textbf{(a)}~Class-averaged accuracy versus class radius $R$: simulation
  (markers) against Theorem~5.5 (line). The dotted line is the $R\to\infty$
  plateau $0.316$, which is set by $\sigma_k/\sigma_\mu$ alone.
  \textbf{(b)}~Quenched centroid versus fresh per-example sampling of the centroids, $y=x$
  line, $r=0.993$, validating self-averaging. \textbf{(c)}~Across-realization variance versus $P$ at fixed
  radius, fitted $\propto P^{-0.97}$, similar to the isotropic model (Proposition C.1.5).}
  \label{fig:v19-synth-aligned}
\end{figure}

We validate the two generative models of the main text on synthetic data, where
the geometry is controlled exactly and the analytic predictions can be checked
without any of the confounds of real representations. Figure~\ref{fig:v19-synth-isotropic}
tests the isotropic uncorrelated model of Section~3, and
Figure~\ref{fig:v19-synth-aligned} tests the centroid-aligned model of Section~5. For each model we verify three claims: (i)~the analytic accuracy
matches the simulation; (ii)~the accuracy is self-averaging, i.e.\ a single
quenched centroid bank gives the same per-class accuracy as a fresh, independently
drawn set of rivals for every example; and (iii)~the fluctuation of the accuracy
across realizations shrinks with the number of classes $P$ and for the isotropic model also with the class radius
$R$ as predicted by the previous section.

\section{Rival Selection}
\label{app sec: K}

For a given example, only a small number of the $P$ classes are genuine
competitors. This is the reason that the theory of
Section~\ref{sec:theory} evaluates the accuracy over a small rival set
$\mathcal{R}_K$. This appendix (i)
compares four possible rival-selection rules on convergence to the full $P$-way classification (ii) shows that the theory's error saturates by $K\approx10$-$20$, so $K=20$ is a
principled choice.

\subsection{Rival-selection rules}
\label{app sec: K:selectors}

Fix a class $k$. We define a ranking method for every candidate
rival $\mu\neq k$:\\
\textbf{Centroid correlations}: $g_\mu=\hat{\mathbf{c}}_{k}\cdot\hat{\mathbf{c}}_{\mu}$ , (descending).
\\
\textbf{Projection variance}: $\sigma_\mu=\sqrt{\mathbb{E}_{k}\!\big[(\delta\mathbf{x}\cdot\hat{\mathbf c}_{\mu})^{2}\big]\big/\mathbb{E}_{k}\!\big[\lVert\delta\mathbf{x}\rVert^{2}\big]}$, (descending).
\\
\textbf{Theory margin}: $(1-g_{\mu})/\left(R\sigma_{\mu}\sqrt{1-\rho^{2}_{\mu}}\right)$, (ascending).  The z-independent argument in Eq.\ref{eq:caligned-acc}. \\
\textbf{Gaussian margin}: $(1-g_\mu)/{\sqrt{\sigma_k^2+\sigma_\mu^2-2\,\sigma_k\sigma_\mu\rho_\mu}}$, (ascending). The exact z-expectation over Eq.\ref{eq:caligned-acc} for one rival.

\subsection{Convergence of the \texorpdfstring{$K$}{K}-way classification accuracy}
\label{app sec: K:threshold}

Restricting the classification to a smaller rival set can only improve accuracy, so the $K$-way accuracy
exceeds the full $P$-way accuracy and decreases monotonically to it as $K$ grows.
Table~\ref{tab:v19-rivals-thr} reports, for each selector, the median number of
rivals a backbone needs before its $K$-way accuracy is within a fixed band of the
full accuracy, aggregated over all $36$ backbones. The margin and variance
selectors predict the $P$-way classification at the smallest $K$; the centroid-correlation and
Gaussian-margin selectors need substantially more rivals at the tight $1\%$ band,
confirming that ranking by projected variance (or by the conditional margin) is
the most efficient way to choose the true competitors. The conclusion is consistent for all modalities. Since the projection variance and the conditional margin perform comparably (within five rivals at every band), we adopt the projection variance in the main text, as it is a single measured quantity and is therefore cheaper to compute and easier to interpret.

\subsection{Optimal \texorpdfstring{$K$}{K} for theory predictions}
\label{app sec: K:saturation}

Increasing $K$ enlarges the rival set the theory integrates over. Table~\ref{tab:v19-rivals-rmse}
tracks the root-mean-square error between the predicted and empirical per-class
accuracies as a function of $K$. The error drops steeply up to $K\approx10$ and is
essentially flat thereafter: at $K=20$ the mean error is within $0.003$ of its
$K\!\to\!300$ floor for $31$ of the $36$ backbones, and the median number of rivals needed to come within $0.001$ of the
floor is $20$, the value we chose for $K$ in the theory throughout.\\

\begin{table}[H]
  \centering
  \footnotesize
  \caption{\textbf{Rivals needed to reach the full accuracy.} Median number of
  rivals $K$ (over $36$ backbones) at which the restricted $K$-way accuracy comes
  within $5/3/2/1\%$ of the full $P$-way accuracy, per selector. Smaller is
  better.}
  \label{tab:v19-rivals-thr}
  \begin{tabular}{lcccc}
    \toprule
    Selector & within $5\%$ & within $3\%$ & within $2\%$ & within $1\%$ \\
    \midrule
    Centroid correlations ($g_\mu$)  & 14 & 24 & 36 & 87 \\
    Projection variance ($\sigma_\mu$) & 14 & 20 & 30 & 54 \\
    Theory margin                      & 11 & 18 & 25 & 49 \\
    Gaussian margin                    & 10 & 19 & 28 & 73 \\
    \bottomrule
  \end{tabular}
\end{table}

\begin{table}[H]
  \centering
  \footnotesize
  \caption{\textbf{Rivals needed within $1\%$, by modality.} Median $K$ to reach
  within $1\%$ of the full accuracy, per selector and modality ($n$ backbones).}
  \label{tab:v19-rivals-mod}
  \begin{tabular}{lccccc}
    \toprule
    Modality & $n$ & Centroids corr.\ & Proj.\ variance & Theory margin & Gaussian margin \\
    \midrule
    Vision             & 15 & 76  & 47 & 45 & 56  \\
    Autoregressive LM  & 14 & 102 & 89 & 82 & 108 \\
    Masked LM          & 4  & 48  & 42 & 35 & 43  \\
    Audio              & 3  & 60  & 39 & 37 & 49  \\
    \bottomrule
  \end{tabular}
\end{table}

\begin{table}[H]
  \centering
  \footnotesize
  \caption{\textbf{The theory's error saturates by $K\approx10$--$20$.} Mean
  per-class accuracy RMSE of the theory at $K=10$ and $K=20$, its large-$K$
  floor, and the number of backbones already within $0.003$ of that floor (large $K$) at
  $K=20$, by modality.}
  \label{tab:v19-rivals-rmse}
\begin{tabular}{lccccc}
  \toprule
  Modality & $n$ & RMSE $K{=}10$ & RMSE $K{=}20$ & floor & within $0.003$ by $K{=}20$ \\
  \midrule
  Vision             & 15 & 0.049 & 0.047 & 0.045 & 13/15 \\
  Audio              & 3  & 0.081 & 0.076 & 0.064 & 0/3   \\
  Masked LM          & 4  & 0.064 & 0.062 & 0.061 & 4/4   \\
  Autoregressive LM  & 14 & 0.080 & 0.075 & 0.075 & 14/14 \\
  \midrule
  All                & 36 & 0.065 & 0.062 & 0.060 & 31/36 \\
  \bottomrule
\end{tabular}
\end{table}

\section{Heavy tail statistics and renormalization}
\label{app sec: lambda}

The theory of Sect.~5 treats the standardized centroid projections $s_\mu$ as
jointly Gaussian. Empirically, the variables $s_\mu$ (see Table.~\ref{tab:defs} for the definition) have a pronounced right tail. Since a
prediction is decided by the largest rival score $\max_\mu t_\mu$, this
positive tail controls the error rate, and it is exactly where a Gaussian
underestimates the extreme value. Fitting the pooled positive projections gives a
power-law
\begin{equation}
p(s)\;\propto\;\big(a^2+s^2\big)^{-\nu/2}
\label{eq:v19-halft}
\end{equation}
obtained by a maximum-likelihood fit of a Student-t distribution of parameters
$a,\nu$ to the positive $s$ values. Here $\nu$ is the tail (power-law) exponent, a
smaller $\nu$ means a heavier tail, and $\nu\!\to\!\infty$ recovers
a Gaussian. A heavier tail inflates the extreme rival projections that produce
errors, so the Gaussian theory is over-optimistic; the single global rescaling
$R\to\lambda R$ used in Section~5 (fit per model) compensates for this by
enlarging the effective within-class radius, and the fitted $\lambda$ tracks the
tail heaviness $1/\nu$ (see Fig.\ref{fig:theory} and Table \ref{tab:v19-nu}).

\begin{figure}[t]
  \centering
  \includegraphics[width=\linewidth]{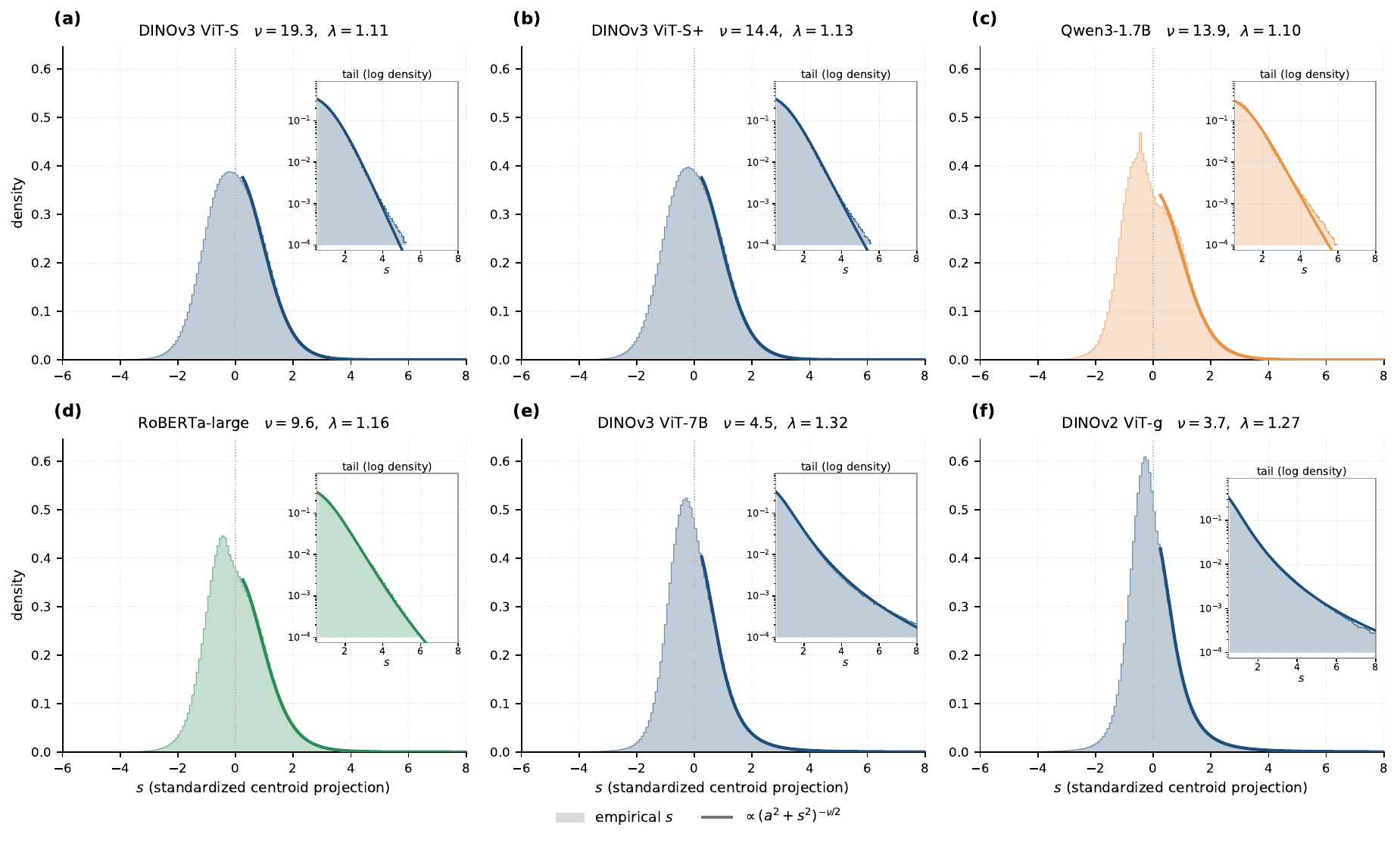}
  \caption{\textbf{Standardized centroid projections have a heavy right tail
  that varies across models.} Density of the standardized projection $s$ (filled)
  with its fitted half-Student-t right tail $p(s)\propto(a^2+s^2)^{-\nu/2}$
  (solid), $K=20$. Panels run light $\to$ heavy tail (left to
  right, top to bottom), with the fitted exponent $\nu$ and model in each title.
  Color encodes modality (vision blue, autoregressive LM orange, masked LM
  green). As $\nu$ decreases, the peak sharpens, and the right tail fattens.}
  \label{fig:v19-sdensity}
\end{figure}

Figure~\ref{fig:v19-sdensity} contrasts the density of $s$ across six backbones
spanning three modalities, ordered from the lightest to the heaviest tail, each
with its fitted curve~\eqref{eq:v19-halft}.

In Sec.~\ref{sec:theory:lambda} we claim that without the additional rescaling of the variability by $\lambda$, the theory is overoptimistic. This is demonstrated explicitly in Table~\ref{tab:lambda=1}. The theory with $\lambda=1$ consistently predicts accuracy high by $6\%$ on average across all backbones, compared to only $0.5\%$ with $\lambda$.

Table~\ref{tab:v19-nu} lists the tail exponent $\nu$ for all $36$ backbones,
grouped by modality and sorted from heaviest to lightest tail, alongside the
rescaling $\lambda$. The exponent varies substantially, from $\nu\approx3.7$
(DINOv2 ViT-g, the heaviest tail on the roster) to $\nu\approx22$ (Qwen3-0.6B, nearly Gaussian); the vision backbones span the widest range, while the audio and language backbones have larger $\nu$ (lighter tail) and less variability between models.
Across the roster the heaviness $1/\nu$ correlates strongly with the fitted
rescaling $\lambda$ (Pearson $r=0.90$ over the $36$ backbones): the heaviest-tail
models require the largest correction ($\lambda$ up to $1.34$) and the
near-Gaussian ones need almost none. The main text
shows this $\lambda$-versus-$1/\nu$ relationship as a scatter (Fig.~\ref{fig:theory}).

\begin{table}[H]
  \centering
  \small
  \caption{\textbf{Without global radius rescaling, the theory is over-optimistic.}
  Per-class accuracy predicted by the centroid-aligned theory of
  Sec.~\ref{sec:theory:lambda}, evaluated at $\lambda=1$ (no rescaling of the
  variability) and at the fitted $\lambda$, compared with the empirical accuracy. The theoretical accuracy with $\lambda=1$ is higher than the empirical one, while the theory with $\lambda$ is within $1\%$ of the empirical one, with significantly better RMSE.}
  \label{tab:lambda=1}
  \begin{tabular}{lccccccc}
    \toprule
    & & & \multicolumn{2}{c}{$\lambda=1$} & \multicolumn{2}{c}{fitted $\lambda$} \\
    \cmidrule(lr){4-5}\cmidrule(lr){6-7}
    Modality & $n$ & $\langle\mathrm{Acc}^{\mathrm{}}\rangle$ & $\langle\mathrm{Acc}^{\mathrm{th}}\rangle$ & RMSE & $\langle\mathrm{Acc}^{\mathrm{th}}\rangle$ & RMSE & $\langle\lambda\rangle$ \\
    \midrule
    Vision            & 15 & 0.812 & 0.876 & 0.080 & 0.819 & 0.047 & 1.228 \\
    Audio             & 3  & 0.613 & 0.657 & 0.087 & 0.617 & 0.076 & 1.084 \\
    Masked LM         & 4  & 0.715 & 0.785 & 0.090 & 0.724 & 0.062 & 1.131 \\
    Autoregressive LM & 14 & 0.522 & 0.583 & 0.096 & 0.524 & 0.075 & 1.111 \\
    \midrule
    All               & 36 & 0.672 & 0.734 & 0.088 & 0.677 & 0.062 & 1.160 \\
    \bottomrule
  \end{tabular}
\end{table}
$\quad$
\begin{table}[H]
\centering
\small
\setlength{\tabcolsep}{5pt}
\renewcommand{\arraystretch}{1.05}
\begin{tabular}[t]{lcc}
\toprule
\textbf{Backbone} & \textbf{$\nu$} & \textbf{$\lambda$} \\
\midrule
\multicolumn{3}{l}{\emph{Vision (ImageNet-1K)}} \\
\quad DINOv2 ViT-g                 & $3.7$  & $1.272$ \\
\quad ConvNeXt-B                   & $4.0$  & $1.254$ \\
\quad ViT-B                        & $4.1$  & $1.344$ \\
\quad DINOv3 CNX-L                 & $4.4$  & $1.305$ \\
\quad DINOv3 ViT-7B                & $4.5$  & $1.316$ \\
\quad DINOv3 CNX-B                 & $4.5$  & $1.296$ \\
\quad ResNet50                     & $4.6$  & $1.228$ \\
\quad DINOv3 CNX-S                 & $5.2$  & $1.229$ \\
\quad DINOv3 ViT-H$^{+}$           & $5.3$  & $1.297$ \\
\quad DINOv3 CNX-T                 & $6.8$  & $1.148$ \\
\quad DINOv3 ViT-L                 & $7.0$  & $1.221$ \\
\quad DINOv3 ViT-B                 & $9.0$  & $1.158$ \\
\quad DINOv3 ViT-S$^{+}$           & $14.4$ & $1.134$ \\
\quad CLIP-L                       & $16.4$ & $1.108$ \\
\quad DINOv3 ViT-S                 & $19.3$ & $1.114$ \\
\cmidrule(lr){1-3}
\quad \textit{$r(\lambda,\,1/\nu)$} & \multicolumn{2}{c}{$\mathbf{0.888}$} \\
\midrule
\multicolumn{3}{l}{\emph{Audio (VGGSound)}} \\
\quad SSLAM                        & $7.9$  & $1.106$ \\
\quad CED                         & $7.9$  & $1.089$ \\
\quad AST                          & $9.5$  & $1.057$ \\
\cmidrule(lr){1-3}
\quad \textit{$r(\lambda,\,1/\nu)$} & \multicolumn{2}{c}{$\mathbf{0.948}$} \\
\bottomrule
\end{tabular}
\hfill
\begin{tabular}[t]{lcc}
\toprule
\textbf{Backbone} & \textbf{$\nu$} & \textbf{$\lambda$} \\
\midrule
\multicolumn{3}{l}{\emph{Masked LM (TinyStories, masked-word)}} \\
\quad RoBERTa-large                & $9.6$  & $1.157$ \\
\quad BERT-large                   & $10.1$ & $1.127$ \\
\quad ModernBERT-large             & $11.3$ & $1.124$ \\
\quad NomicBERT                    & $14.2$ & $1.115$ \\
\cmidrule(lr){1-3}
\quad \textit{$r(\lambda,\,1/\nu)$} & \multicolumn{2}{c}{$\mathbf{0.805}$} \\
\midrule
\multicolumn{3}{l}{\emph{Autoregressive LM (TinyStories, next-token)}} \\
\quad OLMo-2-13B                   & $10.2$ & $1.152$ \\
\quad Qwen2.5-32B                  & $10.9$ & $1.117$ \\
\quad Qwen2.5-7B                   & $11.2$ & $1.109$ \\
\quad Qwen3-8B                     & $11.3$ & $1.128$ \\
\quad Qwen2.5-1.5B                 & $11.3$ & $1.110$ \\
\quad Qwen3-14B                    & $11.3$ & $1.124$ \\
\quad Qwen3-32B                    & $11.4$ & $1.119$ \\
\quad Qwen2.5-14B                  & $11.5$ & $1.112$ \\
\quad Llama-3.1-8B                 & $12.4$ & $1.107$ \\
\quad Qwen2.5-3B                   & $12.6$ & $1.112$ \\
\quad Qwen3-4B                     & $12.7$ & $1.118$ \\
\quad Qwen3-1.7B                   & $13.9$ & $1.103$ \\
\quad Qwen2.5-0.5B                 & $19.2$ & $1.088$ \\
\quad Qwen3-0.6B                   & $21.9$ & $1.059$ \\
\cmidrule(lr){1-3}
\quad \textit{$r(\lambda,\,1/\nu)$} & \multicolumn{2}{c}{$\mathbf{0.887}$} \\
\midrule
\textbf{Overall}                   & \multicolumn{2}{c}{$\mathbf{0.901}$} \\
\bottomrule
\end{tabular}
\caption{\textbf{Right-tail exponent $\nu$ across all backbones.}
$\nu$ is the power-law exponent of the fitted right tail
$p(s)\propto(a^2+s^2)^{-\nu/2}$; a smaller $\nu$
means a heavier tail. $\lambda$ is the global radius rescaling of the theory.
Grouped by modality, sorted from heaviest to lightest tail. Pearson correlations between $\lambda$ and $1/\nu$ are high in every family ($r=0.89,0.95,0.81,0.89$ in vision, audio, masked language models and autoregressive models, respectively) and globally over all models ($r=0.90$).}
\label{tab:v19-nu}
\end{table}

\begin{table}[H]
\centering
\small
\setlength{\tabcolsep}{5pt}
\renewcommand{\arraystretch}{1.05}
\begin{tabular}{lrrrrrrr}
\toprule
\textbf{Backbone} & \textbf{$P$} & \textbf{$N$} & Acc & Acc$^{\mathrm{th}}$ & \textbf{$r$} & \textbf{$\lambda$} & RMSE \\
\midrule
\multicolumn{8}{l}{\emph{Vision (ImageNet-1K)}} \\
\quad ConvNeXt-B                   & 1000 & 1024 & $0.943$ & $0.952$ & $0.954$ & $1.254$ & $0.0256$ \\
\quad ResNet50                     & 1000 & 2048 & $0.861$ & $0.871$ & $0.936$ & $1.228$ & $0.0454$ \\
\quad DINOv3 ViT-H$^{+}$           & 1000 & 1280 & $0.856$ & $0.862$ & $0.960$ & $1.297$ & $0.0390$ \\
\quad DINOv3 ViT-7B                & 1000 & 4096 & $0.855$ & $0.861$ & $0.951$ & $1.316$ & $0.0427$ \\
\quad ViT-B                        & 1000 &  768 & $0.853$ & $0.861$ & $0.959$ & $1.344$ & $0.0378$ \\
\quad DINOv3 ViT-L                 & 1000 & 1024 & $0.845$ & $0.852$ & $0.964$ & $1.221$ & $0.0384$ \\
\quad DINOv3 CNX-L                 & 1000 & 1536 & $0.814$ & $0.819$ & $0.961$ & $1.305$ & $0.0440$ \\
\quad DINOv3 CNX-B                 & 1000 & 1024 & $0.804$ & $0.808$ & $0.958$ & $1.296$ & $0.0455$ \\
\quad DINOv2 ViT-g                 & 1000 & 1536 & $0.795$ & $0.804$ & $0.941$ & $1.272$ & $0.0631$ \\
\quad DINOv3 ViT-B                 & 1000 &  768 & $0.794$ & $0.800$ & $0.954$ & $1.158$ & $0.0460$ \\
\quad CLIP-L                       & 1000 &  768 & $0.792$ & $0.804$ & $0.936$ & $1.108$ & $0.0596$ \\
\quad DINOv3 CNX-S                 & 1000 &  768 & $0.786$ & $0.793$ & $0.956$ & $1.229$ & $0.0481$ \\
\quad DINOv3 CNX-T                 & 1000 &  768 & $0.740$ & $0.749$ & $0.937$ & $1.148$ & $0.0602$ \\
\quad DINOv3 ViT-S$^{+}$           & 1000 &  384 & $0.738$ & $0.742$ & $0.951$ & $1.134$ & $0.0504$ \\
\quad DINOv3 ViT-S                 & 1000 &  384 & $0.706$ & $0.710$ & $0.944$ & $1.114$ & $0.0555$ \\
\cmidrule(lr){1-8}
\quad \textit{mean}                &      &      & $\mathbf{0.812}$ & $\mathbf{0.819}$ & $\mathbf{0.951}$ & $\mathbf{1.228}$ & $\mathbf{0.0468}$ \\
\midrule
\multicolumn{8}{l}{\emph{Audio (VGGSound)}} \\
\quad SSLAM                        &  310 &  768 & $0.628$ & $0.631$ & $0.952$ & $1.106$ & $0.0720$ \\
\quad CED                        &  310 &  768 & $0.628$ & $0.632$ & $0.947$ & $1.089$ & $0.0739$ \\
\quad AST                          &  310 &  768 & $0.583$ & $0.587$ & $0.944$ & $1.057$ & $0.0819$ \\
\cmidrule(lr){1-8}
\quad \textit{mean}                &      &      & $\mathbf{0.613}$ & $\mathbf{0.617}$ & $\mathbf{0.948}$ & $\mathbf{1.084}$ & $\mathbf{0.0759}$ \\
\midrule
\multicolumn{8}{l}{\emph{Masked LM (TinyStories, masked-word)}} \\
\quad RoBERTa-large                &  320 & 1024 & $0.767$ & $0.776$ & $0.944$ & $1.157$ & $0.0542$ \\
\quad ModernBERT-large             &  320 & 1024 & $0.734$ & $0.743$ & $0.942$ & $1.124$ & $0.0614$ \\
\quad BERT-large                   &  320 & 1024 & $0.710$ & $0.720$ & $0.942$ & $1.127$ & $0.0630$ \\
\quad NomicBERT                    &  320 &  768 & $0.649$ & $0.658$ & $0.945$ & $1.115$ & $0.0683$ \\
\cmidrule(lr){1-8}
\quad \textit{mean}                &      &      & $\mathbf{0.715}$ & $\mathbf{0.724}$ & $\mathbf{0.943}$ & $\mathbf{1.131}$ & $\mathbf{0.0617}$ \\
\midrule
\multicolumn{8}{l}{\emph{Autoregressive LM (TinyStories, next-token)}} \\
\quad Qwen2.5-32B                  &  320 & 5120 & $0.567$ & $0.572$ & $0.945$ & $1.117$ & $0.0722$ \\
\quad Qwen3-32B                    &  320 & 5120 & $0.566$ & $0.569$ & $0.955$ & $1.119$ & $0.0665$ \\
\quad OLMo-2-13B                   &  320 & 5120 & $0.563$ & $0.567$ & $0.948$ & $1.152$ & $0.0701$ \\
\quad Qwen3-14B                    &  320 & 5120 & $0.556$ & $0.558$ & $0.954$ & $1.124$ & $0.0694$ \\
\quad Qwen2.5-14B                  &  320 & 5120 & $0.552$ & $0.556$ & $0.942$ & $1.112$ & $0.0758$ \\
\quad Llama-3.1-8B                 &  320 & 4096 & $0.550$ & $0.553$ & $0.943$ & $1.107$ & $0.0772$ \\
\quad Qwen2.5-7B                   &  320 & 3584 & $0.544$ & $0.550$ & $0.940$ & $1.109$ & $0.0756$ \\
\quad Qwen3-8B                     &  320 & 4096 & $0.540$ & $0.542$ & $0.955$ & $1.128$ & $0.0693$ \\
\quad Qwen2.5-3B                   &  320 & 2048 & $0.528$ & $0.531$ & $0.946$ & $1.112$ & $0.0785$ \\
\quad Qwen3-4B                     &  320 & 2560 & $0.519$ & $0.520$ & $0.953$ & $1.118$ & $0.0697$ \\
\quad Qwen2.5-1.5B                 &  320 & 1536 & $0.491$ & $0.494$ & $0.931$ & $1.110$ & $0.0822$ \\
\quad Qwen3-1.7B                   &  320 & 2048 & $0.474$ & $0.474$ & $0.949$ & $1.103$ & $0.0747$ \\
\quad Qwen2.5-0.5B                 &  320 &  896 & $0.452$ & $0.450$ & $0.944$ & $1.088$ & $0.0853$ \\
\quad Qwen3-0.6B                   &  320 & 1024 & $0.406$ & $0.403$ & $0.939$ & $1.059$ & $0.0839$ \\
\cmidrule(lr){1-8}
\quad \textit{mean}                &      &      & $\mathbf{0.522}$ & $\mathbf{0.524}$ & $\mathbf{0.946}$ & $\mathbf{1.111}$ & $\mathbf{0.0750}$ \\
\midrule
\textbf{Overall}                   &      &      & $\mathbf{0.672}$ & $\mathbf{0.677}$ & $\mathbf{0.948}$ & $\mathbf{1.160}$ & $\mathbf{0.0618}$ \\
\bottomrule
\end{tabular}
\caption{\textbf{Centroid-aligned theory}: The theory was tested on all 36 models. For each backbone, we report: number of classes $P$, feature
dimension $N$, empirical prototype accuracy, per-class Pearson $r$ between
predicted and empirical accuracy, the fitted global rescaling $\lambda$, and the
per-class prediction error RMSE. The theory consistently predicts the accuracy well across all models ($r>0.93$).}
\label{app:manymodels-table}
\end{table}


\begin{figure}[H]
  \centering
  \includegraphics[width=\linewidth]{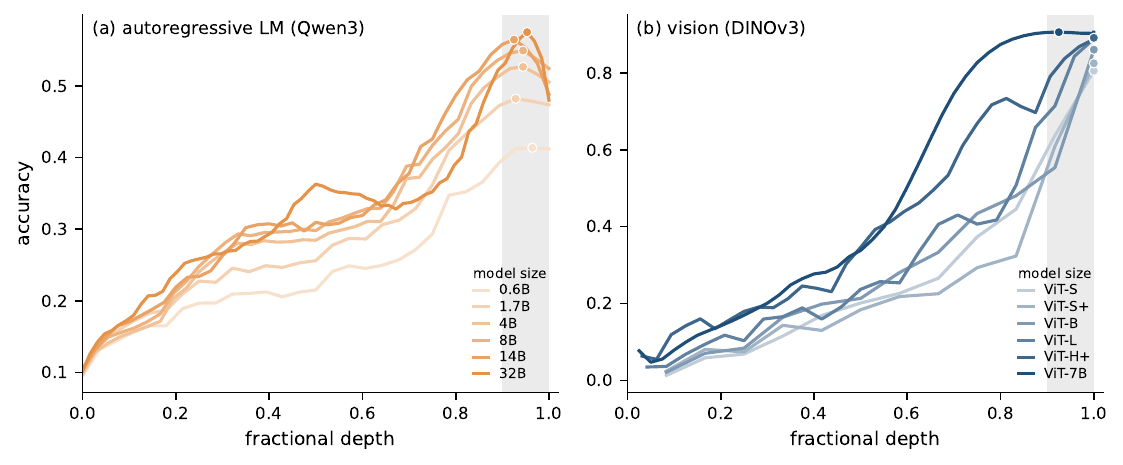}
  \caption{\textbf{The best prototype layer is a few blocks before the final one
  for language, but is the final block for vision.} Per-layer prototype
  classification accuracy versus fractional depth. Color runs from light (small) to dark
  (large) within each modality, and the peak performance is marked. \textbf{(a)}~The six Qwen3
  autoregressive language models; the optimal accuracy is a few blocks before the final layer, with a big drop at the final layer in most models. \textbf{(b)}~the six
  DINOv3 vision transformers. The final layer is the best, with a plateau on DINOv3 7B.}
  \label{fig:v19-depth}
\end{figure}
\section{Depth analysis}
\label{app sec: qwen}

The prototype classifier of Sec.~5 reads out a single hidden layer. Here we analyze how prototype classification accuracy and geometry change across depth, and identify the best layer for this task. For the
autoregressive language models the best layer for content-word prediction is
consistently a few blocks before the final one, and accuracy dips at the last block; for vision the final block is best (in DINOv3 7B, there is a plateau). We subsequently
analyze the geometry of the largest language model, Qwen3-32B across depth, similar to the analysis in Sec.~\ref{sec:scaling} for the Qwen3 family across scale. 
Similar to the scale effect, we see that the best layer has large true centroid correlation compared to rivals $\sigma_k/\sigma_\mu^*$, decorrelates centroids (small $g_\mu^*$), and has small correlations between projections $\rho_\mu^*$. We note that, as observed before with scale, the class radius $R$ does not predict the overall best accuracy.

\begin{table}[H]
  \centering
  \footnotesize
  \caption{\textbf{Qwen3-32B across depth.}
  Class-averaged geometric measures for
  the seven last blocks of Qwen3-32B ($64$ blocks). $^{*}$ marks each
  class's hardest competitor $\mu^{\star}$, measured by $\sigma_\mu$, the projection standard deviation.
  $\langle R\rangle$ is the mean class radius; $\sigma_k/\sigma^{*}_{\mu}$ is the ratio between the standard deviation of the projection onto the true-class and hardest-rival directions. $\sigma^{*}_{\mu}$ is the hardest rival projected std,
  $g^{*}_{\mu}=\hat{\mathbf{c}}_k\!\cdot\!\hat{\mathbf{c}}_{\mu^{\star}}$ the
  centroid overlap and $\rho^{*}_{\mu}$ the correlation between projections. $\nu$ is the
  right-tail exponent. Best measure highlighted in bold.}
  \label{tab:v19-depth-band}
  \begin{tabular}{l r r r r r r r}
    \toprule
    Layer & $\langle R\rangle$ & $\sigma_k/\sigma^{*}_{\mu}$ & $\sigma^{*}_{\mu}$ & $g^{*}_{\mu}$ & $\rho^{*}_{\mu}$ & $\nu$ & Acc \\
    \midrule
    L58 (0.91) & 2.40 & 0.993 & \textbf{0.172} & 0.514 & 0.496 & \textbf{10.9} & 0.521 \\
    L59 (0.92) & 2.32 & 0.996 & 0.175 & 0.511 & 0.502 & 11.0 & 0.542 \\
    L60 (0.94) & 2.24 & 0.999 & 0.175 & 0.510 & 0.474 & \textbf{10.9} & 0.557 \\
    \textbf{L61 (0.95)} & 2.20 & \textbf{1.007} & 0.175 & \textbf{0.503} & 0.451 & 11.4 & \textbf{0.566} \\
    L62 (0.97) & 2.17 & 0.957 & 0.181 & 0.517 & 0.463 & 13.7 & 0.553 \\
    L63 (0.98) & 2.16 & 0.894 & 0.191 & 0.521 & \textbf{0.448} & 18.1 & 0.528 \\
    L64 (final) & \textbf{2.10} & 0.754 & 0.224 & 0.533 & 0.471 & 98.5 & 0.470 \\
    \bottomrule
  \end{tabular}
\end{table}

\section{Theoretical Predictions Across Models and Modalities}
\label{app:manymodels}

The main text validates the centroid-aligned variability theory (Sec.~\ref{sec:theory}) on a
handful of representative backbones. Here we apply it, unchanged, to a comprehensive list of 36 state-of-the-art models, spanning four modalities: 15 vision encoders (ImageNet), 14 autoregressive language models and 4 masked language models
(TinyStories), and 3 audio encoders
(VGGSound). The models have different architectures (transformers and convolutional neural networks), training methods (supervised, self-supervised, and contrastive), and training datasets. For every model, we compute the per-class prediction
with a single fitted global rescaling $\lambda$ and the measured quantities the theory predicts, and compare it to the empirical prototype accuracy class by class. No
per-model tuning beyond the scalar $\lambda$ is used.

Table~\ref{app:manymodels-table} reports, for each backbone, the number of
classes $P$, the feature dimension $N$, the empirical prototype accuracy, the
per-class Pearson correlation $r$ between predicted and empirical accuracy, the
fitted $\lambda$, and RMSE. 

Across all four modalities the per-class correlation between the theory and the
empirical accuracy is high: $r=0.951$ (vision), $0.946$
(autoregressive), $0.943$ (masked) and $0.948$ (audio). Per-class errors have nearly zero bias (mean per modality bias $0.002-0.009$)
and only a few percent mean deviation (mean per modality RMSE $0.047$--$0.076$). 
Thus our theory predicts prototype accuracy class by class across vision, language and
audio representations.

For each model we also reproduce the two per-class panels of the theory vs the empirical accuracy as in
Fig.~\ref{fig:theory}. The four figures show all backbones grouped by modality. The theory not only tracks the general $R$ trend, but often closely follows the individual pattern of the model.

\begin{figure}[t]
\centering
\includegraphics[width=\linewidth]{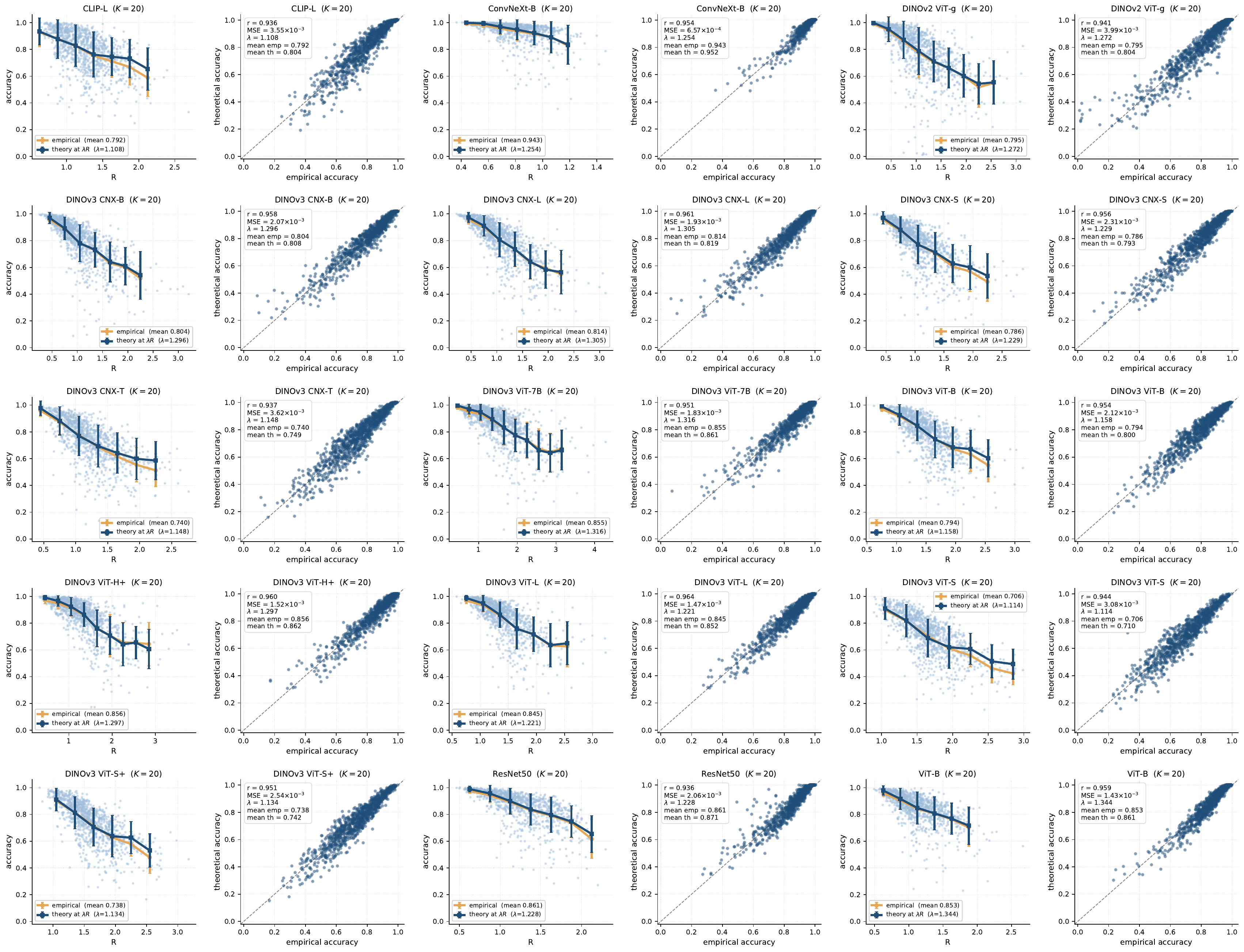}
\caption{15 vision models. For each backbone: (left)
per-class accuracy versus empirical radius $R$, empirical binned mean  and theory binned mean nearly overlap in most models; (right)
per-class predicted versus empirical accuracy with $r$, RMSE and $\lambda$. The dashed line is $y=x$.
Mean Pearson $r=0.951$, mean RMSE $=0.047$. 
}
\label{app:grid-vision}
\end{figure}

\begin{figure}
\centering
\includegraphics[width=\linewidth]{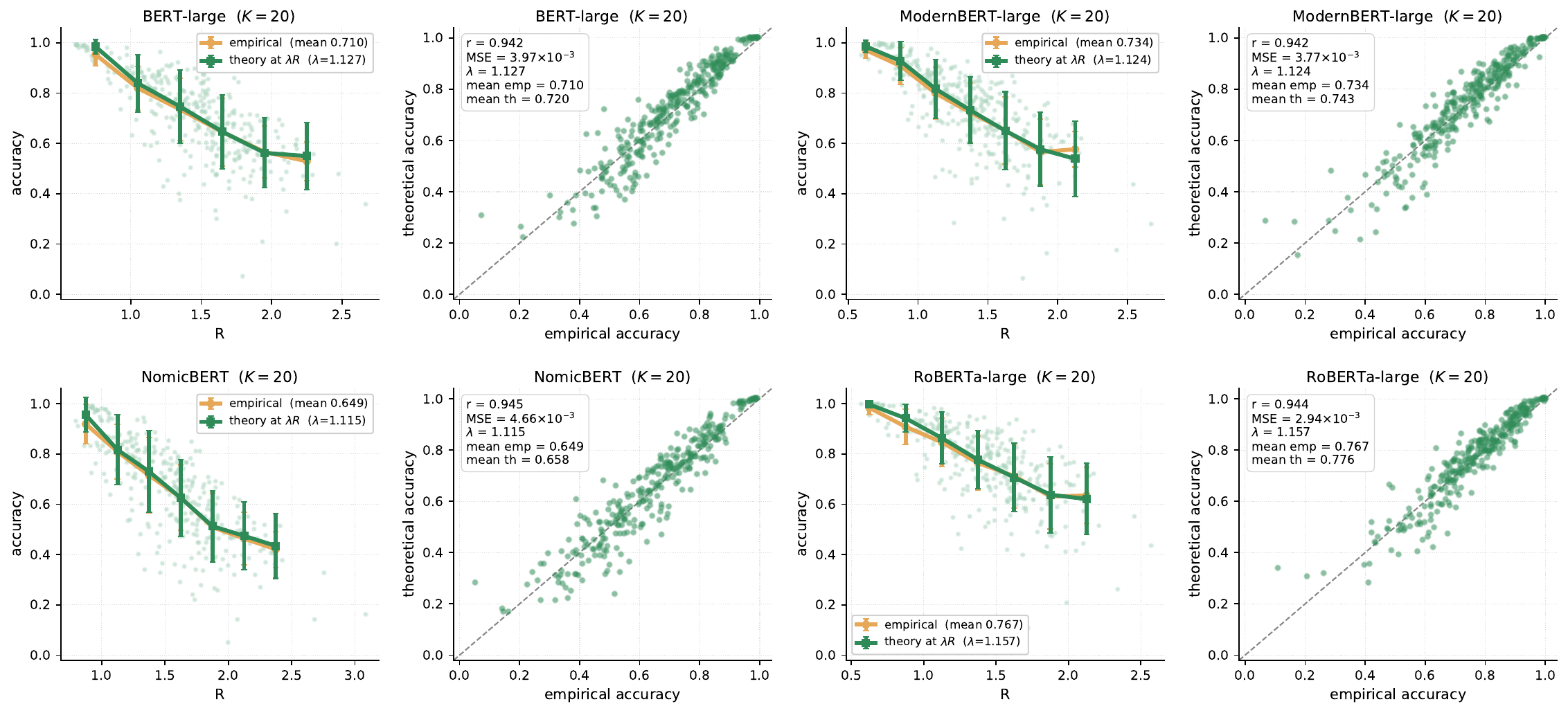}
\caption{Masked language models (4 models, TinyStories content-word prediction,
best layer). Panels as in Fig.~\ref{app:grid-vision}. Mean $r=0.943$, mean
RMSE $=0.062$. 
}
\label{app:grid-masked}
\end{figure}

\begin{figure}
\centering
\includegraphics[width=\linewidth]{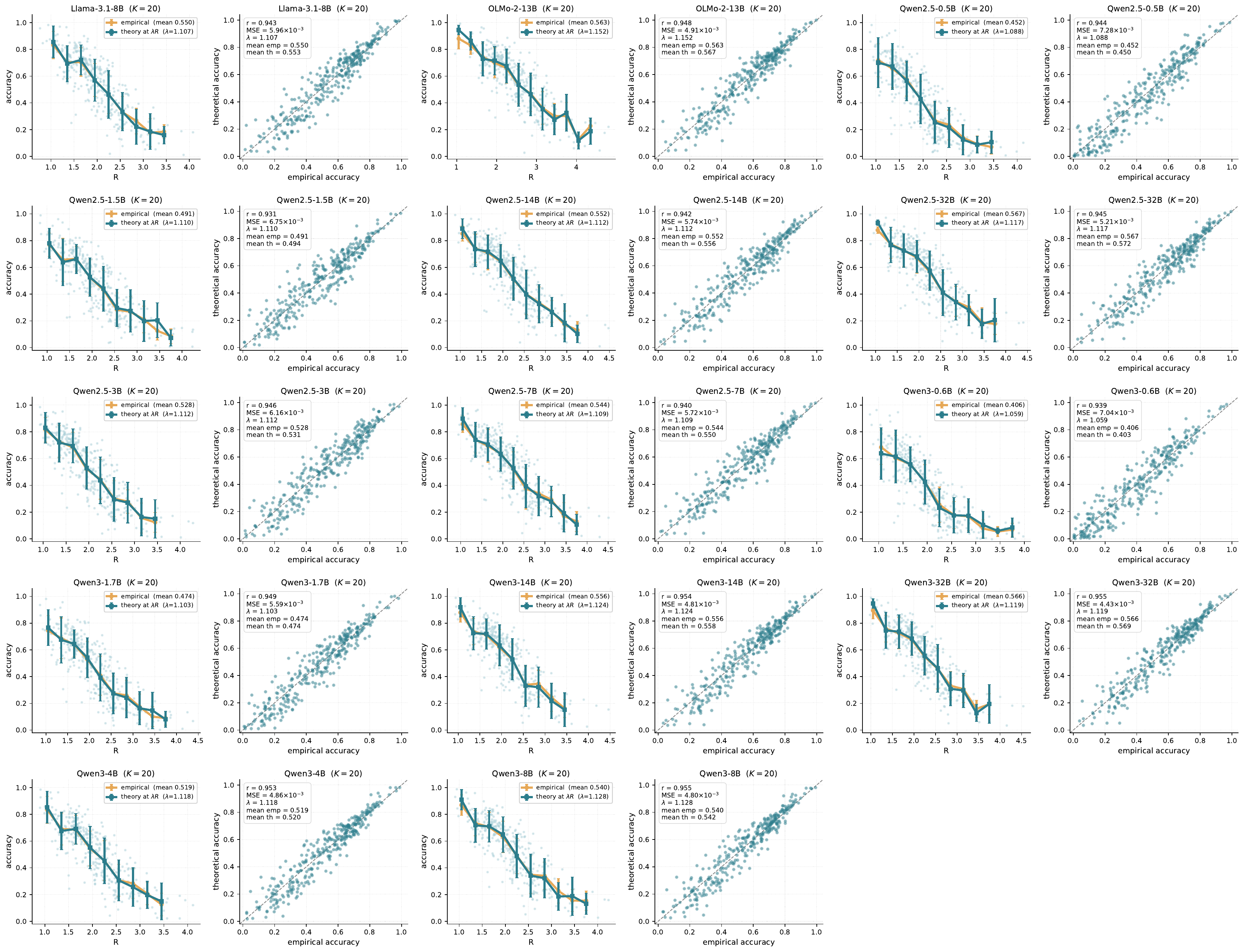}
\caption{Autoregressive language models (14 models, TinyStories content-word
prediction, best layer). Panels as in Fig.~\ref{app:grid-vision}; equally spaced
$R$ bins fixed for this modality. Mean $r=0.946$, mean RMSE $=0.075$. 
}
\label{app:grid-ar}
\end{figure}
\begin{figure}
\centering
\includegraphics[width=\linewidth]{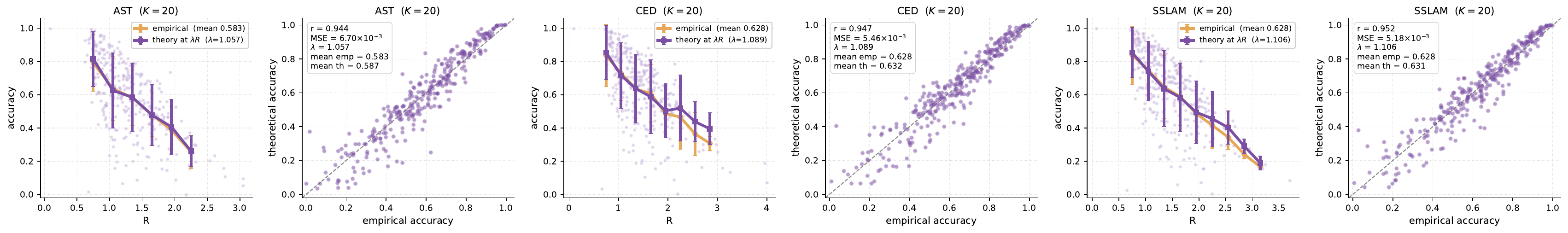}
\caption{Audio (3 encoders, VGGSound training split, final layer). Panels as in
Fig.~\ref{app:grid-vision}. Mean $r=0.948$, mean RMSE $=0.076$. 
}
\label{app:grid-audio}
\end{figure}

\section{Variants of the theory}
\label{app:variants}

The prototype-accuracy theory of the main text (Theorem~5.5) is one point in a small
family of predictors, obtained by making different assumptions about the covariance of the standardized centroid projections $ s_\mu$. Here we compare the main theory
against a simpler baseline and one richer refinement. All three predictors are evaluated
at a fixed rival $K=20$, on the same 36 backbones
and the same per-class accuracy targets. The global rescaling factor $\lambda$ was fitted separately for each variant.

\textbf{Uncorrelated s}: The model is the same as the one presented in Sec.~\ref{sec:theory}, with $\rho_{\mu\neq k}=0$, so that different $s_\mu$ are uncorrelated. The uncorrelated theory fits the data significantly worse than the theory of Sec.~\ref{sec:theory}, where we consider a simple off-diagonal structure, and thus it is not used in the main text.

\textbf{Cluster expansion:} We start from Eq.~\ref{eq:cal-orthant}. We assume that $\Sigma_{\mu\nu}$ has the structure identified in Sec.~\ref{sec:theory}, plus a small correction on the terms outside the diagonal, $\Sigma=\Sigma_0+\delta\Sigma,\enspace \delta\Sigma_{\mu\mu}=0$. We shift the logits to have zero mean for convenience, and write the orthant probability conditioned on $t_k$, which has a diagonal element from the original structure plus the small correction

\begin{equation}
\mathrm{Acc}^{\mathrm{th}}(R)\propto\prod_{\mu\in\mathcal{R}_{K}}\int^{t_{k}-f_{\mu}\left(t_{k}\right)-g_{\mu}}_{-\infty}\!\frac{dt_{\mu}}{\sqrt{2\pi}}\,\exp\!\left(-\frac{1}{2}\sum_{\mu,\nu\in\mathcal{R}_{K}}t_{\mu}\left(\delta_{\mu\nu}\Lambda^{2}_{\mu}+\delta\Sigma_{\mu\nu}\right)^{-1}t_{\nu}\right)
\end{equation},

where $f_{\mu}\left(t_{k}\right)=\rho_{\mu}\sigma_{\mu}\left(t_{k}-1\right)/\sigma_{k}$ is the conditional mean similar to Sec.~\ref{app:caligned-proofs}, and $\Lambda_\mu$ is the conditional standard deviation $\Lambda_\mu=R\sigma_\mu\sqrt{1-\rho_\mu^2}$. Expanding to linear order in $\delta\Sigma_{\mu\nu}$:
\begin{equation}
\mathrm{Acc}^{\mathrm{th}}(R)=\prod_{\mu\in\mathcal{R}_{K}}\int^{t_{k}-f_{\mu}\left(t_{k}\right)-g_{\mu}}_{-\infty}\!\frac{dt_{\mu}}{\sqrt{2\pi\Lambda^{2}_{\mu}}}\,\exp\!\left(-\frac{1}{2}\sum_{\mu\in\mathcal{R}_{K}}\frac{t^{2}_{\mu}}{\Lambda^{2}_{\mu}}\right)\left(1+\frac{1}{2}\sum_{\mu\neq\nu}\frac{\delta\Sigma_{\mu\nu}t_{\mu}t_{\nu}}{\Lambda^{2}_{\mu}\Lambda^{2}_{\nu}}\right)
\end{equation}
Define the conditional margin 
\begin{equation}
F_\mu(t_k)=\frac{t_{k}-f_{\mu}\left(t_{k}\right)-g_{\mu}}{\Lambda_{\mu}},
\end{equation}
and perform the truncated Gaussian integrals
\begin{equation}
\mathrm{Acc}^{\mathrm{th}}(R)=\prod_{\mu\in\mathcal{R}_{K}}\Phi\left(F_{\mu}\left(t_{k}\right)\right)\left(1+\frac{1}{2}\sum_{\mu\neq\nu}\frac{\delta\Sigma_{\mu\nu}}{\Lambda_{\mu}\Lambda_{\nu}}m\left(F_{\mu}\left(t_{k}\right)\right)m\left(F_{\nu}\left(t_{k}\right)\right)\right)
\end{equation}
We used the common inverse Mills ratio $m(x)=\phi(x)/\Phi(x)$, where $\phi(x)$ is a standard normal PDF and $\Phi(x)$ is a standard normal CDF. Substitute $t_k=Rz\sigma_k+1$ and averaging over $z$ we get the final result:
\begin{equation}
    \mathrm{Acc}^{\mathrm{th}}(R)=\mathbb{E}_{z\sim\mathcal{N}\left(0,1\right)}\left[\prod_{\mu\in\mathcal{R}_{K}}\Phi\left(F_{\mu}\left(z\right)\right)\left(1+\frac{1}{2}\sum_{\mu\neq\nu}\frac{\delta\Sigma_{\mu\nu}}{\Lambda_{\mu}\Lambda_{\nu}}m\left(F_{\mu}\left(z\right)\right)m\left(F_{\nu}\left(z\right)\right)\right)\right]
\end{equation}
With
\begin{equation} F_{\mu}\left(z\right)=\frac{\sigma_{k}-\rho_{\mu}\sigma_{\mu}}{\Lambda_{\mu}/R}z+\frac{1-g_{\mu}}{\Lambda_{\mu}}
\end{equation}
Although the cluster expansion theory remains analytic, and fits the data slightly better than the original theory derived in Sec.~\ref{sec:theory}, it adds another $K(K-1)/2$ parameters $\delta\Sigma_{\mu\nu}$ that need to be measured, and it is less transparent than the original simpler theory. Thus, we bring it here as an additional theoretical result with its performance on the 36 backbones we tested (Table.~\ref{tab:variants-lambda-rmse}), but do not use it in the main text.

\begin{table}[H]
\centering
\small
\setlength{\tabcolsep}{5pt}
\renewcommand{\arraystretch}{1.05}
\begin{tabular}{lcccccc}
\toprule
& \multicolumn{2}{c}{\textbf{Uncorrelated}} & \multicolumn{2}{c}{\textbf{Main theory}} & \multicolumn{2}{c}{\textbf{Cluster expansion}} \\
\cmidrule(lr){2-3}\cmidrule(lr){4-5}\cmidrule(lr){6-7}
\textbf{Backbone} & \textbf{$\lambda$} & RMSE & \textbf{$\lambda$} & RMSE & \textbf{$\lambda$} & RMSE \\
\midrule
\multicolumn{7}{l}{\emph{Vision (ImageNet-1K)}} \\
\quad ConvNeXt-B          & $0.979$ & $0.0493$ & $1.254$ & $0.0256$ & $1.281$ & $0.0234$ \\
\quad ResNet50            & $1.114$ & $0.0589$ & $1.228$ & $0.0454$ & $1.320$ & $0.0434$ \\
\quad DINOv3 ViT-H$^{+}$  & $1.212$ & $0.0578$ & $1.297$ & $0.0390$ & $1.317$ & $0.0377$ \\
\quad DINOv3 ViT-7B       & $1.248$ & $0.0574$ & $1.316$ & $0.0427$ & $1.337$ & $0.0402$ \\
\quad ViT-B               & $1.196$ & $0.0593$ & $1.344$ & $0.0378$ & $1.393$ & $0.0339$ \\
\quad DINOv3 ViT-L        & $0.996$ & $0.0733$ & $1.221$ & $0.0384$ & $1.248$ & $0.0366$ \\
\quad DINOv3 CNX-L        & $1.160$ & $0.0706$ & $1.305$ & $0.0440$ & $1.360$ & $0.0417$ \\
\quad DINOv3 CNX-B        & $1.087$ & $0.0799$ & $1.296$ & $0.0455$ & $1.337$ & $0.0420$ \\
\quad DINOv2 ViT-g        & $1.114$ & $0.1166$ & $1.272$ & $0.0631$ & $1.311$ & $0.0561$ \\
\quad DINOv3 ViT-B        & $0.864$ & $0.0929$ & $1.158$ & $0.0460$ & $1.196$ & $0.0416$ \\
\quad CLIP-L              & $0.671$ & $0.1188$ & $1.108$ & $0.0596$ & $1.199$ & $0.0471$ \\
\quad DINOv3 CNX-S        & $0.966$ & $0.0935$ & $1.229$ & $0.0481$ & $1.275$ & $0.0450$ \\
\quad DINOv3 CNX-T        & $0.855$ & $0.1224$ & $1.148$ & $0.0602$ & $1.209$ & $0.0535$ \\
\quad DINOv3 ViT-S$^{+}$  & $0.861$ & $0.0981$ & $1.134$ & $0.0504$ & $1.179$ & $0.0438$ \\
\quad DINOv3 ViT-S        & $0.838$ & $0.1061$ & $1.114$ & $0.0555$ & $1.173$ & $0.0464$ \\
\cmidrule(lr){1-7}
\quad \textit{mean}       & $\mathbf{1.011}$ & $\mathbf{0.0837}$ & $\mathbf{1.228}$ & $\mathbf{0.0468}$ & $\mathbf{1.276}$ & $\mathbf{0.0422}$ \\
\midrule
\multicolumn{7}{l}{\emph{Audio (VGGSound)}} \\
\quad SSLAM               & $0.920$ & $0.1194$ & $1.106$ & $0.0720$ & $1.179$ & $0.0555$ \\
\quad CED               & $0.894$ & $0.1164$ & $1.089$ & $0.0739$ & $1.183$ & $0.0599$ \\
\quad AST                 & $0.861$ & $0.1256$ & $1.057$ & $0.0819$ & $1.173$ & $0.0617$ \\
\cmidrule(lr){1-7}
\quad \textit{mean}       & $\mathbf{0.892}$ & $\mathbf{0.1205}$ & $\mathbf{1.084}$ & $\mathbf{0.0759}$ & $\mathbf{1.178}$ & $\mathbf{0.0590}$ \\
\midrule
\multicolumn{7}{l}{\emph{Masked LM (TinyStories, masked-word)}} \\
\quad RoBERTa-large       & $0.943$ & $0.0856$ & $1.157$ & $0.0542$ & $1.235$ & $0.0477$ \\
\quad ModernBERT-large    & $0.884$ & $0.0985$ & $1.124$ & $0.0614$ & $1.225$ & $0.0511$ \\
\quad BERT-large          & $0.943$ & $0.0930$ & $1.127$ & $0.0630$ & $1.235$ & $0.0492$ \\
\quad NomicBERT           & $0.920$ & $0.1048$ & $1.115$ & $0.0683$ & $1.238$ & $0.0565$ \\
\cmidrule(lr){1-7}
\quad \textit{mean}       & $\mathbf{0.923}$ & $\mathbf{0.0955}$ & $\mathbf{1.131}$ & $\mathbf{0.0617}$ & $\mathbf{1.234}$ & $\mathbf{0.0511}$ \\
\midrule
\multicolumn{7}{l}{\emph{Autoregressive LM (TinyStories, next-token)}} \\
\quad Qwen2.5-32B         & $0.937$ & $0.1069$ & $1.117$ & $0.0722$ & $1.363$ & $0.0755$ \\
\quad Qwen3-32B           & $0.940$ & $0.1004$ & $1.119$ & $0.0665$ & $1.376$ & $0.0707$ \\
\quad OLMo-2-13B          & $0.986$ & $0.0922$ & $1.152$ & $0.0701$ & $1.393$ & $0.0726$ \\
\quad Qwen3-14B           & $0.953$ & $0.0984$ & $1.124$ & $0.0694$ & $1.383$ & $0.0716$ \\
\quad Qwen2.5-14B         & $0.933$ & $0.1093$ & $1.112$ & $0.0758$ & $1.370$ & $0.0773$ \\
\quad Llama-3.1-8B        & $0.923$ & $0.1061$ & $1.107$ & $0.0772$ & $1.373$ & $0.0750$ \\
\quad Qwen2.5-7B          & $0.946$ & $0.1114$ & $1.109$ & $0.0756$ & $1.350$ & $0.0780$ \\
\quad Qwen3-8B            & $0.963$ & $0.0947$ & $1.128$ & $0.0693$ & $1.363$ & $0.0703$ \\
\quad Qwen2.5-3B          & $0.956$ & $0.1058$ & $1.112$ & $0.0785$ & $1.357$ & $0.0780$ \\
\quad Qwen3-4B            & $0.950$ & $0.0993$ & $1.118$ & $0.0697$ & $1.347$ & $0.0685$ \\
\quad Qwen2.5-1.5B        & $0.953$ & $0.1166$ & $1.110$ & $0.0822$ & $1.330$ & $0.0865$ \\
\quad Qwen3-1.7B          & $0.940$ & $0.1003$ & $1.103$ & $0.0747$ & $1.337$ & $0.0718$ \\
\quad Qwen2.5-0.5B        & $0.937$ & $0.1072$ & $1.088$ & $0.0853$ & $1.350$ & $0.0808$ \\
\quad Qwen3-0.6B          & $0.900$ & $0.1096$ & $1.059$ & $0.0839$ & $1.311$ & $0.0772$ \\
\cmidrule(lr){1-7}
\quad \textit{mean}       & $\mathbf{0.944}$ & $\mathbf{0.1042}$ & $\mathbf{1.111}$ & $\mathbf{0.0750}$ & $\mathbf{1.357}$ & $\mathbf{0.0753}$ \\
\midrule
\textbf{Overall}          & $\mathbf{0.965}$ & $\mathbf{0.0960}$ & $\mathbf{1.160}$ & $\mathbf{0.0618}$ & $\mathbf{1.295}$ & $\mathbf{0.0574}$ \\
\bottomrule
\end{tabular}
\caption{Fitted global rescaling $\lambda$ and per-class prediction error RMSE for three
predictors at $K{=}20$, on all 36 backbones: the
uncorrelated-projections baseline ($\rho_\mu{=}0$), the main theory (rank-one true--rival correlation), and the first-order cluster expansion.
$\lambda$ is refitted separately for each predictor.}
\label{tab:variants-lambda-rmse}
\end{table}

\newpage

\end{document}